\documentclass[a4paper]{article}

\usepackage{amsfonts,amsmath,amssymb,indentfirst,mathrsfs,amscd}
\usepackage[mathscr]{eucal}
\usepackage{amsthm}
\usepackage{authblk}

\usepackage{lineno,hyperref}

\usepackage{enumerate}
\usepackage[utf8]{inputenc}
\usepackage[T1]{fontenc}
\usepackage{todonotes}
\usepackage{graphicx}
\usepackage{mdframed}
\usepackage[title]{appendix}

\usepackage{multirow}
\usepackage{interval}
\usepackage{array}
\usepackage{booktabs}
\usepackage{colortbl}
\usepackage{comment}
\usepackage{subcaption}
\usepackage{amsmath}
\usepackage{amssymb}
\usepackage{algorithm}
\usepackage[noend]{algpseudocode}
\usepackage{multirow}
\usepackage{amsthm}

\usepackage{url}

\newcommand{\RR}{\mathbb{R}} 

\newcommand\eatpunct[1]{}
\def\RR{\mathbb{R}}
\newtheorem{theorem}{Theorem}[section]
\newtheorem{proposition}[theorem]{Proposition}

\begin{document}

\title{Improving the quality of generative models\\ through Smirnov transformation}

\author[2,3]{Ángel González-Prieto}
\author[1]{Alberto Mozo}
\author[1]{Sandra Gómez-Canaval}
\author[1]{Edgar Talavera}

\affil[1]{Universidad Politécnica de Madrid.}
\affil[2]{Universidad Complutense de Madrid.}
\affil[3]{Instituto de Ciencias Matem\'aticas (CSIC-UAM-UCM-UC3M).}

\markboth{Journal of \LaTeX\ Class Files,~Vol.~14, No.~8, August~2015}%
{Shell \MakeLowercase{\textit{et al.}}: Bare Demo of IEEEtran.cls for IEEE Journals}

\maketitle

\begin{abstract}
Solving the convergence issues of Generative Adversarial Networks (GANs) is one of the most outstanding problems in generative models. In this work, we propose a novel activation function to be used as output of the generator agent. This activation function is based on the Smirnov probabilistic transformation and it is specifically designed to improve the quality of the generated data. 
In sharp contrast with previous works, our activation function provides a more general approach that deals not only with the replication of categorical variables but with any type of data distribution (continuous or discrete).
Moreover, our activation function is derivable and therefore, it can be seamlessly integrated in the backpropagation computations during the GAN training processes.
To validate this approach, we evaluate our proposal against two different data sets: a) an artificially rendered data set containing a mixture of discrete and continuous variables, and b) a real data set of flow-based network traffic data containing both normal connections and cryptomining attacks. To evaluate the fidelity of the generated data, we analyze both their results in terms of quality measures of statistical nature and also regarding the use of these synthetic data to feed a nested machine learning-based classifier. The experimental results evince a clear outperformance of the GAN network tuned with this new activation function with respect to both a na\"ive mean-based generator and a standard GAN. The quality of the data is so high that the generated data can fully substitute real data for training the nested classifier without a fall in the obtained accuracy. 
This result encourages the use of GANs to produce high-quality synthetic data that are applicable in scenarios in which data privacy must be guaranteed. 

\end{abstract}

\section{Introduction}

In many different domains, the application of machine and deep learning (MDL) techniques requires the availability of considerable amounts of data to take advantage of their powerful learning processes. 
In addition, the applicability of MDL algorithms requires to take into account the evolution of data patterns over time, which implies to produce periodically additional volumes of relevant data for training new MDL models.
Furthermore, producing the required volumes of data in industrial scenarios usually represents a considerable drawback since data gathering and processing tasks tend to be optimized to guarantee services and billing. 
Even if efficient mechanisms for generating and labelling data sets can be implemented, data are increasingly protected by the legal regulations that governments impose to guarantee the privacy of their contents (e.g., European General Data Protection Regulation (GDPR)). These restrictions may discourage the use of real data sets for MDL training and validation purposes.

To address this problem, in the last decade, Generative Adversarial Networks (GANs) \cite{Goodfellow:2014} have gained significant attention due to its ability to estimate the underlying statistical structure of high-dimensional data and generate synthetic data simulating realistic media such as images, text, audio and video \cite{wang2019generative,gao2020generative,jabbar2020survey,pan2019recent}. 
Nowadays, GANs are broadly studied and applied through academic and industrial research in different domains beyond media (e.g., natural language processing, medicine, electronics, networking, and cybersecurity).

Roughly speaking, a GAN model is represented by two independent neural networks, the so-called generator and discriminator, that compete to learn and reproduce the distribution of a real data set. 
After a GAN has been trained, its generator can produce as many synthetic examples as necessary, providing an efficient mechanism for solving the lack of labelled data sets and potential privacy restrictions. 

\subsection*{Problems in GAN generation procedure and related work}

There has been remarkable progress in terms of study of the theoretical properties of GANs, and well as their applications to industrial scenarios. However, less effort has been spent in evaluating GANs quantitatively and qualitatively.
Several measures have been introduced mainly for measuring the image quality,  but there is no consensus yet on which measures best capture the strengths and limitations of GAN models and which of them should be used for fair model comparison. 
Only a discrete number of quantitative criteria have emerged  recently although nearly all of them were designed to measure the quality  of synthetic images \cite{borji2019pros}.

Furthermore, there is no consensus on the deterministic stopping criteria during GAN training that produce high-quality synthetic data.
Several works have already started to benchmark GANs to measure their evolution with respect to the quality of the generated data (for instance, \cite{lucic2017gans,kurach2019large,shmelkov2018good}) and only a few of them have proposed a detention criterium not based on the visual inspection of the synthetic data  (e.g., \cite{cui2021stopping} and \cite{mozo2021synthetic}). 
In fact, the former of these two papers proposes a stopping criterion that is closely related to the signal they generate, so that its direct application in other domains outside signal processing seems not very feasible.

Moreover, in \cite{mozo2021synthetic} we observed that it was not uncommon that GANs fail to discover and replicate with sufficient quality the underlying statistical distribution of real data, in particular, if some feature was a discrete variable (e.g. a categorical feature).
 The recent work \cite{jang2016categorical} proposed a solution  to the generation of categorical variables in the domain of variational autoencoders (VAE) when stochastic neural networks are used, as these variables invalidate them to backpropagate through samples. Authors propose a gradient estimator that replaces the non-differentiable sample from a categorical distribution with a differentiable sample from a Gumbel-Softmax distribution. However, the proposed solution is not general enough since it does not consider continuous variables and only deals with categorical. In addition, they need to hard-code the size of each categorical variable into an array of neurons of size the number of categories of the replicated variable. Some other works in this direction propose to avoid backpropagation through discrete examples \cite{choi2017generating} using and additional pre-trained autoencoder to pass the decoded latent examples to the discriminator.

Finally, it is worth mentioning that the vast majority of the proposed GAN solutions in industrial domains do not apply any systematic measurement procedure for evaluating the GAN evolution during training. This prevents the original data to be replaceable by the generated synthetic data, and forces them to mix real and generated data to achieve acceptable thresholds of quality. Unfortunately, these data augmentation solutions are not applicable in scenarios in which data privacy must be guaranteed as they use a combination of real and synthetic data.
For instance, in the works \cite{Lin2018,navidan2021generative}, a data augmentation method is proposed to generate adversarial attacks against network Intrusion Detection Systems (IDSs). Also, this idea was explored in \cite{Vu2017}, where an augmentation method with GANs was proposed to improve network traffic data sets with highly imbalanced classes.

Nowadays, an increasing number of industrial applications need a mechanism to generate high-quality synthetic data that can fully replace real data in machine learning tasks to avoid privacy violations that might appear when using real data for training and testing purposes.
Therefore, providing a mechanism to generate high-quality labelled data sets that do not incur in privacy breaches will foster cross-development of MDL components by third parties. For example, a telecom provider developing ML-based components to be part of an IDS, can import high-quality synthetic data from a telecom operator to train and validate these ML-based components. As the synthetic data was generated from real data using GANs, the ML component after training will reach the desired level of performance and furthermore, no breach of data privacy will be raised as the telecom operator is not sharing any real data with the telecom provider.

\subsection*{Contribution}

The main contribution of this work is to introduce a novel application of custom activation functions to the last layer of GAN generators to obtain synthetic data that replicates with high fidelity the statistical behaviour of real data. We provide evidence of the robustness of this solution in several scenarios, including when dealing with data that contain discrete features (i.e., variables that follow a discrete distribution). The proposed custom activation function is based on the Smirnov transform (ST) and allows the GAN generator to perfectly mimic any kind of real data distribution.

Roughly speaking, given an initial random variable $Y$ and a target random variable $X$, the Smirnov transformation is a real-valued function $\mathcal{S}_{Y \to X}: \RR \to \RR$ such that the transformed random variable $\mathcal{S}_{Y \to X}(Y)$ distributes as $X$. In other words, $\mathcal{S}_{Y \to X}$ is a function that `bends' the shape of the distribution of the input random variable $Y$ and turns it into the distribution of the output $X$. Notice that the function $\mathcal{S}_{Y \to X}$ is completely deterministic with no stochastic behaviour, and moreover, derivable, which allows it to be seamlessly integrated in the backpropagation computations during the GAN training processes. 

We apply this transformation to the case of the output of a GAN. Empirically, it can be observed that the outputs of a generative network that uses linear activation in the units of the last layer tend to follow a normal-like distribution (e.g.\ unimodal, with non-compact support, approximately symmetric, etc.). 
This behaviour discourages its use when the variables to be replicated are discrete or follow complex continuous distributions. 
To bend this output function to fit the actual distribution to be generated, we propose to use the Smirnov transform $\mathcal{S}_{\mathcal{N} \to X}$ as the activation function in the units of the last layer of the GAN to convert each normal distribution into the objective distribution $X$. This function $\mathcal{S}_{\mathcal{N} \to X}$ can be computed per output unit beforehand just by analyzing the training data set so that it is fixed before starting the training process of the GAN.

To show the effectiveness and utility of this proposal, we have conducted the following analysis and obtained the following conclusions:

\begin{enumerate}
    \item We empirically demonstrate that the data generated with our solution presents a high quality comparable with the original data. This generative power is tested through two different data sets: a) A rendered data set containing a mixture of discrete and continuous variables and b) a real data set of flow-based network traffic data containing benign user connections and cryptomining attacks. 
    \item We provide empirical evidence of the quality improvement of our ST-based GANs with respect to standard GANs and simple mean-based generators. This data fidelity is assessed in two different ways: a) From a statistical perspective, we use two distance functions (based on the $L^1$ distance and Jaccard index) that allow us to quantitatively and graphically compare the evolution of the quality of the data generated by ST and standard GANs during the GAN training; and b) from a practical viewpoint we test the performance of a nested Machine Learning (ML) classifier when the synthetic data generated by ST-based and standard GANs completely substitute real data in the training of the ML classifier, comparing the performance of both ML classifiers when they are tested with real data. 
    \item The results obtained in these tests clearly outperform the existing solutions in the literature. Our proposal generates such a high-quality synthetic samples that can completely replace real data without leading to a fall in the performance of the nested model. To the best of our knowledge, this is the first work in which this goal is achieved outside the image generation domain.
    \item Due to the ill-convergence of the GAN training, we do not use as stopping criterion the usual procedure based on stopping the training after a fixed number of epochs. In substitution, we introduce a novel approach based on the observed performance of a nested ML task that uses the synthetic data produced by the generator at each epoch.
\end{enumerate}

Beyond the high quality of the obtained data itself, this approach has several collateral relevant consequences in GAN generation procedure, among which we can highlight the following:

\begin{enumerate}
    \item Our solution provides a more general approach that deals not only with the replication of categorical variables but with any type of data distribution (continuous or discrete). 
    To convert a standard GAN into a ST-based GAN we only need to change the activation functions of the last layer units of the generator by the corresponding IST functions. 
    The IST activation functions are pre-computed before starting the GAN training and therefore, the additional computational effort can be considered negligible in the context of a full GAN training. 
    In addition, when categorical or discrete variables are considered, it is not necessary to adapt the architecture of the generative network to the structure of the data, as it happens in the existing solutions. On the contrary, we only need to change the activation function of each output neuron in the generator with the corresponding ST activation, independently of the number of categories of the replicated variable.
    
    \item The proposed ST-based activation function is derivable and therefore, it can be seamlessly integrated it in the backpropagation computations during the GAN training processes.
  
    \item The generation procedure avoids the privacy violations that could appear when real data is used for different tasks (e.g., in MDL training and testing processes, and when sharing data in cross-developments or federated environments).
    
    \item The generative power of our solution can be used to alleviate the existing shortage problem of available (labelled) data sets in many domains.    

\end{enumerate}

\section{Brief review of GANs}

Let $\Omega$ be a probability space and let us consider a random vector $X: \Omega \to \RR^n$. Typically, neither the space $\Omega$ nor the variable $X$ are fully known, and only very limited access to them is provided, for instance, through a collection of samples. The aim of a generative model is to `replicate' $X$ as reliably as possible, but admitting certain variability. In other words, the goal is the following: given a fixed probability space $\Lambda$, called the latent space, creating a random vector $G: \Lambda \to \RR^n$ such that $G$ is `near' to $X$ in some sense (typically, in distribution).

The proposal of a Generative Adversarial Network (GAN) is to tune two different functions: a generator $G: \Lambda \to \RR^n$ and a discriminator $D: \RR^n \to \RR$. The generator $G$ will be trained to generate as faithful samples to $X$ as possible, while the discriminator $D$ is a binary classifier optimized for distinguishing between `real' samples of $X$ and `fake' samples generated by $G$. Throughout this paper, we shall adopt the convention that the output of $D$ is the likelihood of a sample $x$ to be real, measured in logits. In other words, if $\sigma(t) = (1+e^{-t})^{-1}$ is the logistic function, then $\sigma(D(x))=1$ means that $D$ believes that $x$ is a real sample whereas $\log(D(x))=0$ stands for $D$ believing that $x$ is a fake sample.

The classification error suffered by the classifier $D$ is thus
$$  
\mathbb{E}_\Omega \left[-D(X)\right] + \mathbb{E}_\Lambda \left[D(G)\right],
$$
where $\mathbb{E}_{\Omega}$ and $\mathbb{E}_{\Lambda}$ denote the mathematical expectation on $\Omega$ and $\Lambda$ respectively. It is customary to weight this error with an increasing concave function $f: \RR \to \RR$ so that we instead consider the error function of $D$
\begin{equation}\label{eq:cost-GAN}
\mathcal{E}(G,D) = \mathbb{E}_\Omega \left[f(-D(X))\right] + \mathbb{E}_\Lambda \left[f(D(G))\right].
\end{equation}
Typical choices for $f$ are $f(s) = \log\left(\sigma(t)\right)$ as in \cite{Goodfellow:2014}, or $f(s) = s$ as in the Wasserstein GAN (WGAN) \cite{Arjovsky-WGAN}.

To improve this error, we will suppose that both $G$ and $D$ depend on some parameters (usually, they are implemented as neural networks) and we shall adjust these parameters to optimize the error $\mathcal{E}$. However, notice that the parameters of the discriminator $D$ are trained to minimize $\mathcal{E}$, whereas the generator $G$ aims to cheat $D$, so it seeks to maximize $\mathcal{E}$. This gives rise to the competitive game
\begin{equation}\label{eq:game-GAN}
    \max_{G}\,\min_{D} \mathcal{E}(G,D) = \max_{G}\,\min_{D}\;\mathbb{E}_\Omega f\left[-D(X)\right] + \mathbb{E}_\Lambda f\left[D(G)\right].
\end{equation}
Beware of the sign conventions. Sometimes in the literature, the error function (\ref{eq:cost-GAN}) is weighted with the decreasing function $f(-s)$, so the resulting cost function is equivalent to the one presented here but the objectives of the functions $G$ are exchanged: $G$ aims to minimize it while $D$ tries to maximize it.

In this way, the goal of the training process of a GAN system is to look for Nash equilibria of (\ref{eq:game-GAN}). These are pairs $(G_0,D_0)$ of a generator and a discriminator such that the function $G \mapsto \mathcal{E}(G, D_0)$ has a local maximum at $G=G_0$ and the function $D \mapsto \mathcal{E}(G_0,D)$ has a local minimum at $D=D_0$. Nash equilibria exhibit very good theoretical properties of probabilistic nature. For instance, in the assumption that a perfect discriminator is reachable at a Nash equilibrium, the Jensen-Shannon divergence between the synthetic distribution $G$ and the original one $X$ is minimized. Similarly, Wasserstein's Earth-moving distance is minimized when a WGAN is applied \cite{Arjovsky-WGAN}.

Despite of that, the problem of finding Nash equilibria in GANs is still essentially open. In the seminar paper \cite{Goodfellow:2014}, it is proposed a simple optimization procedure by alternating gradient descend optimization. However, as pointed out in \cite{Nagarajan-Kolter}, the game (\ref{eq:game-GAN}) to be optimized is not a convex-concave problem, so in general the convergence of these simple training methods is not guaranteed. To stabilize this training process, several heuristic methods have been proposed, such as feature matching, minibatch discrimination, or semi-supervised training \cite{Salimans-Goodfellow:2016}; the introduction of spurious noise \cite{Sonderby:2017, Arjovsky-Bottou:2017} or the application of regularization methods based on gradient penalty \cite{Roth-Lucchi}. 

Finally, it is worth mentioning that most of these training methods have been designed to be applied to the case in which the input data $X$ are graphical images. In this scenario, the statistical properties of the color distribution among the pixels foster the convergence of the GAN, which may achieve high quality results. Nevertheless, when the data to be replicated exhibit different statistical properties (say, categorical features, heavy tailed non-normal distribution, strong domain restrictions...), to the best of our knowledge no general purpose training method is known and the results are typically very poor \cite{mozo2021synthetic}.

\section {The Smirnov transform}

Borrowed from the field of theoretical probability, there is a mathematical transformation that will be very useful for our purposes. In this section, we shall outline the main concepts and properties of this map. For further information, please refer to \cite{ross1976first} or \cite{devroye2006nonuniform}.

Suppose that $X: \Omega \to \RR$ is a random variable and let $F_X(x) = \mathbb{P}(X \leq x)$ its cumulative distribution function. In the case that $F_X: \RR \to [0,1]$ is a continuous increasing function, then its inverse $F_X^{-1}:[0,1] \to \RR$ is a well-defined function called the quantile function. Otherwise, we can still define an analogue of the quantile function by setting
\begin{equation}\label{eq:quantile-function}
    F_X^{-1}(p) = \min_{z} \{F_X(z) \geq p\}.
\end{equation}
Recall that $F_X$ is non-decreasing and right continuous, so the set $F_X^{-1}([p, \infty))$ is actually a final segment including the leftmost endpoint. The value $F_X^{-1}(p)$ is thus nothing but the infimum of this set, which is actually a minimum by right continuity. A key feature of the quantile function is the following result.

\begin{proposition}\label{prop:inv-trans}
Let $U \sim U[0,1]$ be the uniform continuous random variable with support $[0,1]$. Then $F_X^{-1}(U) \sim X$.
\begin{proof}
This is a very well-known result whose proof can be found, for instance, in \cite{devroye2006nonuniform}. We include here a brief proof for the convenience of the reader. Notice that, for any $x \in \RR$ and $0 \leq p \leq 1$, we have that $\min_{z} \{z\,|\,F_X(z) \geq p\} \leq x$ if and only if $x \in \{z\,|\,F_X(z) \geq p\}$, which means that $p \leq F_X(x)$. Therefore, we get that, for all $x \in \RR$
$$
    \mathbb{P}\left(F_X^{-1}(U) \leq x\right) = \mathbb{P}\left(\min_{z} \{F_X(z) \geq U\} \leq x\right) = \mathbb{P}\left(U \leq F_X(x)\right) = F_X(x).
$$
Thus, the cumulative distribution function of the random variable $F_X^{-1}(U)$ coincides with $F_X$, as we wanted to show.
\end{proof}
\end{proposition}

There is also a partial reciprocal result, but it requires to impose extra assumptions on $F_X$.

\begin{proposition}\label{prop:dir-trans}
If $F_X$ is an increasing continuous function, then $F_X(X) \sim U[0,1]$.
\begin{proof}
Again, the proof can be found in \cite{devroye2006nonuniform}. Since $F_X$ is increasing and continuous, its punctual inverse is well defined. Hence, for all $x \in [0,1]$, we have
$$
 \mathbb{P}\left(F_X(X) \leq x\right) = \mathbb{P}\left(X \leq F_X^{-1}(x)\right) = F_X\left(F_X^{-1}(x)\right) = x.
$$
Thus, $\mathbb{P}\left(F_X(X) \leq x\right) = x$ for all $x \in [0,1]$, while $\mathbb{P}\left(F_X(X) \leq x\right) = 0$ for $x < 0$ and $\mathbb{P}\left(F_X(X) \leq x\right) = 1$ for $x > 1$, which shows that $X$ distributes as a uniform random variable.
\end{proof}
\end{proposition}

Some remarks are in order. In the previous proposition, the hypothesis that $F_X$ is invertible is actually too strong. Repeating the type of arguments of Proposition \ref{prop:inv-trans}, it can be shown that $F_X(X) \sim U[0,1]$ provided that $P(X=x) = 0$ for any $x$ in the support of $X$. This holds, for instance, for all continuous random variables. In the case that $F_X$ is discontinuous, a closed formula for the cumulative distribution function of $F_X(X)$ can be still obtained. Indeed, if $x \in \RR$ is a continuity point, then $\mathbb{P}\left(F_X(X) \leq x\right) = x$ as usual; but if $x \in \RR$ lies in the middle of a jump singularity in which $F_X$ jumps to a value $x^+ > x$ (in other words, $x^+ = F_X^{-1}(F_X(x))$), then we have that $\mathbb{P}\left(F_X(X) \leq x\right) = x^+$.

Propositions \ref{prop:inv-trans} and \ref{prop:dir-trans} allow us to transform any random variable into another distribution desired. We state this as a theorem since it is crucial for our later developments. The proof is just a straightforward combination of the aforementioned results.

\begin{theorem}\label{thm:smirnov}
Let $X$ be an arbitrary random variable and let $Y$ be a continuous random variable. Set $F_X$ and $F_Y$ for the cumulative distribution functions of $X$ and $Y$, respectively. Then, we have that
$$
    \mathcal{S}_{Y \to X}(Y) := F_X^{-1}\left(F_Y(Y)\right)
$$
is a random variable that distributes as $X$. The new random variable $\mathcal{S}_{Y \to X}(Y)$ is called the \emph{Smirnov transform of $Y$ into $X$}.
\end{theorem}

Notice that Proposition \ref{prop:inv-trans} does not require any assumption on the distribution of $X$, so the target distribution may be anything. However, to apply Proposition \ref{prop:dir-trans}, we need that the original variable $Y$ is continuous. Recall that, in the case that $F_X$ is not continuous or increasing, the quantile function $F_Y^{-1}$ is defined as in (\ref{eq:quantile-function}).

\subsection{Empirical estimation of the Smirnov transform}

Let us consider the scenario of Theorem \ref{thm:smirnov}, in which we want to transform a random variable $Y$ into another random variable $X$. Typically, the distribution of $Y$ will be known, but the actual distribution of $X$ might be unclear (for instance, because it is a very involved phenomenon).

To address this issue, we propose to estimate it through the so-called empirical cumulative distribution function, denoted by $\hat{F}_X$. To this purpose, we shall have access to a collection of samples $x_1, \ldots, x_m$ of $X$, and we define $\hat{F}_X$ by
$$
\hat{F}_X(x) = \frac{1}{m}\sum_{i=1}^m \chi_{(-\infty, x_i]}(x),
$$
where $\chi_{A}$ is the characteristic function of the set $A$, that is, $\chi_{A}(x) = 1$ if $x \in A$ and $\chi_{A}(x) = 0$ otherwise.

By the Glivenko-Cantelli theorem, when $n \to \infty$, the empirical cumulative distribution function $\hat{F}_X$ converges to the real distribution function $F_X$ in the $L^\infty$ distance almost surely. This means that, for large $n$, $\hat{F}_X$ is a very good estimator of $F_X$. In particular, we can approximate the Smirnov transform $\mathcal{S}_{Y \to X}(Y)$ by the \emph{empirical Smirnov transform}
$$
    \hat{\mathcal{S}}_{Y \to X}(Y) = \hat{F}_X^{-1}\left(F_Y(Y)\right).
$$
Here, $\hat{F}_X^{-1}$ is defined as in (\ref{eq:quantile-function}) by $\hat{F}_X^{-1}(p) = \min_{z} \{\hat{F}_X(z) \geq p\}$.

\subsection{Smirnov transform as GAN activation}

After this probabilistic digression, let us come back to the problem of training GANs. Suppose that we want to generate a random vector $X = (X^1, \ldots, X^n)$ of which only some samples $x_1 = (x_1^1, \ldots, x_1^n), \ldots, x_m = (x_m^1, \ldots, x_m^n) \in \RR^n$ are known. For this purpose, we want to train a GAN made of a generator $G = (G_1, \ldots, G_n): \Lambda \to \RR^n$ and a discriminator $D: \RR^n \to \RR$.

The key problem is that, typically, without further tuning, the output distribution of each of the random variables $G_i: \lambda \to \RR$ is approximately normal. This is related with the mode-collapse problem \cite{thanh2020catastrophic}, a well-reported behaviour of the GANs in which, when they try to generate non-normal variables, the output data tend to not be variate, and the generator degenerates to synthesize only small variations of a prototypical example of $X$. In some sense, the network $G$ stucks in a particular example of $X$ that cheats $D$ very efficiently and ignores any other type of data to generate. This is optimal from the point of view of the GAN game (\ref{eq:game-GAN}), but leads to a degenerate behaviour in which the real distribution of $X$ is not recovered.

To address this problem, in this work we propose to facilitate the job of the generator $G$ by using as activation function a customized function able to capture the statistical subtleties of $X$. To be precise, let us denote by $F_{\mathcal{N}}$ the distribution function of the standard normal distribution, that is
$$
    F_{\mathcal{N}}(x) = \frac{1}{\sqrt{2\pi}}\int_{-\infty}^{\infty} e^{-s^2/2}\,ds. 
$$
Additionally, using the sample $x_1 = (x_1^1, \ldots, x_1^n), \ldots, x_m = (x_m^1, \ldots, x_m^n) \in \RR^n$ of the $n$-dimensional random vector $X = (X^1, \ldots,X^n)$ we generate the $n$ marginal empirical cumulative distribution functions $\hat{F}_{X^1}, \ldots, \hat{F}_{X^n}$. With this information, we create a new activation function as the juxtaposition of the Smirnov transformations from the standard normal distribution to $X_i$
$$
    \hat{\mathcal{S}}_X = \left(\hat{\mathcal{S}}_{\mathcal{N} \to X^1}, \ldots, \hat{\mathcal{S}}_{\mathcal{N} \to X^n}\right) = \left(\hat{F}_{X^1}^{-1}\circ F_\mathcal{N}, \ldots, \hat{F}_{X^n}^{-1}\circ F_\mathcal{N}\right): \RR^n \to \RR^n.
$$
To speed up the evaluation of the function $\hat{\mathcal{S}}_X$, as well as to avoid the vanishing gradient problems derivated from the piecewise constant nature of the functions $\hat{F}_{X^1}^{-1}$, instead of using $\mathcal{S}$ we shall interpolate each of the component functions $\hat{\mathcal{S}}_{\mathcal{N} \to X^i}$ through a numerical interpolation method (typically, spline interpolation \cite{de1978practical}) to get approximate functions $\tilde{\mathcal{S}}_{\mathcal{N} \to X^i}$. Now, the interpolated global function is given by
$$
    \tilde{\mathcal{S}}_X = \left(\tilde{\mathcal{S}}_{\mathcal{N} \to X^1}, \ldots, \tilde{\mathcal{S}}_{\mathcal{N} \to X^n}\right): \RR^n \to \RR^n.
$$
In this way, as the activation function of the output layer of the generator network $G$ we shall use the function $\tilde{\mathcal{S}}_X$. The gradients of the error suffered by this function will be propagated towards the initial layers of the network as usual in the usual backpropagation algorithm. Notice that, provided that the interpolation method returns $C^1$-functions $\tilde{\mathcal{S}}_{\mathcal{N} \to X^i}$ (i.e. derivable with continuous derivatives), then the activation function $\tilde{\mathcal{S}}_X$ is a differentiable map. This is the case, for instance, of spline interpolation, allowing us to deal with discrete distributions even though their underlying quantile function is not smooth.

As a final comment, notice that this Smirnov transformation converts random vectors with normal marginal distributions into random vectors with approximately marginal distribution $X^i$. However, the global dependence between the different output variables is not captured by $\tilde{\mathcal{S}}_X$. Far from being a problem (which may be addressed for example with copulas \cite{nelsen1999introduction}), this is an advantage of this approach: with the use of $\tilde{\mathcal{S}}_X$ as activation function, we ease the job of the generator network of generating the marginal distribution, so that it can focus on the capture of the non-linear interrelations among the different components, which is typically the hardest part to generate.

\section{Performance metrics}
\label{sec:metrics}

We propose to evaluate GANs performance using two different types of metrics. The first set of metrics is inspired by the $L^1$ functional distance and the Jaccard coefficient \cite{tanimoto1958elementary} and aims to quantify the similarity of the synthetic data with respect to the real data from a statistical perspective, considering the joint distribution of data features. 
%
%
On the other hand, the second set of metrics attempts to quantify the performance of synthetic data when it is used as a substitute for real data in the training of a ML classifier that aims to distinguish among the different types of data contained in the data set. To apply this metric, it is obvious to assume that the real data are labelled and that the data set contains more than one type of data.

These two types of metrics will be used to compare the similarity between real and synthetic distributions, and the later set will also be applied to implement a stopping criterion for GAN training that will allow us to select the best generators producing high-quality synthetic data.
It is worth noting that in preliminary experiments we compared the aforementioned metrics with standard Jensen-Shannon and Wasserstein distributional distances. We finally decided to include only the former in our experiments, since the latter sometimes exhibited oscillatory behaviours that were not present in the former. 

\subsection{$L^1$ distance and Jaccard index}
\label{sec:distances}

These two metrics try to measure the difference between the probabilistic distributions of real and synthetic data. They are based on two well-known statistical coefficients applied for hypothesis testing and probabilistic distances: the $L^1$-metric and the standard Jaccard coefficient.

Although both metrics use the probability density function of the two data distributions to compute the distance, they can be straightforwardly extended to a more practical scenario where the density functions of the data distributions are not known. Instead, we shall compute an empirical estimator through the histogram to replace the probability density function.

In the following, we shall sketch briefly the main ideas involved in the construction and estimation of these quality metrics. For further details, please refer to \cite{mozo2021synthetic}.

\paragraph{Empirical probability density function}

Let us suppose that we have samples $x_1, \ldots, x_n \in \RR^d$ of a $d$-dimensional random vector $X$. Let us choose a partition of the support of $X$ into disjoint cubes $C_1, \ldots, C_s$. For simplicity, we shall take all the cubes $C_i$ of the same volume. The empirical probability density function $h_X: \RR^d \to \RR$ is the function
$$
    h_X(x) = \frac{1}{n} \sum_{j=1}^s \left|\{x_i \in C_j\}\right|\chi_{C_j}(x),
$$
where $\chi_{C_j}$ is the characteristic function of the cube $C_j$ (i.e.\ $\chi_{C_j}(x) = 1$ if $x \in C_j$ and is $0$ otherwise) and $\left|\{x_i \in C_j\}\right|$ stands for the number of samples that belong to the cube $C_j$. By the Glivenko-Cantelli theorem \cite{van1996glivenko}, the empirical probability density function $h_X$ is a faithful estimator of the actual probability density function of $X$.

\paragraph{$L^1$ Distance} 
Given two continuous $d$-dimensinal random vectors $X$ and $Y$ with probability density functions $f_X$ and $f_Y$, we can consider the $L^1$ distance between their probability density functions, that is
$$
    d_{L^1}(X,Y) = \int_{\RR^d}|f_X(s) - f_Y(s)|\,ds.
$$
Notice that $d_{L^1}(X,Y) = 0$ if and only if $X=Y$ almost sure.

However, in applications, it is not common to explicitly know the probability density functions of $X$ and $Y$. Instead, from a collection of samples $x_1, \ldots, x_n$ and $y_1, \ldots, y_m$ of $X$ and $Y$, respectively, we can compute their empirical probability density functions $h_X$ and $h_Y$. In this way, the empirical $L^1$ distance can be taken as
$$
    d_{L^1}^{\textrm{emp}}(X,Y) =  \int_{\mathbb{R}^d} |h_X(s) - h_Y(s)|\,ds = L\sum_{j = 1}^{s} |h_X(C_j)-h_Y(C_j)|,
$$
where $L$ is a constant depending only on the volume and number of elements of the partitions taken and $h_X(C_j)$ (resp.\ $h_Y(C_j)$) is the value of the function $h_X$ (resp.\ $h_Y$) on $C_j$. In this empirical setting, we have that $d_{L^1}^{emp}(X,Y) = 0$ if and only if $h_X(C_j) = h_Y(C_j)$, i.e.\ if and only if the number of samples of $X$ on each cube $C_j$ equals the number of samples of $Y$.

\paragraph{Jaccard index} This metric is designed to compare the supports of two distributions. In this way, instead of looking at the particular distribution function, the aim of this metric is to determine whether the two random variables satisfy the same value constraints.

Suppose that we have two random variables $X$ and $Y$ with supports $\textrm{supp}(X)$ and $\textrm{supp}(Y)$, respectively. The Jaccard index of $X$ and $Y$ is 
$$
    J(X,Y) = \frac{|\textrm{supp}(X) \cap \textrm{supp}(Y)|}{|\textrm{supp}(X) \cup \textrm{supp}(Y)|},
$$
where $|A|$ stands for the Lebesgue measure (i.e.\ the volume) of the measurable set $A$. This coefficient takes values in the interval $[0,1]$ and the larger the value of $J(X,Y)$ the more similar the empirical supports.

Again, if the real support is not known, we can still estimate it through the empirical probability density functions as
$$
    J^{smp}(X,Y) = \frac{|\textrm{supp}(h_X) \cap \textrm{supp}(h_Y)|}{|\textrm{supp}(h_X) \cup \textrm{supp}(h_Y)|}.
$$

\subsection{Nested ML performance}\label{sec:nested-ml}
\label{sec:sub:nested_ML}

The second set of metrics attempts to quantify the performance of synthetic data when it is used as a substitute for real data for training a ML classifier.

To be precise, suppose that our data set of real data, let us call it DS, is labelled for a supervised classification ML task. In other words, DS contains instances of $s \geq 2$ different classes which are appropriately labelled. For the sake of notational simplicity, we shall consider the case $s=2$ of binary classification (as appears in the experiments of this work), but the approach can be straightforwardly generalized.

In order to train a ML model, as usual, we can split DS into two data sets, DS1 and DS1, with similar statistical properties. In this way, DS1 can be used for training a ML classifier, whereas DS2 is reserved for testing its accuracy through the standard classification quality measures: $F_1$-score, precision and recall.


Nevertheless, apart from the training the ML classifier, DS1 can also be used to train GANs aiming to replicate its data. Hence, using DS1 we train two GANs $(\Lambda_0, G_{0}, D_0)$ and $(\Lambda_1, G_1, D_1)$ to synthesize data with label $0$ and $1$ respectively. Choose $N, M > 0$ and draw samples $x_1^0, \ldots, x_N^0$ and $x_1^1, \ldots, x_M^1$ of the latent spaces $\Lambda_0$ and $\Lambda_1$ respectively. Then, using the generators $G_0$ and $G_1$, we create a new fully synthetic training data set $\textrm{DS1}'$ with $N + M$ instances joining the synthetic data generated by both GANs.

With this new synthetic dataset $\textrm{DS1}'$, we train a standard ML classifier (say, a random forest classifier \cite{kam1995random}). Then, screening the precision, recall, and $F_1$-score of the classifier against DS2, we are able to measure the quality of the generated data: the higher these measures, the better the synthetic data that was generated by $(\Lambda_0, G_{0}, D_0)$ and $(\Lambda_1, G_1, D_1)$. Hence, large values of these coefficients point out that the synthetic data generated by $G_0$ and $G_1$ can be used to faithfully substitute the real instances. Observe that no real data is used for such training purposes, although real data is always used for testing. 

Additionally, as a baseline comparison for the metrics obtained with GAN synthetic data, we can also consider the ML classifier trained with DS1 and compute its performance metrics with DS2 as the testing data set. In this way, we can compare the performance of the ML classifier trained with GAN synthetic data against the benchmark-level performance obtained using real data during the training of the ML classifier.
Notice that our approach highly differs from many existing works that only mix real with synthetic data (e.g., data augmentation solutions), which can generate data privacy breaches as real data is present in the resultant data set.

\paragraph{Marginal quality evaluation}

From a practical perspective, beyond the aforementioned process, we can also evaluate the marginal quality of each of the generators before jointly evaluating the quality of $G_0$ and $G_1$. We generate only one of the types of data, say label $0$, and we mix the synthetic samples of label $0$ with real samples of label $1$ obtained from DS1. This data set is used to train a ML classifier and then the classifier is tested on DS2 to get the performance measures. This process is repeated for each type of data.
In this way, the corresponding ML accuracy coefficients will only measure the ability of $G_0$ to generate label $0$, regardless of the fitness of $G_1$ and vice versa.

In our experiments, we apply a variant of this approach that computes the performance of the ML classifier at each of the training epochs of the GAN.
In this way, we are able to screen the evolution of the training and to relate it to the quality of the generated data. 
In particular, this idea enables a novel stopping criterion: when the GAN training epochs do not produce any significant enhancement in the performance of the ML classifier, the training process of the GAN is stopped.
It is worth noting that this approach allows to train each type of GAN in parallel and therefore, each training can be stopped at different epochs when no significant enhancement is observed in a particular GAN.

\paragraph{Choice of the best GAN model}

After each GAN is trained, the joint performance of both types of synthetic data is computed.
Using the whole set of GANs obtained during the marginal quality evaluation would imply to compute the ML performance for each pair $(G_0, G_1)$ of generators $G_0$ and $G_1$ at each of their training epochs. This leads to a quadratic number of generators to be tested in the ML task, both for training and testing, to obtain the full set of measurements. 

However, we  observed experimentally that drawing roughly a dozen samples by choosing uniformly at random one generator of each type of data tends to produce results equivalent to the brute force approach of trying all possible combinations. 
In addition, we applied more elaborated strategies based on performance elitism, ordering the generators of each type of traffic by $F_1$-score and choosing generators at random only from the subset containing the best generators.

Finally, we would like to remark that, since we have generative models able to create as many samples as needed, we can choose the number of generated samples $N$ and $M$ as large as desired. If we choose $N$ and $M$ in the same range as the number of instances in the original dataset DS1, we obtain a synthetic data set with very similar characteristics to the original one. In particular, any unbalancing between classes will remain. However, other choices can be made. For instance, we can decide to take $N=M$, so that the obtained data set corrects the unbalanced situation, or to take $N$ and $M$ much larger than the size of the original dataset, so that we increment the amount of data available for the ML classifier. It is worthy to mention that, even though this solution gives rise to a balanced dataset as with data augmentation procedures, the proposed solution is substantially stronger than simple data augmentation: the synthetic data is not a simple enrichment of the original data set but a completely new data set.

\section{Experiments}

\subsection{Dataset description}\label{sec:dataset-description}

To demonstrate the applicability of the proposed solution, we have defined two  different use cases to experimentally validate the versatility of ST-based GANs.
The general objective of the two use cases is to test whether GAN-generated synthetic traffic can fully replace real traffic in ML problems where the use of real data could generate privacy breaches.

\paragraph{First use case: Rendered data set}

For the first use case considered in this work, we have designed two fictitious data sets (denoted by DS1-r and DS2-r). As previously commented in subsection \ref{sec:sub:nested_ML}, we train GANs with DS1-r, while DS2-r is used   for training the ML-classifier that will assess the quality of the synthetic data generated by the GANs. In addition, DS1-r is used for training the benchmark ML classifier.
Both data sets contain entries of two different data types (i.e., 2 labels) and are perfectly balanced, containing $400,000$ entries for each label. The data sets are composed of four variables, two of them contain continuous values and the other two contain discrete values. Each of the entries of the data set was generated by sampling from two random vectors (one per label class) made of independent random variables with the distributions shown in Table \ref{tab:distribution-rendered} (see also their histogram in Figures \ref{fig:SYN-distr-l0} and \ref{fig:SYN-distr-l1}.

\begin{table}[ht!]
    \centering\scriptsize
    \begin{tabular}{|c|c|c|c|c|}
        \hline\textbf{Label} & \textbf{Feature 0} & \textbf{Feature 1} & \textbf{Feature 2} & \textbf{Feature 3} \\\hline\hline
        Label $0$ & $\begin{matrix} \textrm{Normal} \\ (\mu = 0, \sigma = 1)\end{matrix}$ & $\begin{matrix} \textrm{Binomial} \\ (n = 15, p = 0.3)\end{matrix}$ & $\begin{matrix}\textrm{Exponential} \\ (\sigma = 3)\end{matrix}$ & $\begin{matrix}\textrm{Poisson} \\ (\lambda = 1.0)\end{matrix}$ \\\hline
        Label $1$ & $\begin{matrix} \textrm{Normal} \\ (\mu = 0, \sigma = 1)\end{matrix}$ & $\begin{matrix} \textrm{Discrete uniform} \\ (\textrm{supp.\ } [0, 15])\end{matrix}$ & $\begin{matrix}\textrm{Snedecor }F \\(\nu_1 = \nu_2 = 3)\end{matrix}$ & $\begin{matrix}\textrm{Poisson} \\ (\lambda = 2.0)\end{matrix}$\\\hline
    \end{tabular}
    \caption{Distributions used to render data sets DS1-r and DS2-r. Each of the features was drawn independently.}
    \label{tab:distribution-rendered}
\end{table}

This use case aims to demonstrate that the use of linear activation functions at the last layer of the generator fails to replicate complex data distributions such as the ones we have rendered and in particular, the variables representing discrete data distributions. On the contrary, we show that, using ST-based activation functions, we are able to perfectly replicate from a statistical point of view, both continuous and discrete variables, even if the data variables follow complex statistical distributions.

It is worth noting that the statistical distributions of the two types of data have been generated in such a way that they are similar on average, which makes the task of an ML classifier more complicated when we want to train it to correctly identify the two types of data. Indeed, if the synthetic data have not been generated with sufficient fidelity to the two real data distributions, because the means of their 4 variables are so close, the ML classifier trained with synthetic data will obtain a significantly worse performance in terms of accuracy, precision and recall than a benchmark classifier trained with real data. 

Finally, observe that some of the distributions of the 8 variables (4 per data type) have been generated with statistical patterns different from the Gaussian distribution (Table \ref{tab:distribution-rendered}) to demonstrate that the generators with linear activation do not replicate with precision the real distributions when they are not Gaussian or discrete, and that on the contrary, when the generators have activation functions based on ST, the distributions of the synthetic data exactly replicate the real variables even if their distributions follow statistical patterns very different from Gaussian distributions (e.g. discrete distribution. 

\paragraph{Second use case: Network traffic}

The second use case aims to evaluate the replication by GANs of data coming from a real scenario in the cybersecurity domain. The real data used in this experiment were previously generated in a realistic network laboratory called the Mouseworld lab \cite{pastor2018mouseworld}. The Mouseworld is a network digital twin created at Telef\'onica R+D facilities that allows deploying complex network scenarios in a controlled way.
In this lab, a set of virtual machines were deployed for the generation of regular network traffic (e.g., web and video flows) jointly with cryptomining clients connected to public mining pools in the Internet \cite{pastor2020detection}. 

We ran the experiment twice for one hour, collected the transmitted packets, and generated two data sets (denoted by DS1-c and DS2-c) each with 4 millions of flow-based entries  containing statistics of the connections.
Normal traffic connections were labelled with $0$ and cryptomining ones with $1$. 
It is worth noting that both data sets are totally unbalanced, containing only $4,000$ entries of criptomining connections. 

A set of 59 statistical features were extracted from each TCP connection, although we selected a reduced set of 4 for our experiments: (a) number of bytes sent from the client, (b) average round-trip time observed from the server, (c) outbound bytes per packet, and (d) ratio of packets inbound / packets outbound. 
These four features were selected as they exhibit two interesting properties for our generative experiments that were previously commented in the first use case: (i) each feature presents a different statistical behaviour far from a Gaussian distribution and (ii) the mean of each feature in the two types of traffic (normal and cryptomining) were close, which makes the task of an ML classifier more complicated when we want to train it to correctly identify the two types of data.

The nature of both types of traffic is very different, a fact that will be reflected in the quality of the GAN-generated data. The normal traffic has a great variety since it is composed of many types of connections (e.g., video, audio, web elements, and multimedia elements). On the contrary, the cryptomining connections are handled by a reduced set of protocols and therefore, their statistical patterns are not expected to differ substantially. Due to the greater diversity that normal traffic connections exhibit when compared to cryptomining connections, GANs that try to replicate label-$0$ data will perform slightly worse than their counterparts that replicate label-$1$ data.

\subsection{Proposed architecture}

Aiming to mimic synthetic data with several types of behaviour,  we adopted in preliminary experiments a well-known conditional GAN model, the so-called Auxiliary Classifier GAN (AC-GAN) \cite{odena2017conditional}, as the architecture to generate at the same time all types of variables. In both use cases, the ACGAN did not produce an adequate performance when replicating the two types of data and moreover, it generated significant oscillations in the convergence process. For that reason, we opted to use a different GAN for each type of data to be generated.

To get rid of the mode collapse problems that frequently appear during GAN training, we adopted as a reference model the WGAN architecture \cite{Arjovsky-WGAN}, in which a  Wasserstein loss function is used as the loss function instead of a standard cross-entropy function. We tested two different strategies to enforce the required Lipschitz constraint in the cost function, weight clipping (\cite{Arjovsky-WGAN}) and gradient penalty (\cite{gulrajani2017improved}), not observing any significant enhancement in the convergence of the GAN training. Therefore, we chose a WGAN architecture with no additional strategy to enforce the Lipschitz constraint and with a discriminator with small learning rates as heuristic to avoid reaching mode collapse situations.

Regarding that the statistical nature of the 4 features to be synthetically replicated in both use cases did not exhibit any topological structure or time relationship among them, convolutional or recurrent networks would not take any advantage of it. Therefore, we selected fully connected neural networks (FCNNs) as the architectural model for both the discriminant and generator networks.
LeakyRelu functions were used as activation layer in all layers, except for the last layer of the generator. 
Based on previous experiments \cite{mozo2021synthetic}, no filtering based on the output of the discriminator was applied to discard synthetic data, nor  was noise added to synthetic or real data during the training of the discriminator to help GAN convergence.

\begin{table*}[!t]
\centering
\caption{WGAN hyperparameters.}
\label{table:hyperparameters}
\resizebox{1.\linewidth}{!}{%
\begin{tabular}{|>{\hspace{0pt}}m{0.0027\linewidth}>{\hspace{0pt}}m{0.304\linewidth}|>{\hspace{0pt}}m{0.252\linewidth}|>{\hspace{0pt}}m{0.284\linewidth}|>{\hspace{0pt}}m{0.284\linewidth}|>{\hspace{0pt}}m{0.344\linewidth}|}
\cline{4-6}  \multicolumn{1}{c}{} & \multicolumn{1}{>{\hspace{0pt}}m{0.304\linewidth}}{} &  & \textbf{Range of values} & \textbf{Rendered data} & \textbf{Network data} \\ 
\hline

 & \multirow{ 12}{*}{\LARGE \textbf{Generator}}  & \textit{\# layers} \par{} \textit{\# units per layer} & {[}2..6] \par{} {[}100..10000] & $G_0, G_1 \in [500,3000,5000,400,4]$  & $G_0 \in [200,500,3000,500,4]$ \par{}  $G_1 \in [600,3000,1000,4]$ \\ 
 \cline{3-6}
 &  & latent vector & Fixed value \par{} (100, 123) & $100$ & $123$ \\ 
\cline{3-6}
 &  & noise for latent vector \par{} (distribution, std) & distr = {[}norm, uniform] \par{} std = [0.1..100] & uniform, std = 1.5 & normal, std = 0.5\\ 
\cline{3-6}
 &  & batch normalization & {[}True..False] & True & True \\ 
\cline{3-6}
 &  & regularization: \par{} L2, dropout & L2 = $[1e-5..10]$ \par{} Dropout = $[0..1]$
 & L2 \par{} $G_0,G_1=0.1$ \par{} Dropout \par{} $G_0,G_1 = 0$
 & L2 \par{} $G_0,G_1=0$ \par{}  Dropout\par{} $G_0,G_1 = 0$  \\ 
\cline{3-6}
 
 &  & LeakyRelu alpha & Fixed value $(0.15)$ & $G_0=G_1= 0.2$ \par{} $D_0=D_1=0.15$ & $G_0=G_1= 0.2$ \par{} $D_0=D_1=0.15$ \\ 
\hline\hline
&  & learning rate & Default value \par{} $(0.001)$ & $G_0, G_1= 0.001$  & $G_0, G_1= 0.001$ \\ 
\cline{3-6}

&  \multirow{ 7}{*}{\LARGE \textbf{Discriminator}}   &  \textit{\# layers} \par{} \textit{\# units per layer} & {[}2..6] \par{} {[}100..10000] &  $D_0, D_1 \in [280,503,177,23]$ &  $D_0 \in [380,800,600,177,23]$ \par{}  $D_1\in[280,903,500,23]$\\ 
\cline{3-6}
 &   & batch normalization & {[}True..False] & True & True \\ 
\cline{3-6}
  &  & regularization: \par{} L2, dropout & L2 = $[1e-5..10]$ \par{} Dropout = $[0..1]$
 & L2  \par{} $D_0,D_1=0.001$ \par{} Dropout \par{} $D_0,D_1=0$
 & L2 \par{}  $D_0=0.02$, $D_1=0.05$ \par{} Dropout\par{} $D_0=0.1$, $D_1=0.15$ \\ 
\cline{3-6}
 &  & LeakyRelu alpha & Fixed value $(0.15)$ &  $D_0=D_1=0.15$ & $D_0=D_1=0.15$ \\ 
\cline{3-6}
 &  & learning rate & Default value \par{} $(0.001)$ & $D_0, D_1= 0.0001$ &  $D_0, D_1= 0.001$\\ 
\hline
\end{tabular}
}
\end{table*}

\subsection{Experimental results}

In this section, we analyze the performance on each of the use cases of Section \ref{sec:dataset-description} of four different generative approaches: (i) when real data is used and no generation occurs, (ii) with a simple mean-based generator, (iii) with a standard WGAN, and (iv) with a WGAN with ST-based activation function. As we will show, the ST-based solution outperforms both the standard GAN and the simple mean-based generator, reaching an accuracy in a nested ML classifier similar to the one obtained with real data.

As ML classifier, we selected a Random Forest model with 300 estimators. In the first use case, we limited each tree depth to 20 to avoid some overfitting effects that appeared in preliminary experiments. No depth limit was applied to trees in the second use case.
In the first use case, a balanced set of samples were obtained for each label ($200,000$ samples per label totalling $400,000$ samples) for training and testing. In the second use case, we kept the original ratio of the two labels (many more normal traffic connections than cryptomining ones)  and we got $400,000$ samples of label $0$ (normal traffic) and only $4,000$ of label $1$ (cryptomining connections) for training and testing. In the testing process, we establish additional decision thresholds of $0.2$, $0.4$, $0.6$ and $0.8$  to the default $0.5$ in order to analyse the results of the default and the best performing threshold. Finally, training and testing of ML classifiers were run 100 times in all experiments to minimise biased behaviours during sampling and training.

For training WGANs, we used previous knowledge from past experiments \cite{mozo2021synthetic},  and applied the set of hyperparameters detailed in Table \ref{table:hyperparameters} performing a blind random search in the hyperparameter space guided by the $F_1$-score obtained in a nested ML-model that was executed  evaluating the marginal quality of the generator after $10$ mini-batch trains (see Section \ref{sec:sub:nested_ML}).
For each type of data, the WGAN selected was the one that obtained the best $F_1$-score for the nested classifier in any of its mini-batches.
As optimization algorithms, we used Adam for generators and RMSProp for discriminators and the  binary cross-entropy loss function was substituted by the Wasserstein loss. The hyper-parameters chosen for the generator and discriminator in each use case are detailed in the last two columns in Table \ref{table:hyperparameters}.

\subsubsection{Real data}
To establish an upper bound on the expected performance of the nested ML classifier, a benchmark classifier was trained $100$ times for each of the two use cases with real samples from the first data set DS1 and tested using samples from the second data set DS2.
The first row in Table \ref{tab:SYN_mix} summarises the obtained $F_1$-score values and confusion matrices in testing for the best decision threshold and the default threshold ($0.5$) in the first use case (rendered data). The first row of Table \ref{tab:CR_mix} summarises the same information for the second use case (cryptomining attack).

Alongside, Figure \ref{fig:Mix_SYN_B_R} plots the histogram of the statistical distribution of $F_1$-score results obtained in the first use case  when a ML classifier was trained with DS1-r and tested with DS2-r. %
Similarly, Figure \ref{fig:Mix_CR_B_R} shows  the histogram of the statistical distribution of  $F_1$-score results obtained in the second use case  when a ML classifier was trained with DS1-c and tested with DS2-c.

\subsubsection{Mean generator}

Opposed to the previous results, we established a baseline in our experiments through a na\"ive generator designed to generate new data by adding Gaussian noise to the means of the data features.
The standard deviation of the noise was manually adjusted to produce the best results and hence, a more challenging baseline. Each n\"iaive model was tested with the corresponding DS2 data set of each use case.

\begin{figure*}[!ht]
\begin{mdframed}
\centering

\begin{subfigure}[t]{0.99\textwidth}
\centerline{
  \includegraphics[width=0.49\linewidth]{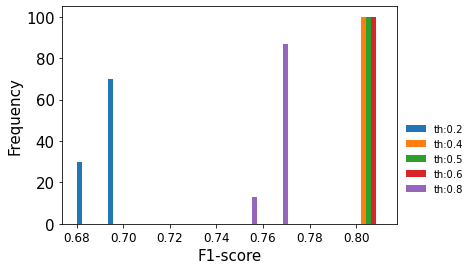} 
  \includegraphics[width=0.49\linewidth]{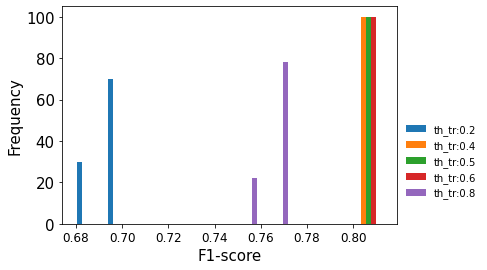}}
\caption{Real data.}\label{fig:Mix_SYN_B_R}
\end{subfigure}
\medskip

\begin{subfigure}[t]{.99\textwidth}
\centering
\vspace{0pt}
\includegraphics[width=0.49\linewidth]{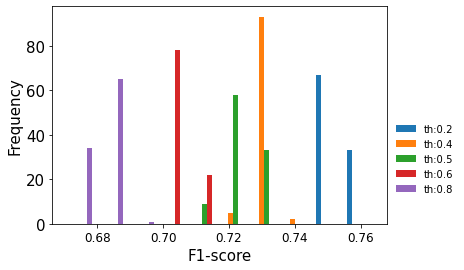}
\hfill
\includegraphics[width=0.49\linewidth]{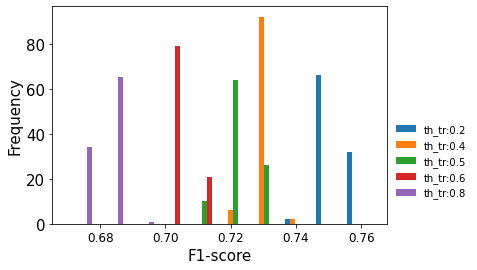}
\caption{Naive generator with means.}\label{fig:Mix_SYN_B_M}
\end{subfigure}

\end{mdframed}
\medskip
\begin{minipage}[t]{.99\textwidth}
\caption{Rendered data set (first use case). Baseline. $F_1$-score on DS2-r (left) and DS1-r (right) using  as training data set: (\ref{fig:Mix_SYN_B_R}) a real data set (DS1-r)  and  (\ref{fig:Mix_SYN_B_M}) a naive noise generator with means. Results for decision thresholds of 0.2, 0.4, 0.5, 0.6 and 0.8 are represented. Each experiment was run 100 times.} \label{fig:Mix_SYN_B}
\end{minipage}

\end{figure*}

\begin{figure*}[!ht]
\begin{mdframed}
\centering

\begin{subfigure}[t]{0.99\textwidth}
\centerline{
  \includegraphics[width=0.49\linewidth]{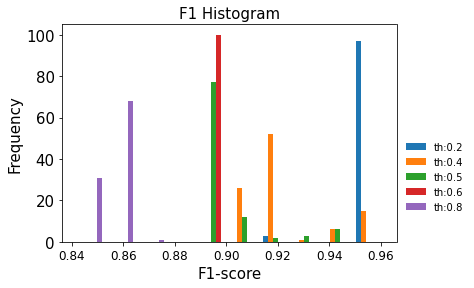} 
  \includegraphics[width=0.49\linewidth]{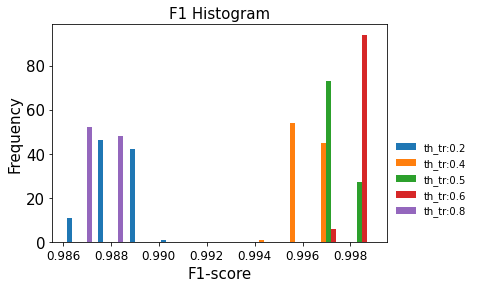}}
\caption{Real data.}\label{fig:Mix_CR_B_R}
\end{subfigure}
\medskip

\begin{subfigure}[t]{.99\textwidth}
\centering
\vspace{0pt}
\includegraphics[width=0.49\linewidth]{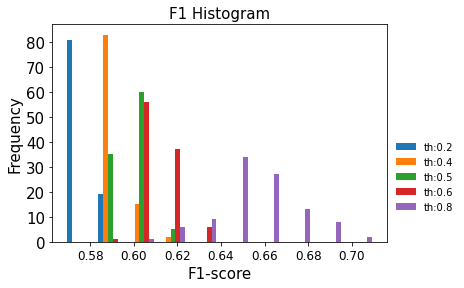}
\hfill
\includegraphics[width=0.49\linewidth]{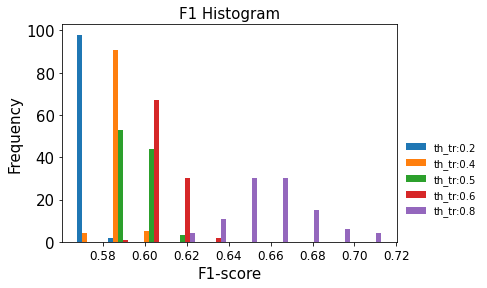}
\caption{Naive generator with means.}\label{fig:Mix_CR_B_M}
\end{subfigure}

\end{mdframed}
\medskip
\begin{minipage}[t]{.99\textwidth}
\caption{Cryptomining attack data set (second use case). Baseline.
$F_1$-score on DS2-c (left) and DS1-c (right) using  as training data set: (\ref{fig:Mix_CR_B_R}) a real data set (DS1-r)  and  (\ref{fig:Mix_CR_B_M}) a naive noise generator with means. Results for decision thresholds of 0.2, 0.4, 0.5, 0.6 and 0.8 are represented.  Each experiment was run 100 times.} \label{fig:Mix_CR_B}
\end{minipage}

\end{figure*}

A summary of the baseline results obtained with this na\"ive model can be found in the second rows of Table \ref{tab:SYN_mix} and Table \ref{tab:CR_mix}, for the first and second use cases respectively. Histograms showing the $F_1$-score values obtained after running the mean generators $100$ times  are shown for each use case in Figure \ref{fig:Mix_SYN_B_M} and Figure \ref{fig:Mix_CR_B_M}.

\subsubsection{GANs: Linear and custom activation}

For the two use cases of Section \ref{sec:dataset-description}, we ran a set of experiments to obtain the performance of a ML classifier trained with GAN synthetic data and tested with DS2 data sets (DS2-r for the first use case and DS2-c for the second). GANs were always trained using DS1 data sets (DS1-r for the first use case and DS1-c for the second). The GANs were trained during a fixed set of $25,000$ mini-batches (50 epochs). Every 10 mini-batches, the GAN generator model was saved for posterior use, and the obtained $L^1$ and Jaccard distances of synthetic and real data were computed. 
In addition, as mentioned in Section \ref{sec:nested-ml}, a training data set was generated mixing GAN synthetic data of the current label with real data sampled from the other label of DS1. Using this hybrid data set, we trained a ML classifier, and then the ML model was tested with DS2. The obtained $F_1$-score represents the marginal performance of the GAN synthetic data for the current label and, in addition, provides a potential early stopping criterion for GAN training. Note that although both DS1 and DS2 contain real data, the performance against DS2 provides a more reliable measure, since DS1 was used for training the GAN and therefore, the GAN generator could have learnt specific information only contained in DS1. 

After training standard and ST-based WGANs for the two labels, we run the following experiment for each type of GAN (standard and ST-based) to highlight the advantage of the proposed ST solution: 
For each label, a WGAN generator is selected uniformly at random among all models stored previously every 10 mini-batches. Then, a completely synthetic data set is produced using the generator of label $0$ and the generator of label $1$. Using this synthetic data set we train the ML classifier and test its performance with DS2 obtaining the $F_1$-score value and the confusion matrix. 
This process was run $100$ times to compare the statistical distribution of the obtained $F_1$-score values for the standard and ST-based WGANs.
In addition, we repeated the experiment not selecting uniformly at random each label generator among the whole set of stored generators for a label but among the top 10 sorted by the marginal $F_1$-score for this label (i.e., using $F_1$-score elitism). 
In this way, we explored whether it is more efficient to search the best performing label 0 and 1 generators among all stored generators or using the $F_1$-score elitism. In addition, we analyse whether the ST-based solution performs better than the standard WGAN using this elitism.


\paragraph{Distances from synthetic to real data}

\begin{figure*}[!t]
\begin{subfigure}[t]{.245\textwidth}
\centering
\includegraphics[width=1\linewidth]{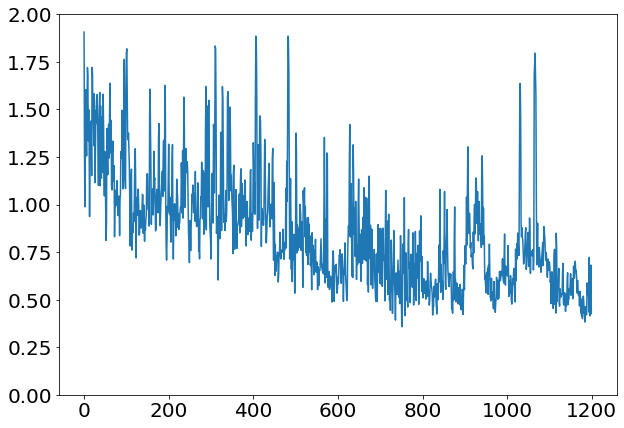} 
\caption{$L^1$ distance. L0.}
\label{fig:syn-vainilla-kolmo-l0}
\end{subfigure}
\begin{subfigure}[t]{.245\textwidth}
\centering
\includegraphics[width=1\linewidth]{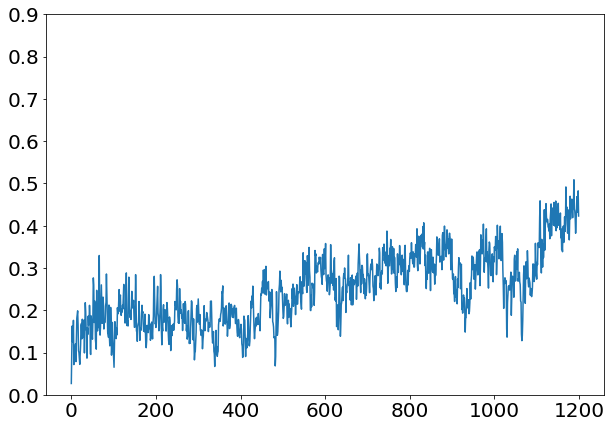} 
\caption{Jaccard index. L0.}
\label{fig:syn-vainilla-jacc-l0}
\end{subfigure}
\begin{subfigure}[t]{.245\textwidth}
\centering
\includegraphics[width=1\linewidth]{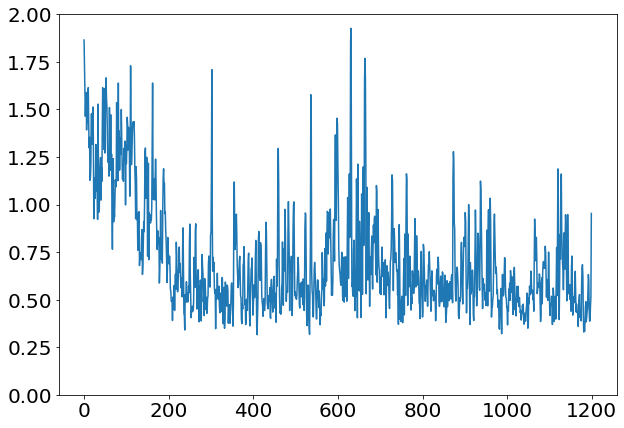} 
\caption{$L^1$ distance. L1.}
\label{fig:syn-vainilla-kolmo-l1}
\end{subfigure}
\begin{subfigure}[t]{.245\textwidth}
\centering
\includegraphics[width=1\linewidth]{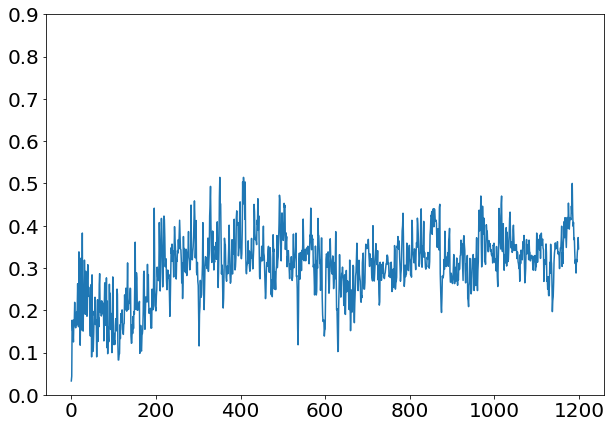} 
\caption{Jaccard index. L1.}
\label{fig:syn-vainilla-jacc-l1}
\end{subfigure}

\medskip
\begin{subfigure}[t]{.245\textwidth}
\centering
\includegraphics[width=1\linewidth]{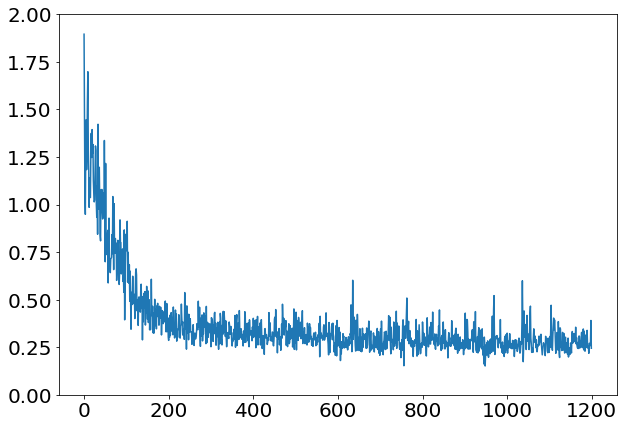} 
\caption{$L^1$ distance. L0.}
\label{fig:syn-custom-kolmo-l0}
\end{subfigure}
\begin{subfigure}[t]{.245\textwidth}
\centering
\includegraphics[width=1\linewidth]{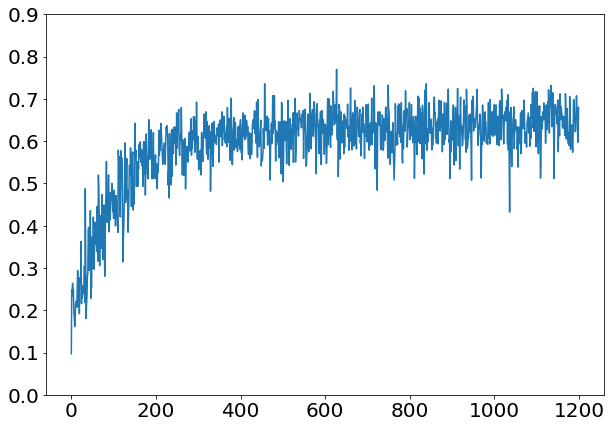} 
\caption{Jaccard index. L0.}
\label{fig:syn-custom-jacc-l0}
\end{subfigure}
\begin{subfigure}[t]{.245\textwidth}
\centering
\includegraphics[width=1\linewidth]{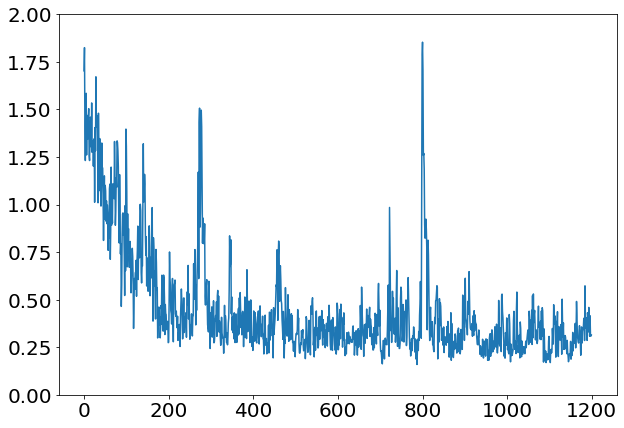} 
\caption{$L^1$ distance. L1.}
\label{fig:syn-custom-kolmo-l1}
\end{subfigure}
\begin{subfigure}[t]{.245\textwidth}
\centering
\includegraphics[width=1\linewidth]{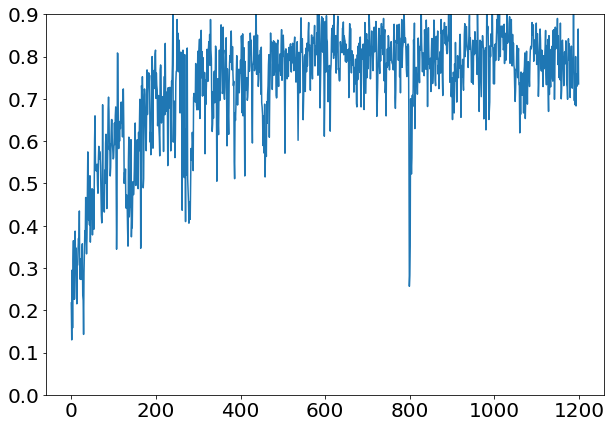} 
\caption{Jaccard index. L1.}
\label{fig:syn-custom-jacc-l1. Label 1.}
\end{subfigure}

\caption{Rendered data (use case \#1). Evolution of  $L^1$ distance and Jaccard index,  using GAN generators with linear activation (top row) and ST-based activation (bottom row) for labels 0 and 1. The $x$-axis represents the GAN training epochs (1 epoch $=$ 50 ticks).}
\label{fig:SYN_distancias}
\end{figure*}

Figure \ref{fig:SYN_distancias} shows the evolution of $L_1$ distance and Jaccard index during the GAN training for label $0$ and label $1$ in the first use case. It can be observed from both labels that the $L_1$ distance curve for the ST-based WGAN stabilises faster, shows less oscillations and achieves smaller values (around $0.3$ for both labels when GAN  training is stabilised) than in the standard WGAN with linear activation (around $0.6$ for both labels in the minimum points of the curves).
With respect to the Jaccard index, the results of the ST-based WGAN conclude in a similar way: The curves for both labels achieve high values (around $0.7$ for label $0$ and $0.9$ for label $1$), stabilise faster (in $4$ epochs for label $0$ and $6$ epochs for label $1$) and do not exhibit  significant oscillations. On the contrary, the Jaccard curves of the standard WGAN for the two labels show a bad performance with values not greater than $0.4$.
These results highlight that in the first use case the similarity of the synthetic data generated by the ST-based WGAN generator and the real data is much higher than when the synthetic data is generated by the standard WGAN generator. 

\begin{figure*}[!t]
\begin{subfigure}[t]{.245\textwidth}
\centering
\includegraphics[width=1\linewidth]{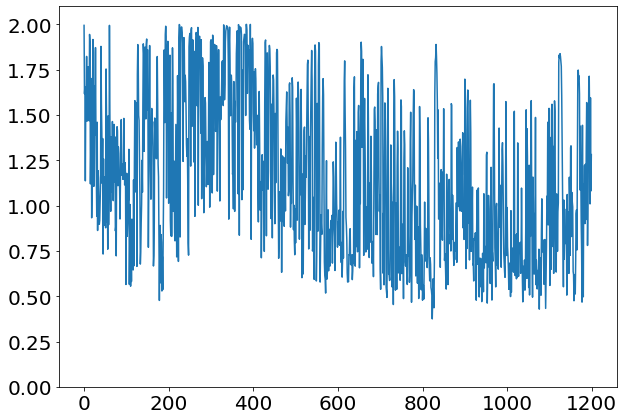} 
\caption{$L^1$ distance. L0.}
\label{fig:vainilla-kolmo-l0}
\end{subfigure}
\begin{subfigure}[t]{.245\textwidth}
\centering
\includegraphics[width=1\linewidth]{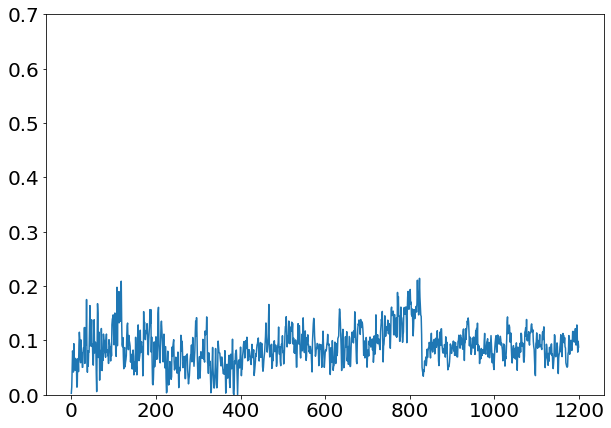} 
\caption{Jaccard index. L0.}
\label{fig:vainilla-jacc-l0}
\end{subfigure}
\begin{subfigure}[t]{.245\textwidth}
\centering
\includegraphics[width=1\linewidth]{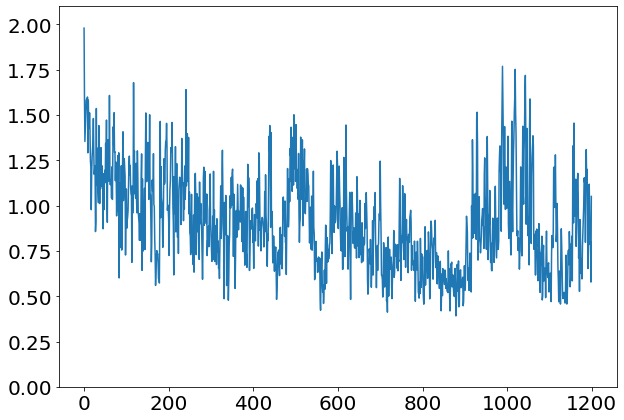} 
\caption{$L^1$ distance. L1.}
\label{fig:vainilla-kolmo-l1}
\end{subfigure}
\begin{subfigure}[t]{.245\textwidth}
\centering
\includegraphics[width=1\linewidth]{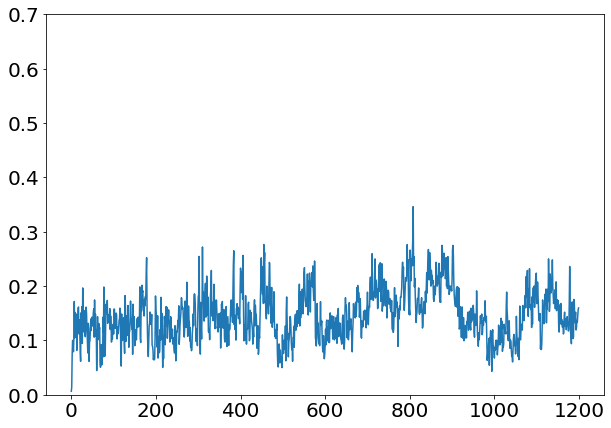} 
\caption{Jaccard index. L1.}
\label{fig:vainilla-jacc-l1}
\end{subfigure}

\medskip
\begin{subfigure}[t]{.245\textwidth}
\centering
\includegraphics[width=1\linewidth]{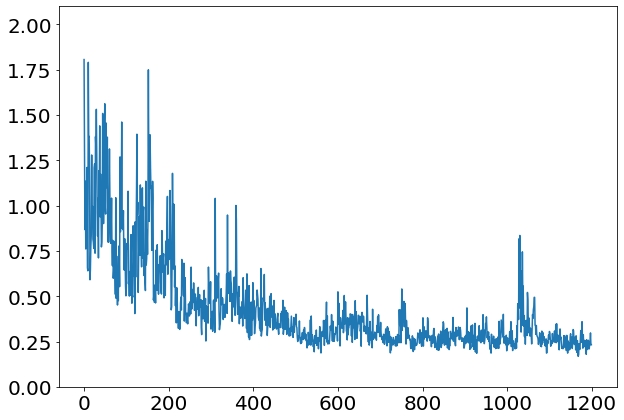} 
\caption{$L^1$ distance. L0.}
\label{fig:custom-kolmo-l0}
\end{subfigure}
\begin{subfigure}[t]{.245\textwidth}
\centering
\includegraphics[width=1\linewidth]{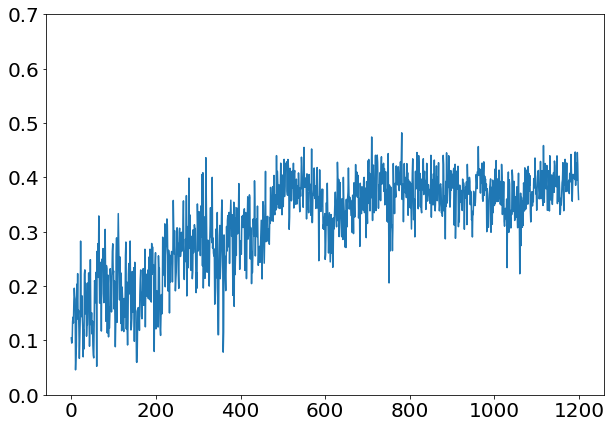} 
\caption{Jaccard index. L0.}
\label{fig:custom-jacc-l0}
\end{subfigure}
\begin{subfigure}[t]{.245\textwidth}
\centering
\includegraphics[width=1\linewidth]{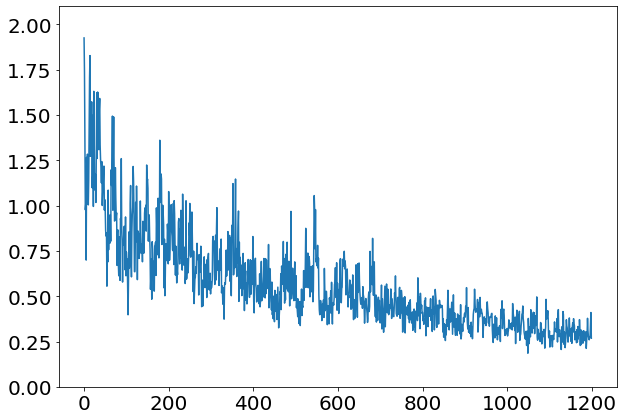} 
\caption{$L^1$ distance. L1.}
\label{fig:custom-kolmo-l1}
\end{subfigure}
\begin{subfigure}[t]{.245\textwidth}
\centering
\includegraphics[width=1\linewidth]{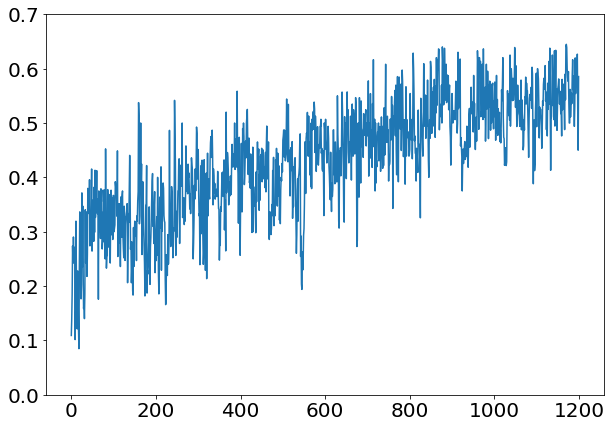} 
\caption{Jaccard index. L1.}
\label{fig:custom-jacc-l1. Label 1.}
\end{subfigure}

\caption{Cryptomining attack scenario (use case \#2). Evolution of  $L^1$ distance and Jaccard index,  using GAN generators with linear activation (top row) and ST-based activation (bottom row) for labels 0 (normal traffic) and 1 (cryptomining connections). The $x$-axis represents the GAN training epochs (1 epoch $=$ 50 ticks).}
\label{fig:CR_distancias}
\end{figure*}

The $L^1$ distance and Jaccard index results for the second use case are shown in Figure \ref{fig:CR_distancias}. 
Similarly to the first use case, the distance in ST-based WGAN stabilises quickly ($10$ epochs for label $0$ and $20$ for label $1$), with a very low value (around $0.25$) and without significant oscillations indicating that the quality of the generated synthetic data is high. On the contrary, standard WGAN suffers from remarkable oscillations and the distance value is not small (from $0.65$ to $1.6$ in label $0$ and from $0.6$ to $1.25$ for label $1$), which highlights that the similarity of the synthetic data generated by the GAN and the real data is not very high.
Analyzing the Jaccard coefficient in the figure for the ST-based WGAN curve, it can be seen that values around $0.4$ and $0.6$ are obtained for labels $0$ and $1$. It is intuited in the figure that with more training epochs the former values would continue to grow. In contrast, the standard WGAN curve quickly stabilises around a very small value of $0.1$ for both labels, which shows that in this case, the statistical distributions of real and synthetic data are quite different.

We can conclude that in both use cases, the ST-based WGAN replicates for the two labels the statistical behaviour of the real data (DS1 data set) with high precision and requiring only a few training epochs. In addition, the quality of the synthetic data produced throughout the training process does not suffer from significant oscillations. 
On the contrary, the standard WGAN replicates with worse quality the statistical distribution of the real data, needs more training epochs, and the quality of the generated data suffers from high oscillations during the training process, which prevents its use in real applications.

\paragraph{Evolution of synthetic data quality}

\begin{figure*}[ht!]
\centering
\centering
\begin{subfigure}[t]{.32\textwidth}
\includegraphics[width=1\linewidth]{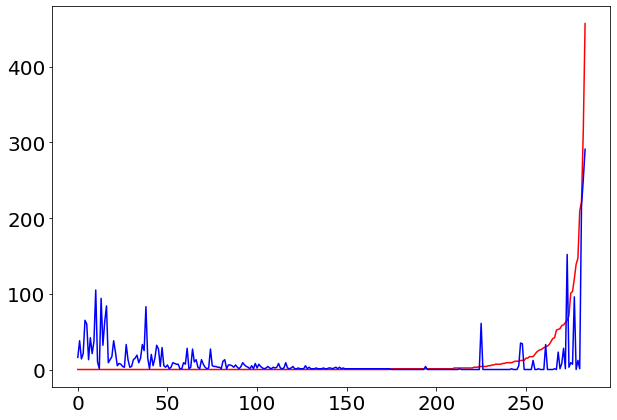} 
\caption{Std WGAN, L0, Epoch 0}\label{fig:E_1_SYN_et0_FA_L}
\end{subfigure}
%
%
\begin{subfigure}[t]{.32\textwidth}
\includegraphics[width=1\linewidth]{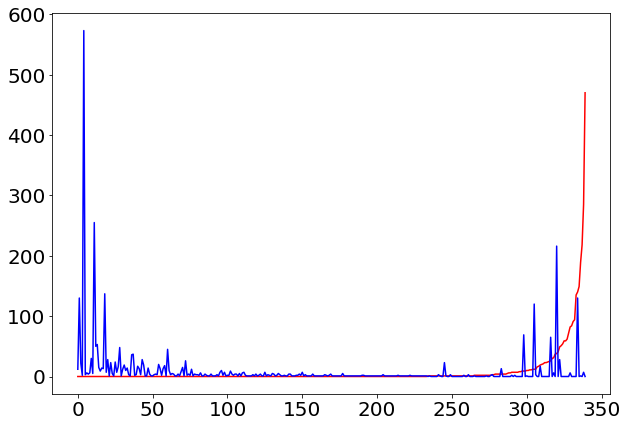} 
\caption{Std WGAN, L0, Epoch 2}\label{fig:E_100_SYN_et0_FA_L}
\end{subfigure}
\begin{subfigure}[t]{.32\textwidth}
\includegraphics[width=1\linewidth]{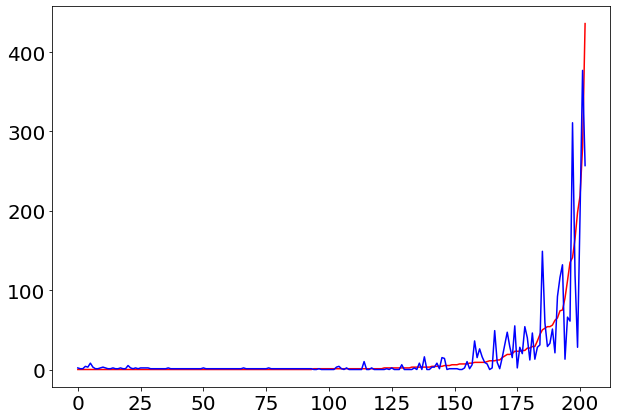} 
\caption{Std WGAN, L0, Epoch 20}\label{fig:E_1000_SYN_et0_FA_L}
\end{subfigure}
\medskip


\medskip

\centering
\begin{subfigure}[t]{.32\textwidth}
\includegraphics[width=1\linewidth]{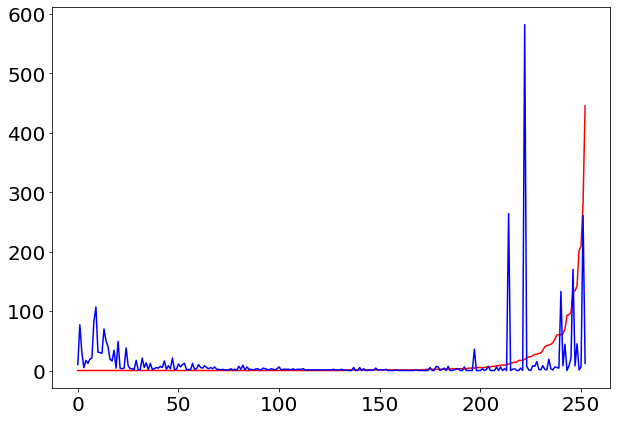} 
\caption{ST-WGAN, L0, Epoch 0}\label{fig:E_1_SYN_et0_FA_S}
\end{subfigure}
%
%
\begin{subfigure}[t]{.32\textwidth}
\includegraphics[width=1\linewidth]{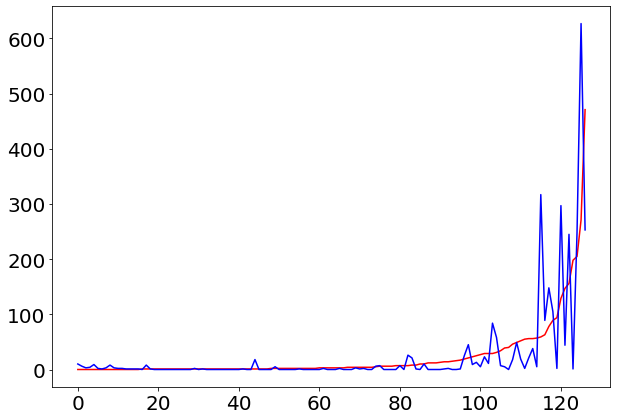} 
\caption{ST-WGAN, L0, Epoch 2}\label{fig:E_100_SYN_et0_FA_S}
\end{subfigure}
%
\begin{subfigure}[t]{.32\textwidth}
\includegraphics[width=1\linewidth]{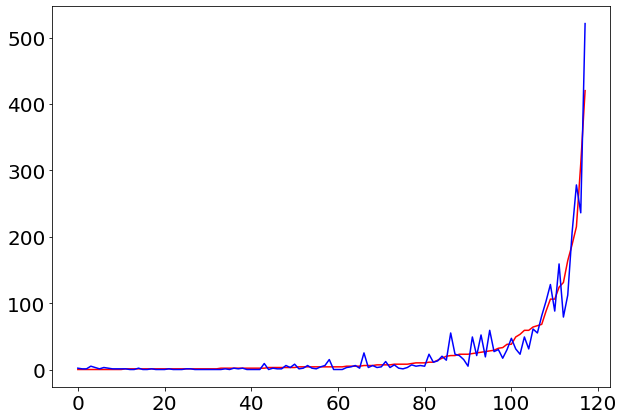} 
\caption{ST-WGAN, L0, Epoch 20}\label{fig:E_1000_SYN_et0_FA_S}
\end{subfigure}

\caption{Rendered data set (first use case). Comparison of synthetic (blue) and real (red) data distributions using WGAN generators for label 0 with linear activation (top row) (\ref{fig:E_1_SYN_et0_FA_L},  \ref{fig:E_100_SYN_et0_FA_L} and \ref{fig:E_1000_SYN_et0_FA_L} ) and with ST-based activation (bottom row) (\ref{fig:E_1_SYN_et0_FA_S}, \ref{fig:E_100_SYN_et0_FA_S}, and \ref{fig:E_1000_SYN_et0_FA_S}) in different epochs (1, 2 and 20).}
\label{fig:SYN_et0_EVO}
\end{figure*}

\begin{figure*}[ht!]

\centering
\centering
\begin{subfigure}[t]{.32\textwidth}
\includegraphics[width=1\linewidth]{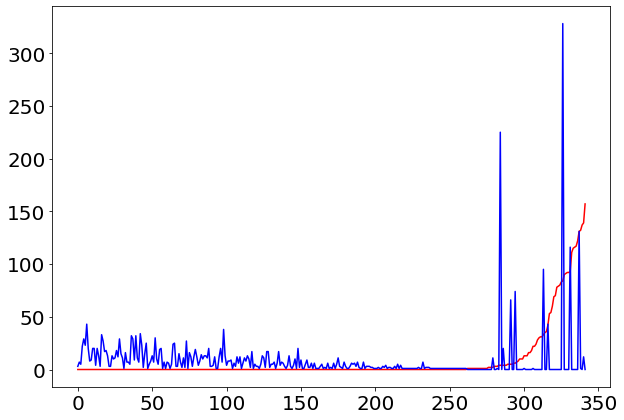} 
\caption{Std WGAN, L1, Epoch 0}\label{fig:E_1_SYN_et1_FA_L}
\end{subfigure}
\begin{subfigure}[t]{.32\textwidth}
\includegraphics[width=1\linewidth]{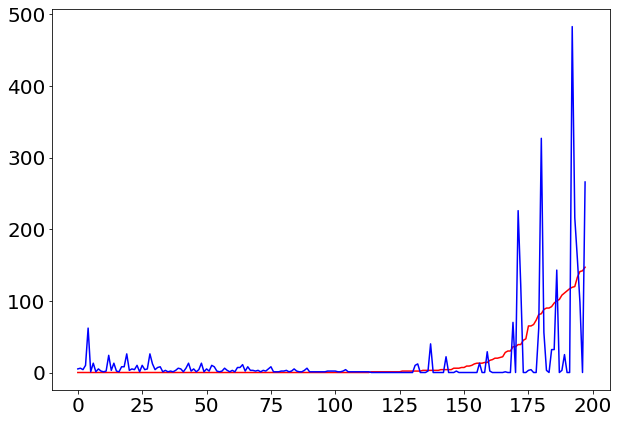} 
\caption{Std WGAN, L1, Epoch 2}\label{fig:E_100_SYN_et1_FA_L}
\end{subfigure}
\begin{subfigure}[t]{.32\textwidth}
\includegraphics[width=1\linewidth]{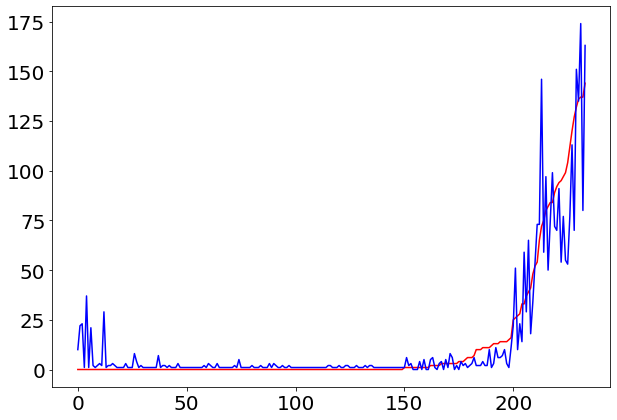} 
\caption{Std WGAN, L1, Epoch 20}\label{fig:E_1000_SYN_et1_FA_L}
\end{subfigure}

\medskip


\medskip

\centering
\begin{subfigure}[t]{.32\textwidth}
\includegraphics[width=1\linewidth]{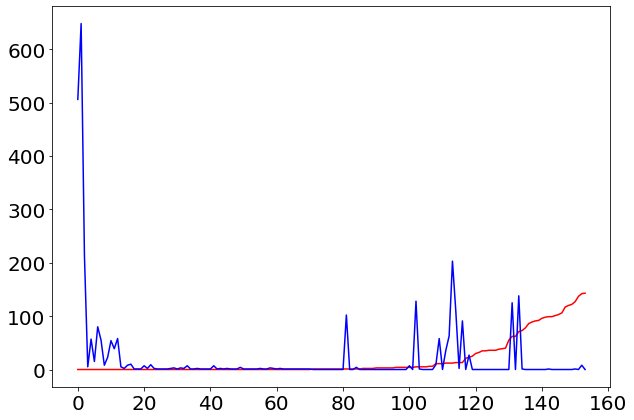} 
\caption{ST-WGAN, L1, Epoch 0}\label{fig:E_1_SYN_et1_FA_S}
\end{subfigure}
\begin{subfigure}[t]{.32\textwidth}
\includegraphics[width=1\linewidth]{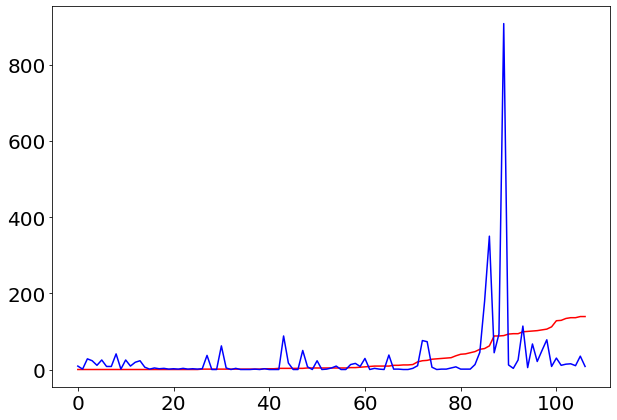} 
\caption{ST-WGAN, L1, Epoch 2}\label{fig:E_100_SYN_et1_FA_S}
\end{subfigure}
\begin{subfigure}[t]{.32\textwidth}
\includegraphics[width=1\linewidth]{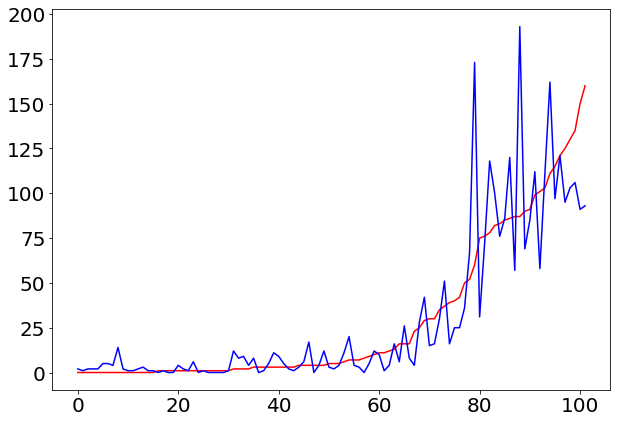} 
\caption{ST-WGAN, L1, Epoch 20}\label{fig:E_1000_SYN_et1_FA_S}
\end{subfigure}

\caption{Rendered data set (first use case). Comparison of synthetic (blue) and real (red) data distributions using WGAN generators for label 1 with linear activation (top row) (\ref{fig:E_1_SYN_et1_FA_L}, 
\ref{fig:E_100_SYN_et1_FA_L}, 
and \ref{fig:E_1000_SYN_et1_FA_L}) and with ST-based activation (bottom row) (\ref{fig:E_1_SYN_et1_FA_S}, 
\ref{fig:E_100_SYN_et1_FA_S}, 
and \ref{fig:E_1000_SYN_et1_FA_S}) in different epochs 
(0, 2 and 20). The $4$-dimensional vector has been flattened into by sorting by frequency in ascending order on the $x$-axis.
}
\label{fig:SYN_et1_EVO}
\end{figure*}

\begin{figure*}[ht!]
\centering
\centering
\begin{subfigure}[t]{.32\textwidth}
\includegraphics[width=1\linewidth]{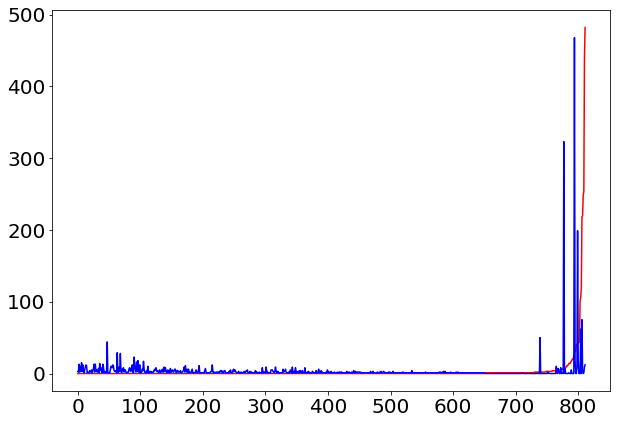} 
\caption{Std WGAN, L0, Epoch 0}\label{fig:E_1_CR_et0_FA_L}
\end{subfigure}
%
%
\begin{subfigure}[t]{.32\textwidth}
\includegraphics[width=1\linewidth]{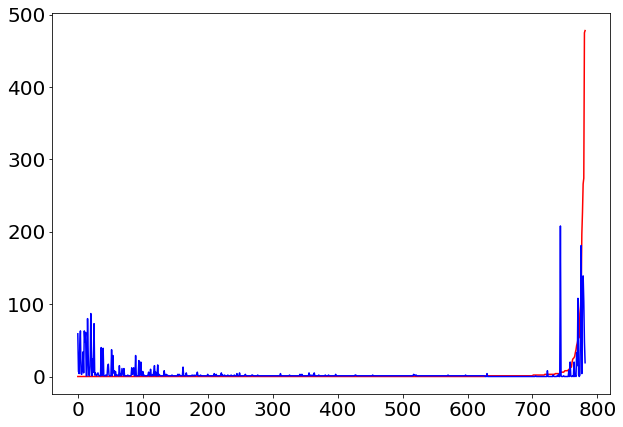} 
\caption{Std WGAN, L0, Epoch 2}\label{fig:E_100_CR_et0_FA_L}
\end{subfigure}
\begin{subfigure}[t]{.32\textwidth}
\includegraphics[width=1\linewidth]{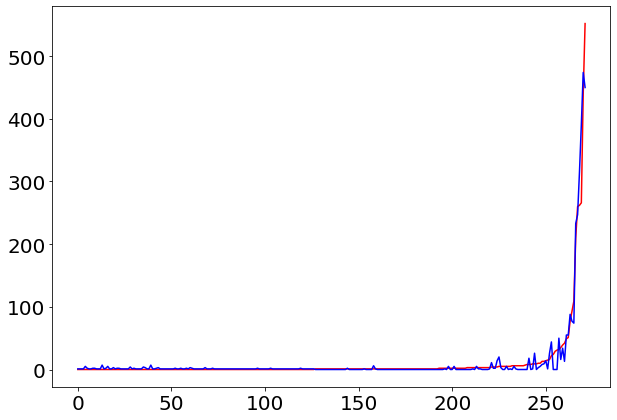} 
\caption{Std WGAN, L0, Epoch 20}\label{fig:E_1000_CR_et0_FA_L}
\end{subfigure}
\medskip


\medskip

\centering
\begin{subfigure}[t]{.32\textwidth}
\includegraphics[width=1\linewidth]{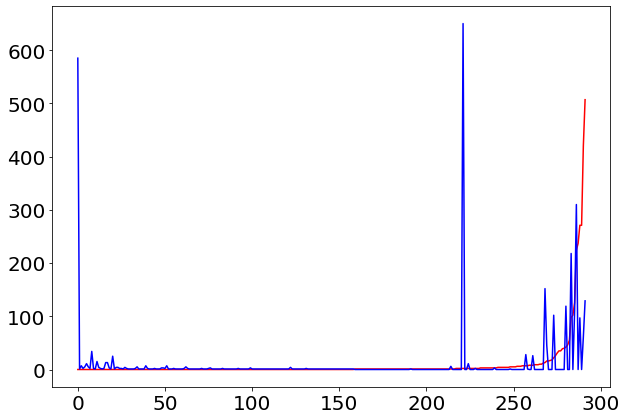} 
\caption{ST-WGAN, L0, Epoch 0}\label{fig:E_1_CR_et0_FA_S}
\end{subfigure}
%
%
\begin{subfigure}[t]{.32\textwidth}
\includegraphics[width=1\linewidth]{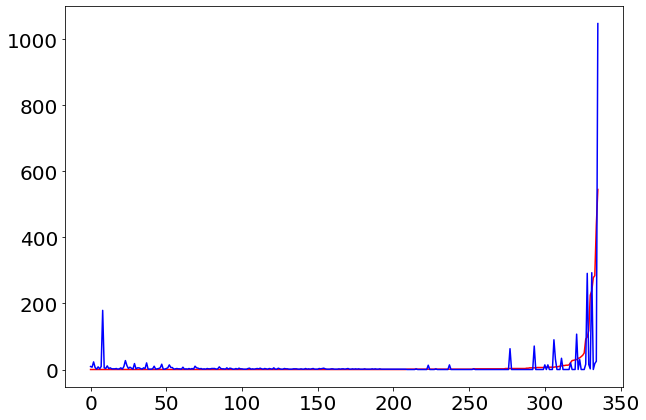} 
\caption{ST-WGAN, L0, Epoch 2}\label{fig:E_100_CR_et0_FA_S}
\end{subfigure}
%
\begin{subfigure}[t]{.32\textwidth}
\includegraphics[width=1\linewidth]{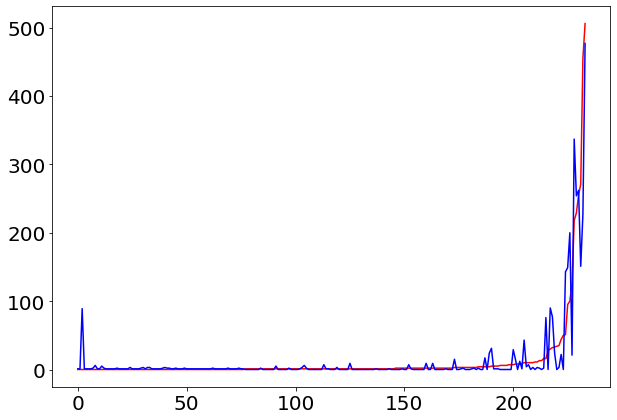} 
\caption{ST-WGAN, L0, Epoch 20}\label{fig:E_1000_CR_et0_FA_S}
\end{subfigure}

\caption{Cryptomining attack (second use case). Comparison of synthetic (blue) and real (red) data distributions using GAN generators for label 0 with linear activation (top row) (\ref{fig:E_1_CR_et0_FA_L},  \ref{fig:E_100_CR_et0_FA_L} and \ref{fig:E_1000_CR_et0_FA_L}) and with ST-based activation (bottom row) (\ref{fig:E_1_CR_et0_FA_S},  \ref{fig:E_100_CR_et0_FA_S} and \ref{fig:E_1000_CR_et0_FA_S}) in different epochs (0, 2 and 20). 
}
\label{fig:CR_et0_EVO}
\end{figure*}

\begin{figure*}[ht!]

\centering
\centering
\begin{subfigure}[t]{.32\textwidth}
\includegraphics[width=1\linewidth]{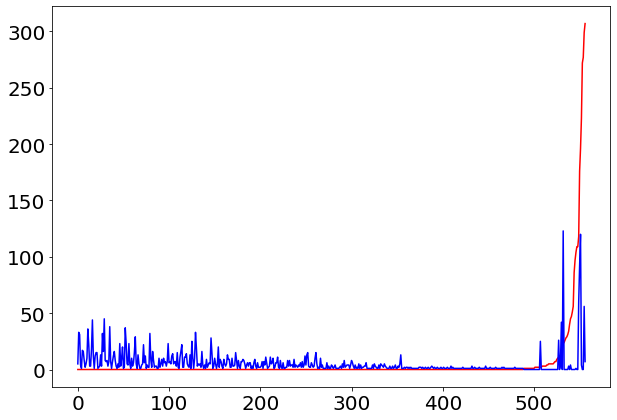} 
\caption{Std WGAN, L1, Epoch 0}\label{fig:E_1_CR_et1_FA_L}
\end{subfigure}
\begin{subfigure}[t]{.32\textwidth}
\includegraphics[width=1\linewidth]{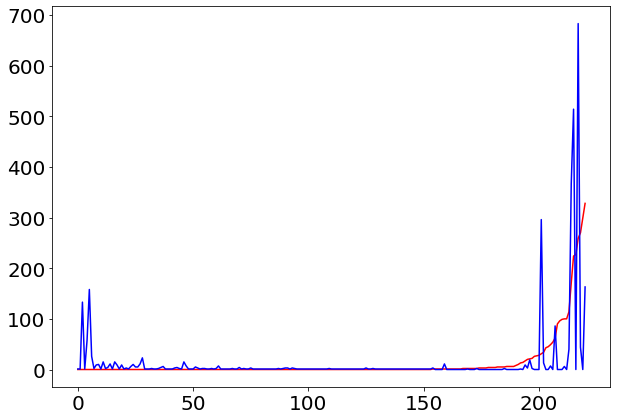} 
\caption{Std WGAN, L1, Epoch 2}\label{fig:E_100_CR_et1_FA_L}
\end{subfigure}
\begin{subfigure}[t]{.32\textwidth}
\includegraphics[width=1\linewidth]{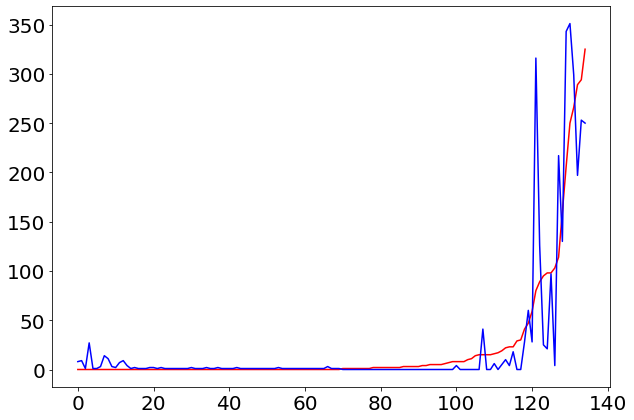} 
\caption{Std WGAN, L1, Epoch 20}\label{fig:E_1000_CR_et1_FA_L}
\end{subfigure}

\medskip


\medskip

\centering
\begin{subfigure}[t]{.32\textwidth}
\includegraphics[width=1\linewidth]{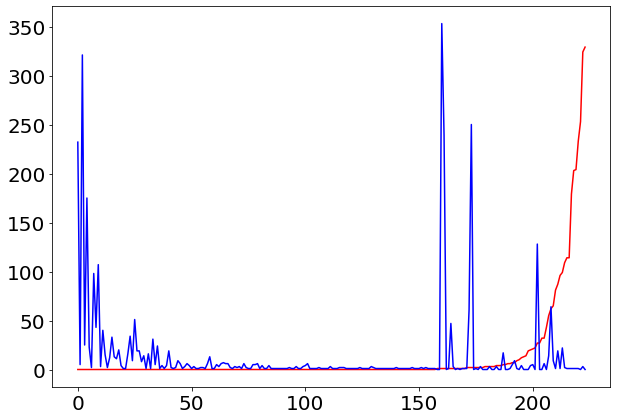} 
\caption{ST-WGAN, L1, Epoch 0}\label{fig:E_1_CR_et1_FA_S}
\end{subfigure}
\begin{subfigure}[t]{.32\textwidth}
\includegraphics[width=1\linewidth]{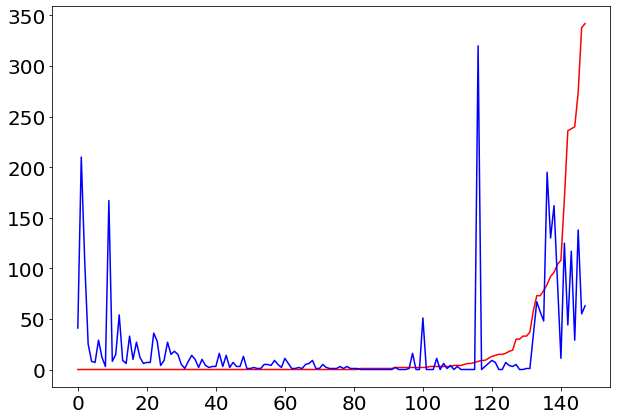} 
\caption{ST-WGAN, L1, Epoch 2}\label{fig:E_100_CR_et1_FA_S}
\end{subfigure}
\begin{subfigure}[t]{.32\textwidth}
\includegraphics[width=1\linewidth]{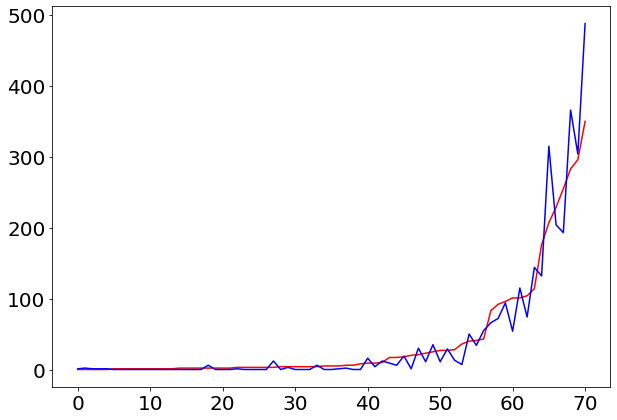} 
\caption{ST-WGAN, L1, Epoch 20}\label{fig:E_1000_CR_et1_FA_S}
\end{subfigure}

\caption{Cryptomining attack (second use case). Comparison of synthetic (blue) and real (red) data distributions using GAN generators for label 1 with linear activation (top row) (\ref{fig:E_1_CR_et1_FA_L}, 
\ref{fig:E_100_CR_et1_FA_L}, 
and \ref{fig:E_1000_CR_et1_FA_L}) and with ST-based activation (bottom row) (\ref{fig:E_1_CR_et1_FA_S}, 
\ref{fig:E_100_CR_et1_FA_S}, 
and \ref{fig:E_1000_CR_et1_FA_S}) in different epochs 
(0, 2 and 20).
}
\label{fig:CR_et1_EVO}
\end{figure*}

In this section, we analyze the evolution of the quality of the synthetic data generated by the standard and ST-based WGANs with respect to the real data. Figures \ref{fig:SYN_et0_EVO} and \ref{fig:SYN_et1_EVO} for the first use case, and Figures \ref{fig:CR_et0_EVO} and \ref{fig:CR_et1_EVO} for the second, show graphically the obtained distributions. To this end, we compare and plot samples of real and synthetic data distributions at different mini-batches (1, 100, and 1000) corresponding to the epochs $0$, $2$ and $20$ respectively.

%
%


To ease the visualization of the above-mentioned figures, we have flattened the four histograms corresponding to the four features of the data set into a $1$-dimensional plot. To be precise, each of the points in the $x$-axis of the plot represents a cube (a bin) in which the empirical probability density function was computed (see Section  \ref{sec:distances}).
In this way, comparing the cubes of two samples, we can infer whether the two data distributions are similar or not.
For example, samples of two data distributions with elements placed in different cubes would point to data distributions with significant differences. On the contrary, if the elements of both distributions are mapped to the same cubes and the number of elements in each cube is similar for both samples, we could infer that the two data distributions are similar. 
%

To plot and compare the synthetic and real data distributions, the cubes of the real data sample are sorted by the number of elements in each cube in ascending order. In this way, the marks on the $x$-axis represent the cubes as  ordered for the real data sample. Therefore, the real data curve always exhibits an ascending shape.  
The $y$-axis represents the number of sample elements that are placed in each cube.
It is worth noting that higher numbers on the $x$-axis indicate 
that the WGAN has created many nonexistent elements (in the real data set) that are assigned to new bins.
These bins containing elements outside the real data domain are placed on the left side of the WGAN curve since the real data curve has no elements in such bins.

In the first use case, it can be observed in Figure \ref{fig:SYN_et0_EVO} that the standard WGAN for label $0$ creates many more nonexistent data elements than the ST-based WGAN. Considering that the real data samples generate around $80$ cubes, at epoch $2$, the standard WGAN created $350$ cubes and the ST-based WGAN only $120$, and when they reach epoch $20$, the standard WGAN generated $200$ cubes and the ST-based WGAN remained at $120$. A similar behaviour is observed in Figure \ref{fig:SYN_et1_EVO} for label $1$, where the standard WGAN, in the three epochs discussed above, doubles the number of cubes generated by the ST-WGAN.
The situation in the second use case is similar to the first. The ST-based WGAN generates significantly fewer cubes than the standard WGAN for both labels (Figure \ref{fig:CR_et0_EVO} and Figure \ref{fig:CR_et1_EVO}).


\begin{figure*}[!t]
\centering
\begin{subfigure}[t]{.24\textwidth}
\includegraphics[width=1\linewidth]{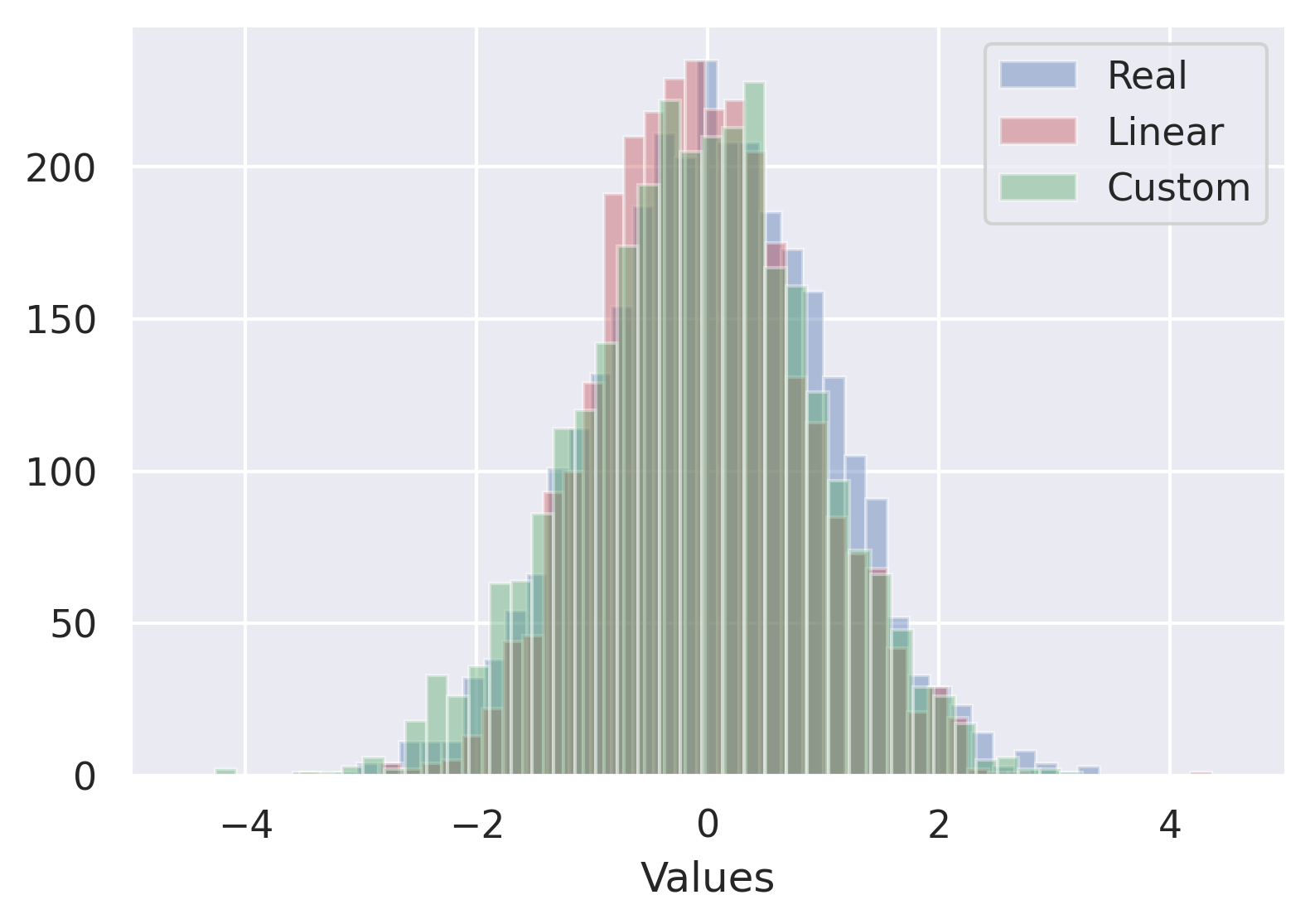} 
\caption{Feature 0.}\label{fig:SYN-distr-l0-f0}
\end{subfigure}
\begin{subfigure}[t]{.24\textwidth}
\includegraphics[width=1\linewidth]{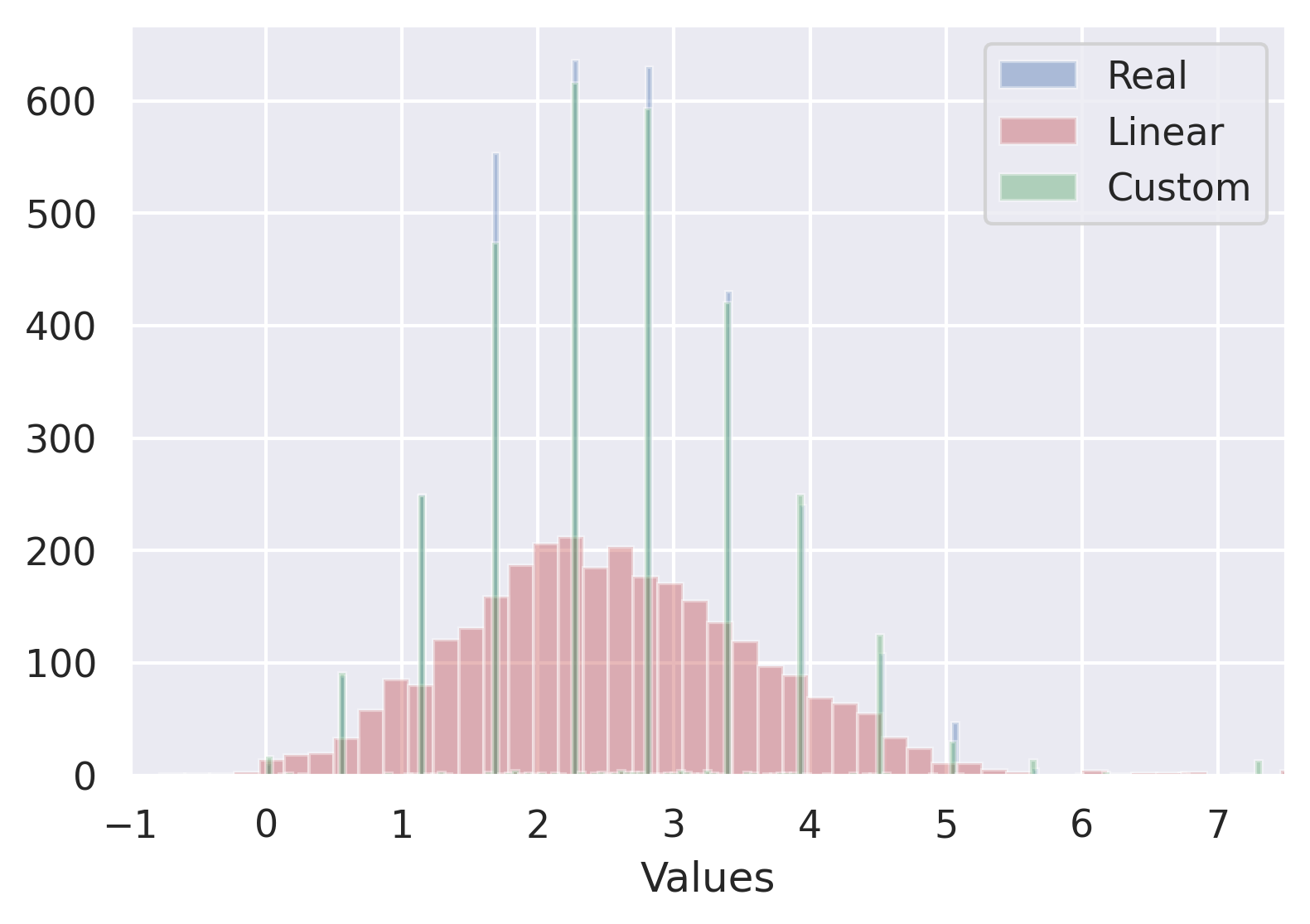} 
\caption{Feature 1.}\label{fig:SYN-distr-l0-f1}
\end{subfigure}
\begin{subfigure}[t]{.24\textwidth}
\includegraphics[width=1\linewidth]{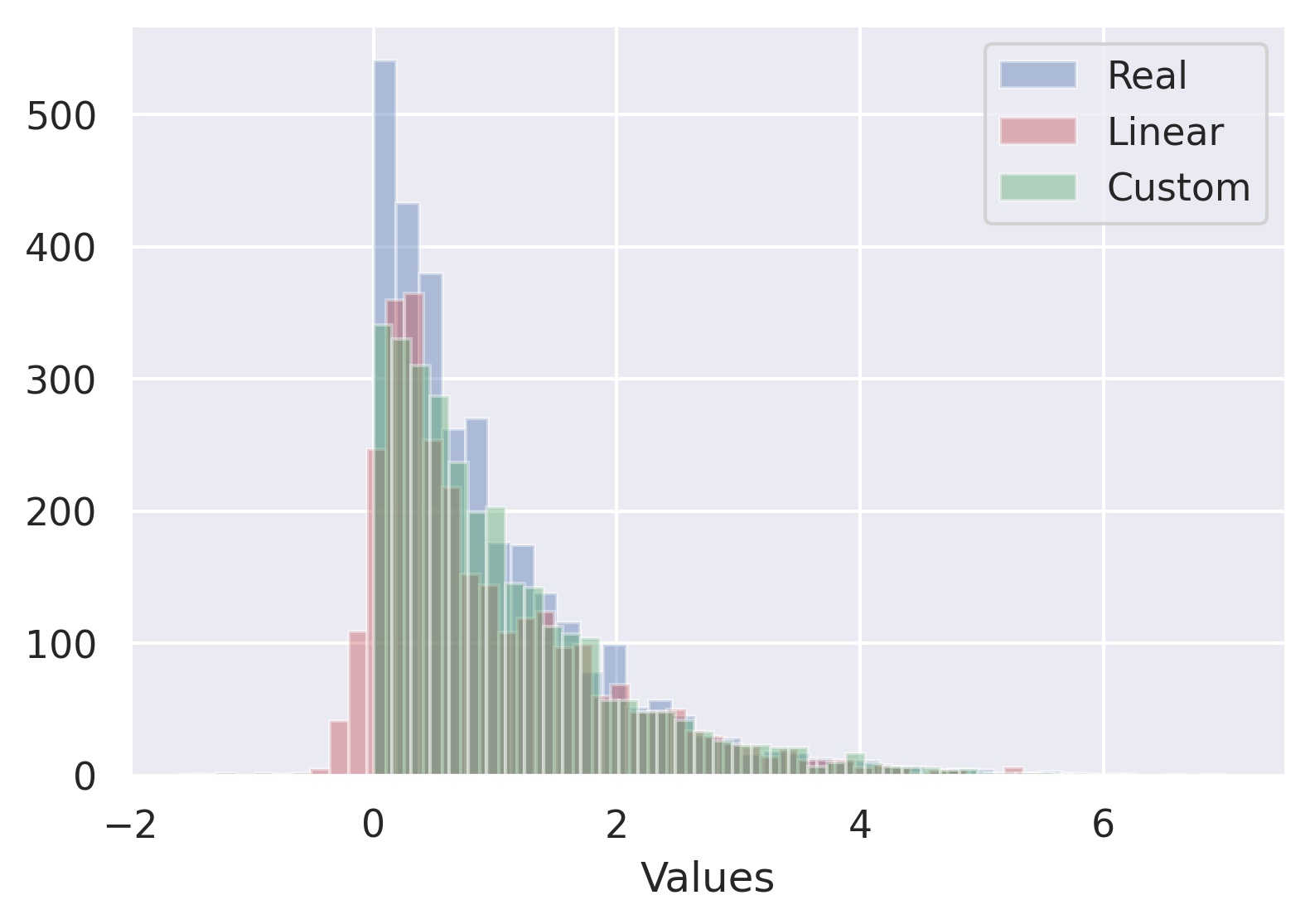} 
\caption{Feature 2.}\label{fig:SYN-distr-l0-f2}
\end{subfigure}
\begin{subfigure}[t]{.24\textwidth}
\includegraphics[width=1\linewidth]{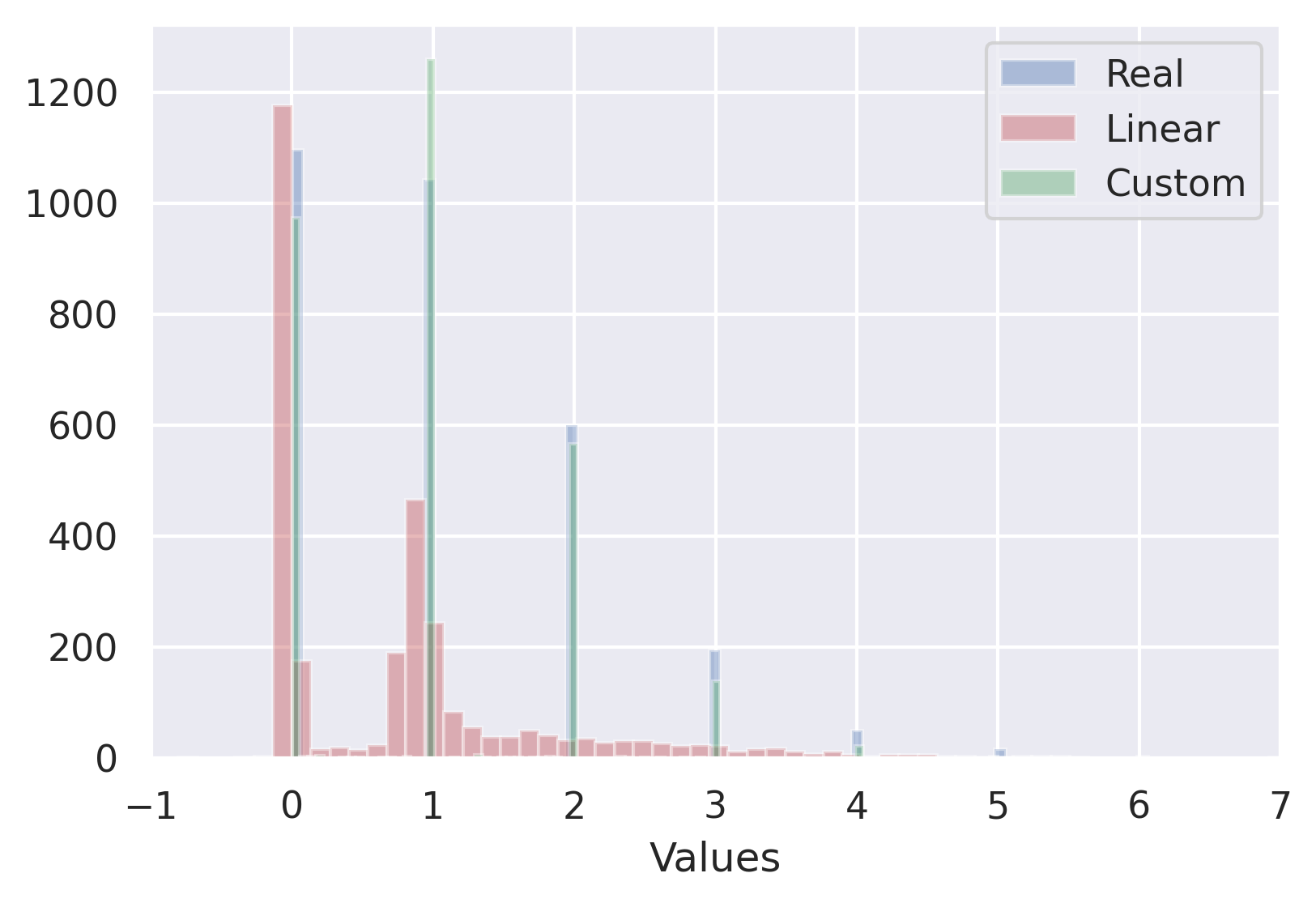} 
\caption{Feature 3.}\label{fig:SYN-distr-l0-f3}
\end{subfigure}

\begin{subfigure}[t]{.24\textwidth}
\includegraphics[width=1\linewidth]{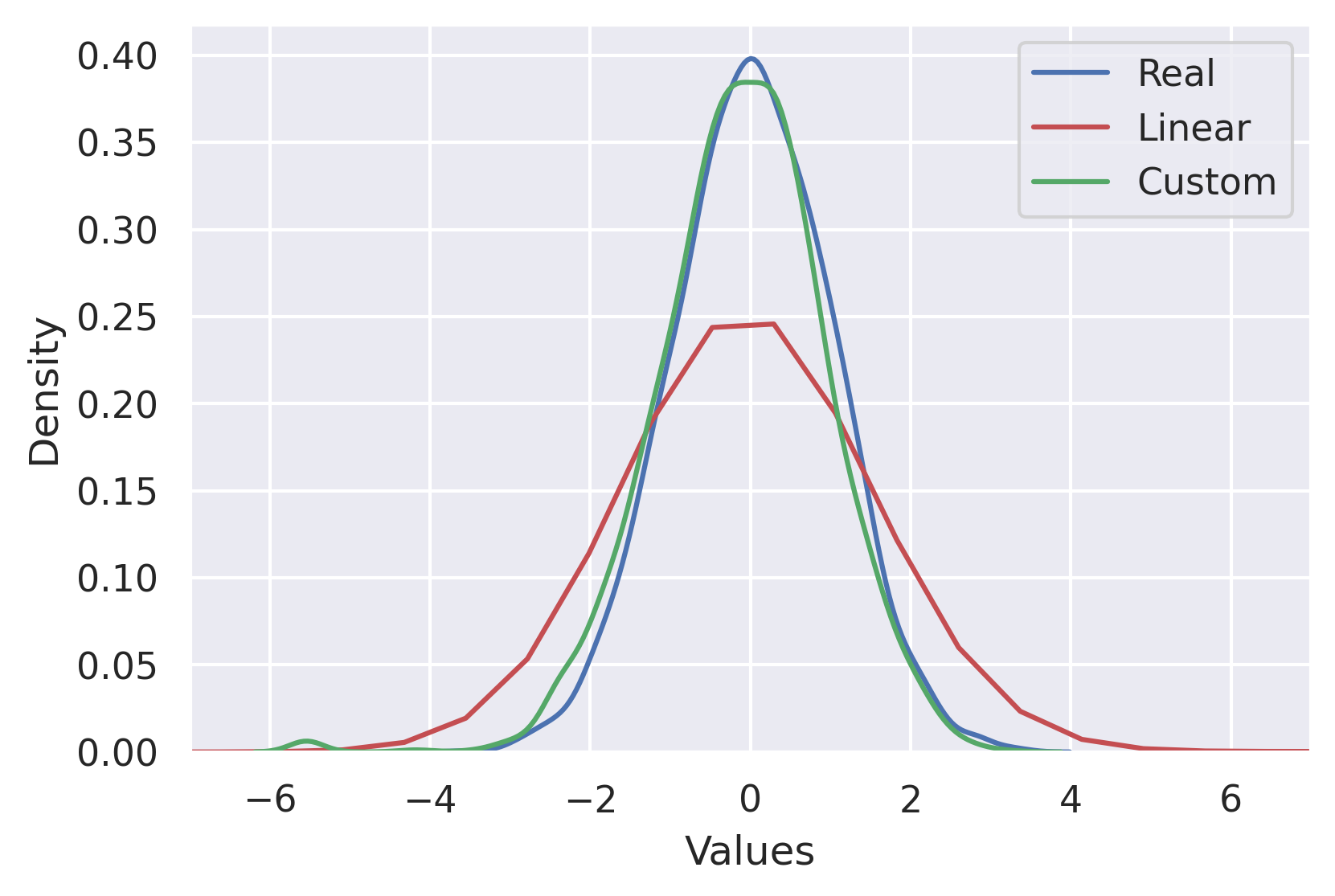} 
\caption{Feature 0.}\label{fig:SYN-distr-kde-l0-f0}
\end{subfigure}
\begin{subfigure}[t]{.24\textwidth}
\includegraphics[width=1\linewidth]{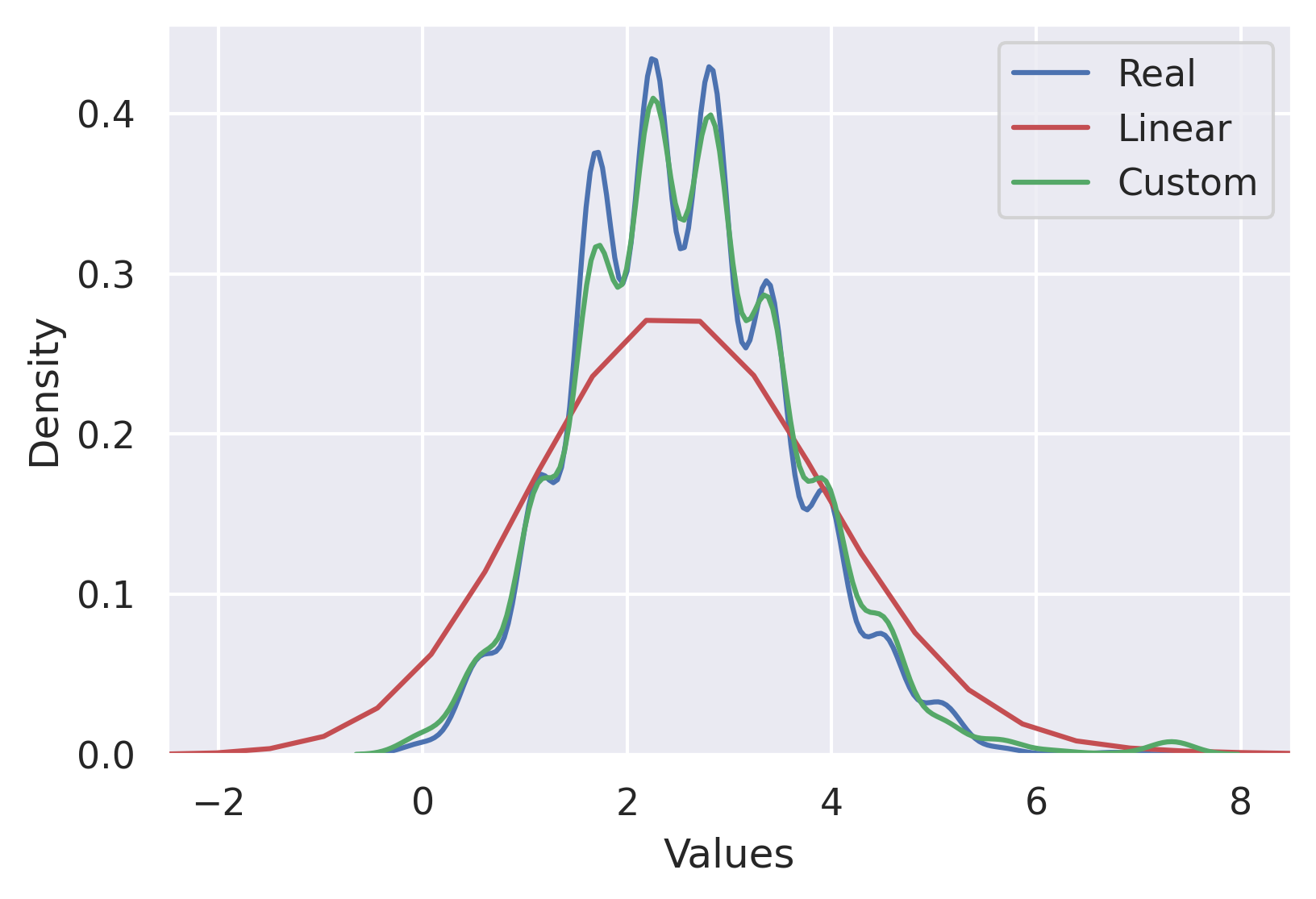} 
\caption{Feature 1.}\label{fig:SYN-distr-kde-l0-f1}
\end{subfigure}
\begin{subfigure}[t]{.24\textwidth}
\includegraphics[width=1\linewidth]{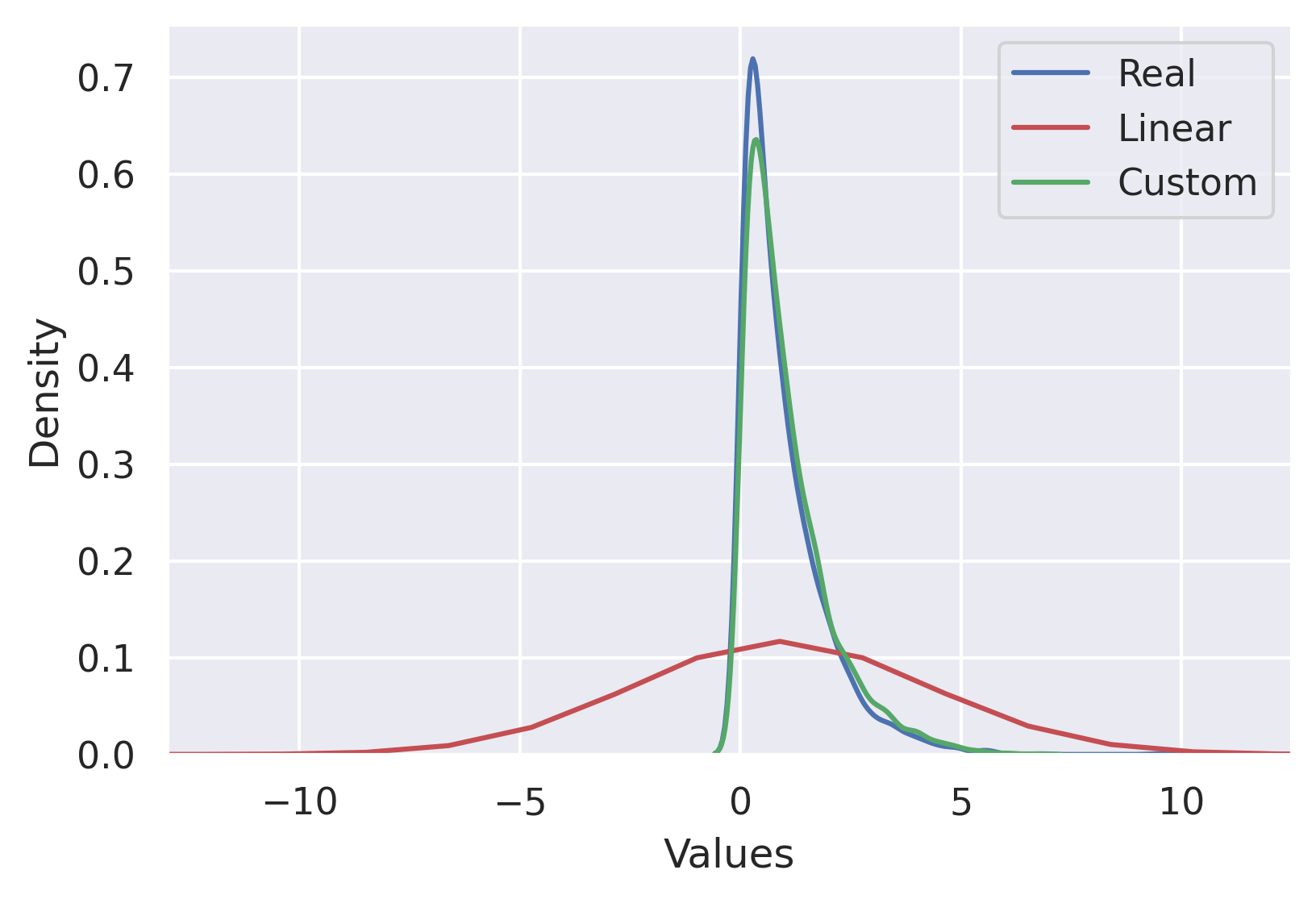} 
\caption{Feature 2.}\label{fig:SYN-distr-kde-l0-f2}
\end{subfigure}
\begin{subfigure}[t]{.24\textwidth}
\includegraphics[width=1\linewidth]{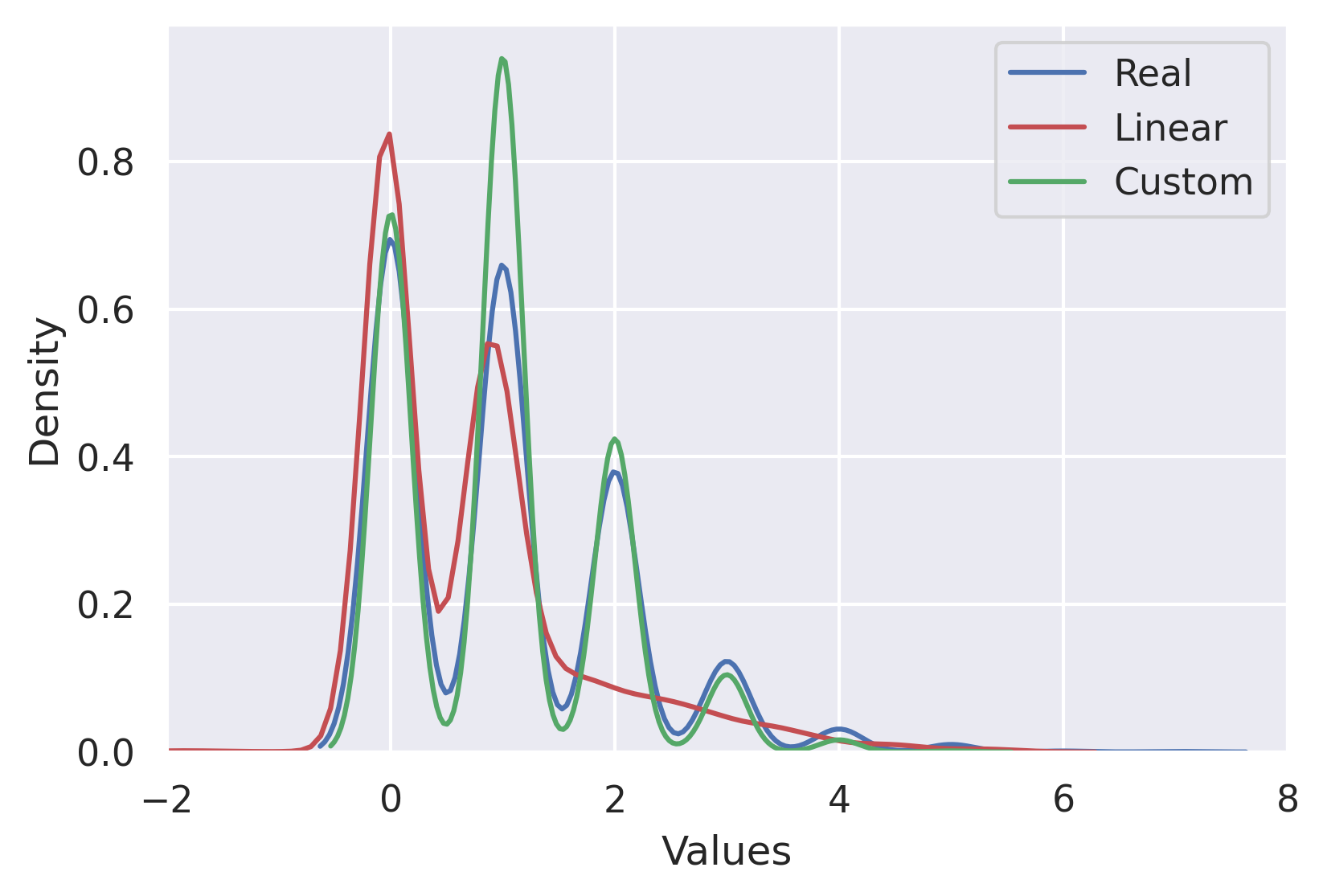} 
\caption{Feature 3.}\label{fig:SYN-distr-kde-l0-f3}
\end{subfigure}

\caption{Rendered data set (first use case). Frequency distribution from Label $0$ of real data and standard and ST-based WGAN's synthetic data. Histogram (top row) and Kernel Density Estimator (KDE) function (bottom row) are shown for the four variables.}
\label{fig:SYN-distr-l0}
\end{figure*}


\begin{figure*}[!t]
\centering
\begin{subfigure}[t]{.24\textwidth}
\includegraphics[width=1\linewidth]{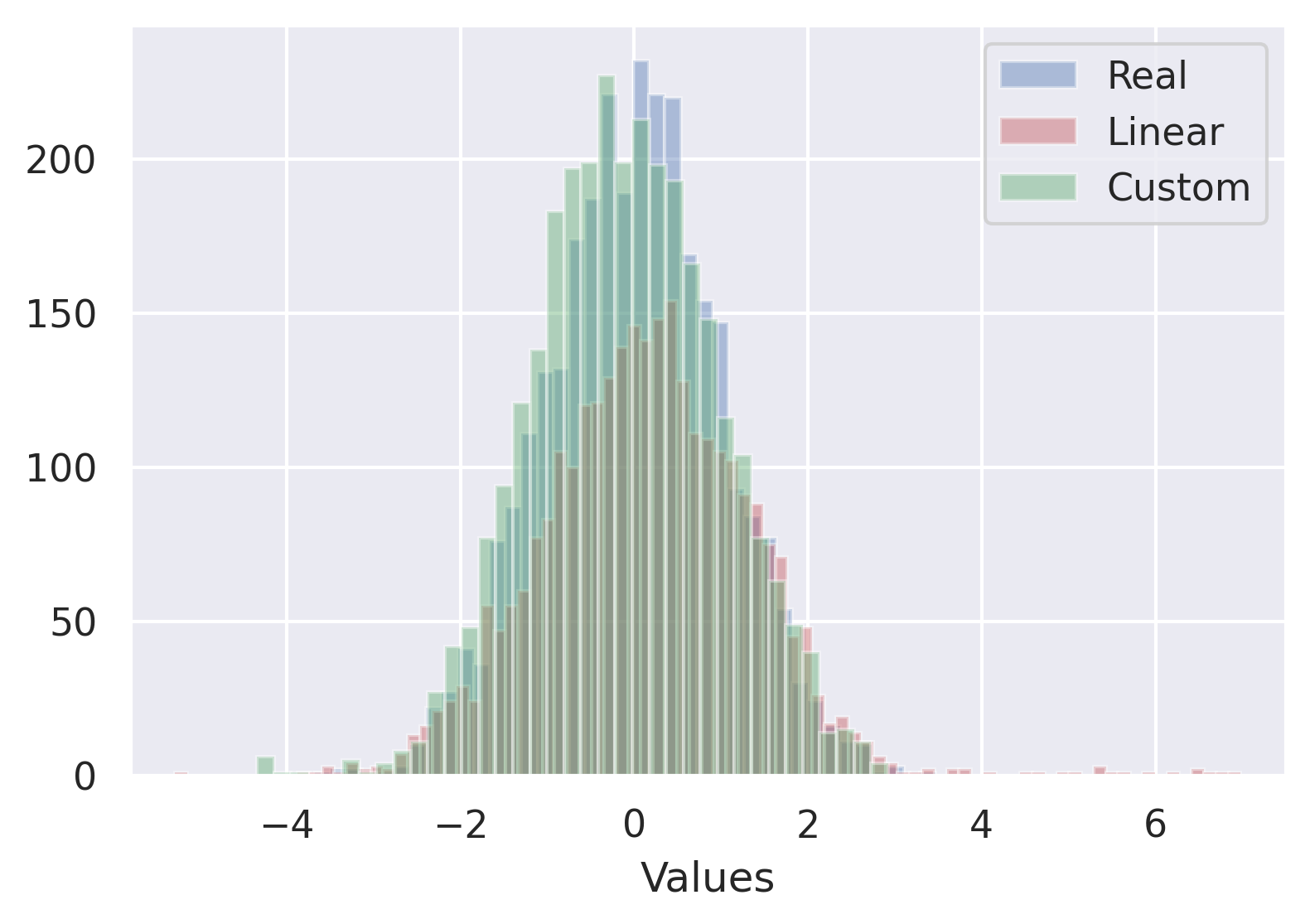} 
\caption{Feature 0.}\label{fig:SYN-distr-l1-f0}
\end{subfigure}
\begin{subfigure}[t]{.24\textwidth}
\includegraphics[width=1\linewidth]{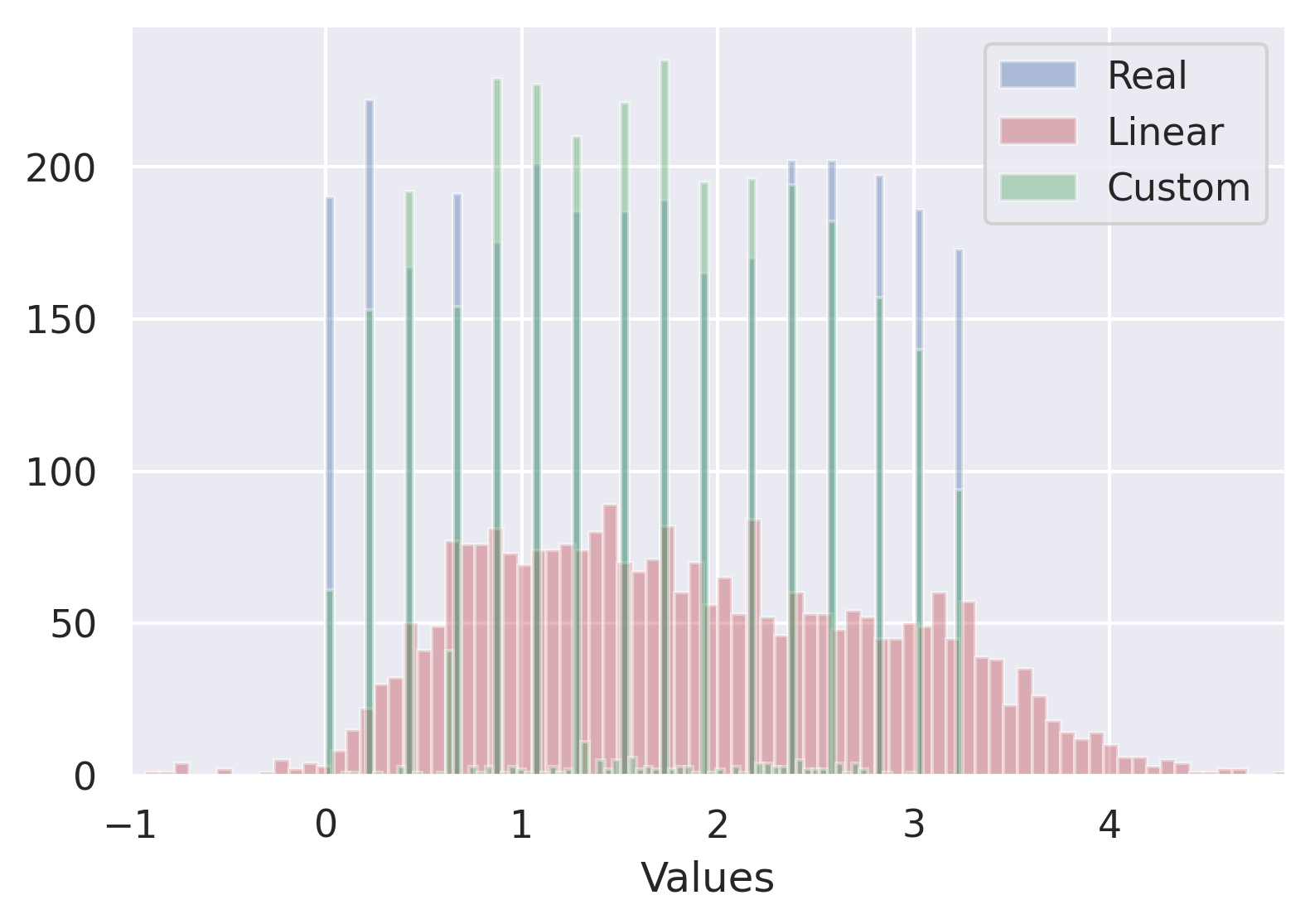} 
\caption{Feature 1.}\label{fig:SYN-distr-l1-f1}
\end{subfigure}
\begin{subfigure}[t]{.24\textwidth}
\includegraphics[width=1\linewidth]{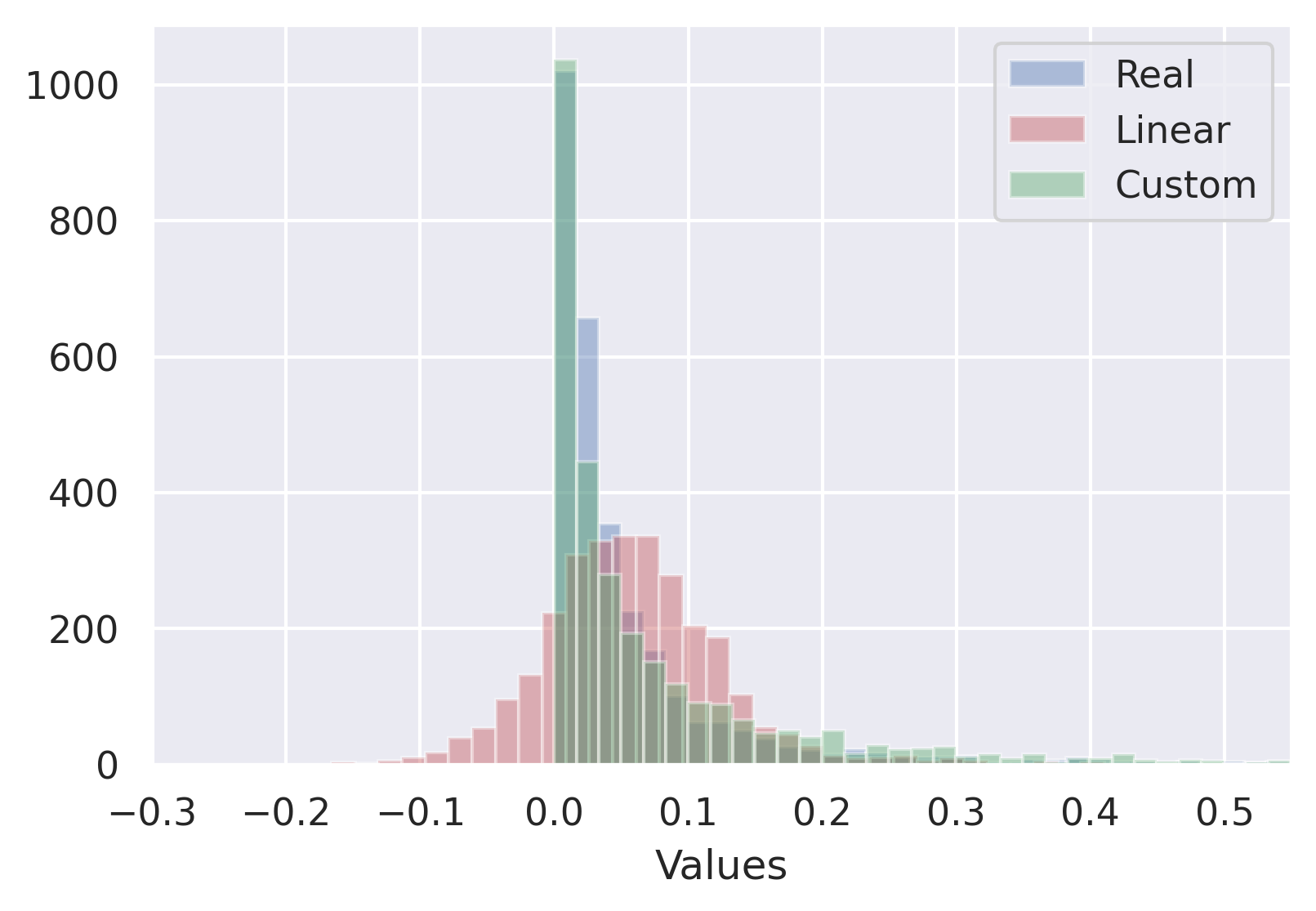} 
\caption{Feature 2.}\label{fig:SYN-distr-l1-f2}
\end{subfigure}
\begin{subfigure}[t]{.24\textwidth}
\includegraphics[width=1\linewidth]{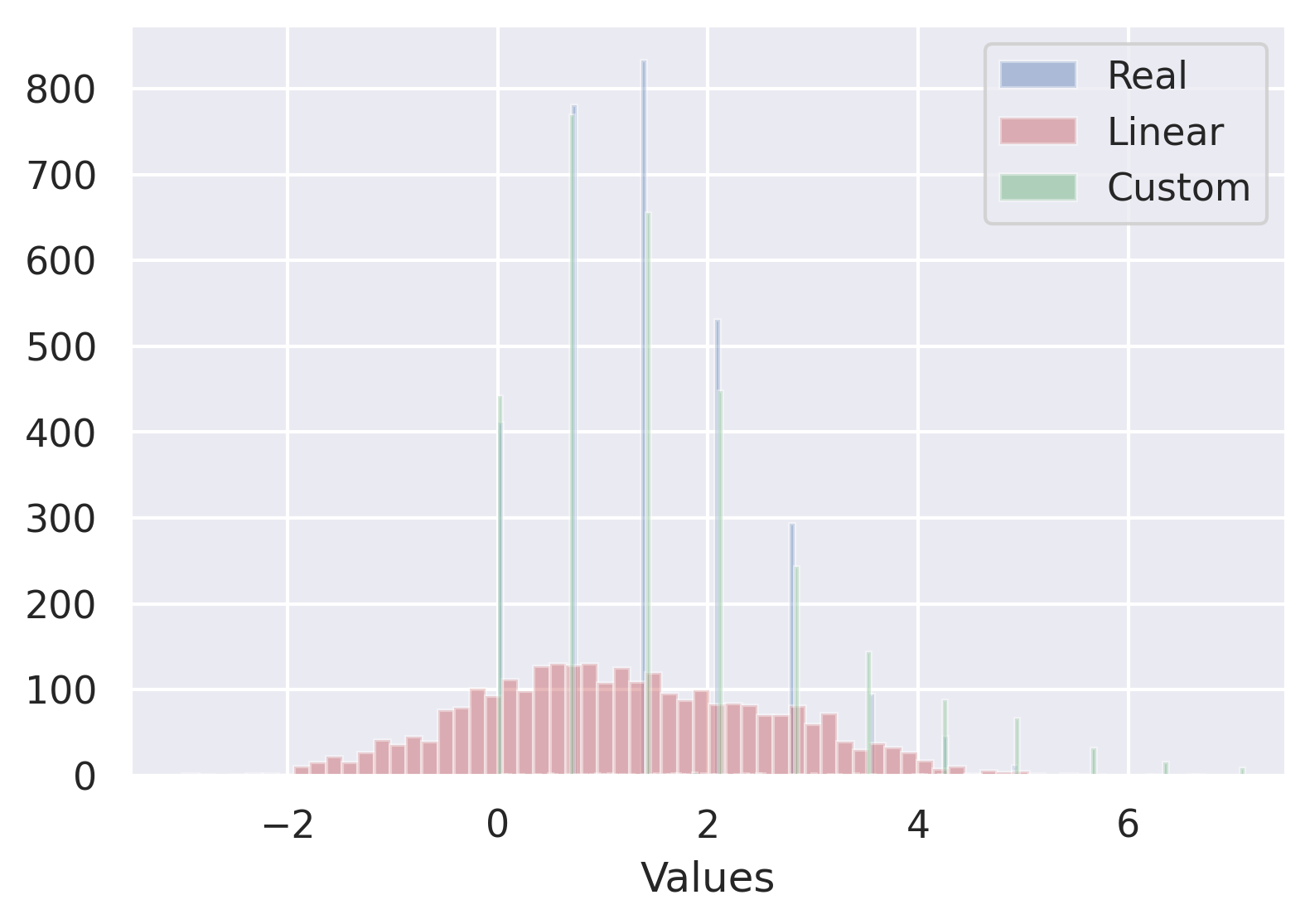} 
\caption{Feat 1. Real data.}\label{fig:SYN-distr-l1-f3}
\end{subfigure}

\begin{subfigure}[t]{.24\textwidth}
\includegraphics[width=1\linewidth]{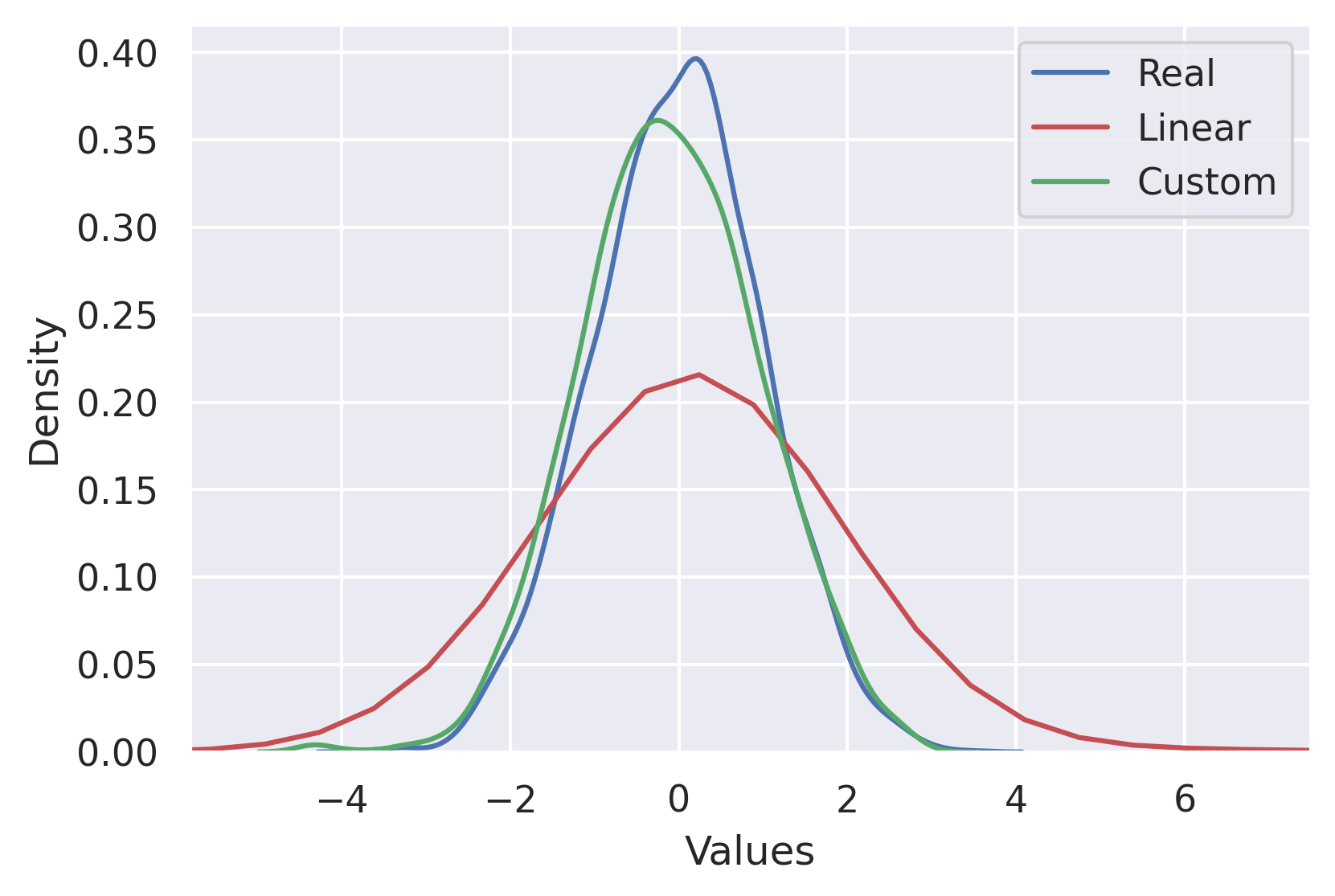} 
\caption{Feature 0.}\label{fig:SYN-distr-kde-l1-f0}
\end{subfigure}
\begin{subfigure}[t]{.24\textwidth}
\includegraphics[width=1\linewidth]{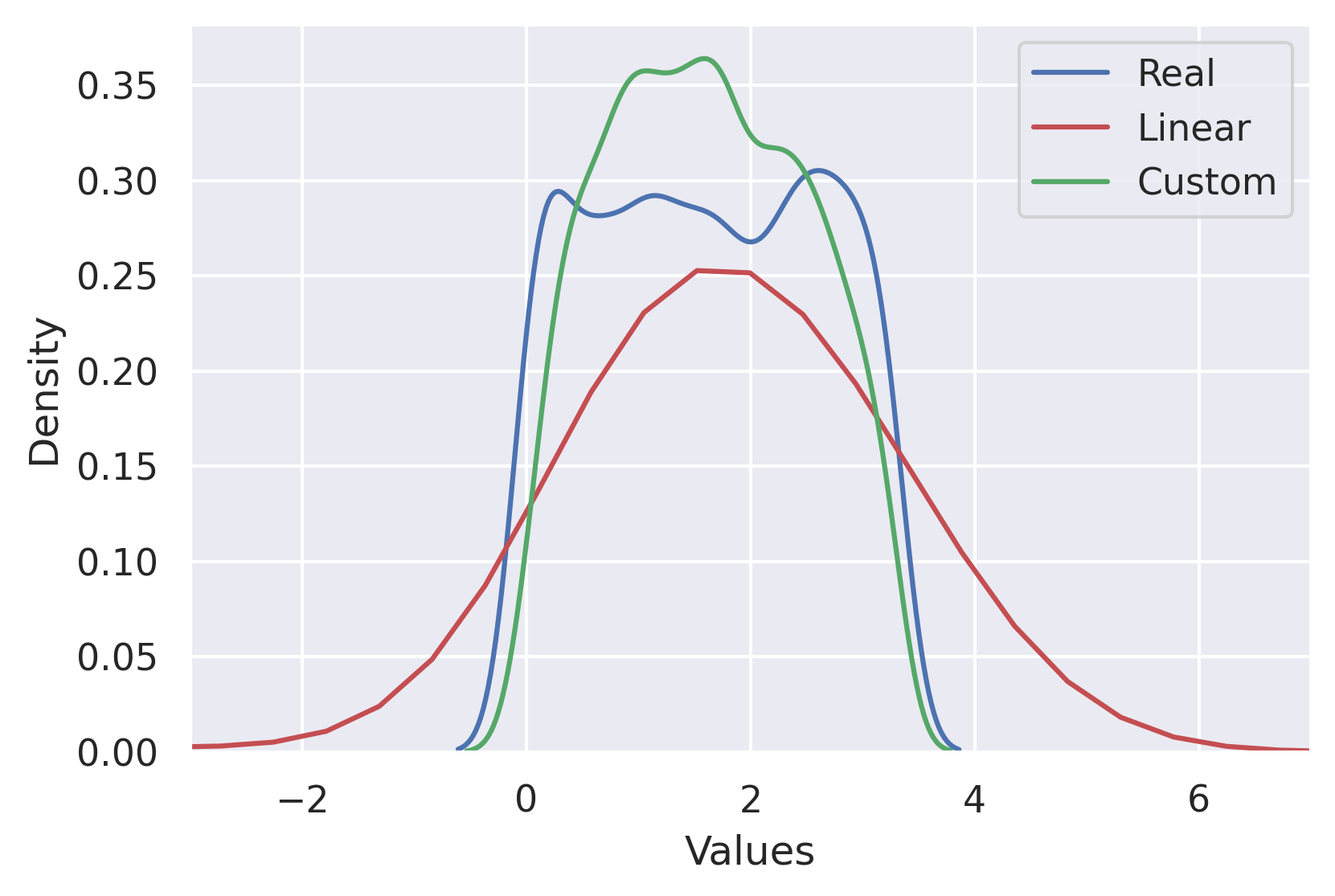} 
\caption{Feature 1.}\label{fig:SYN-distr-kde-l1-f1}
\end{subfigure}
\begin{subfigure}[t]{.24\textwidth}
\includegraphics[width=1\linewidth]{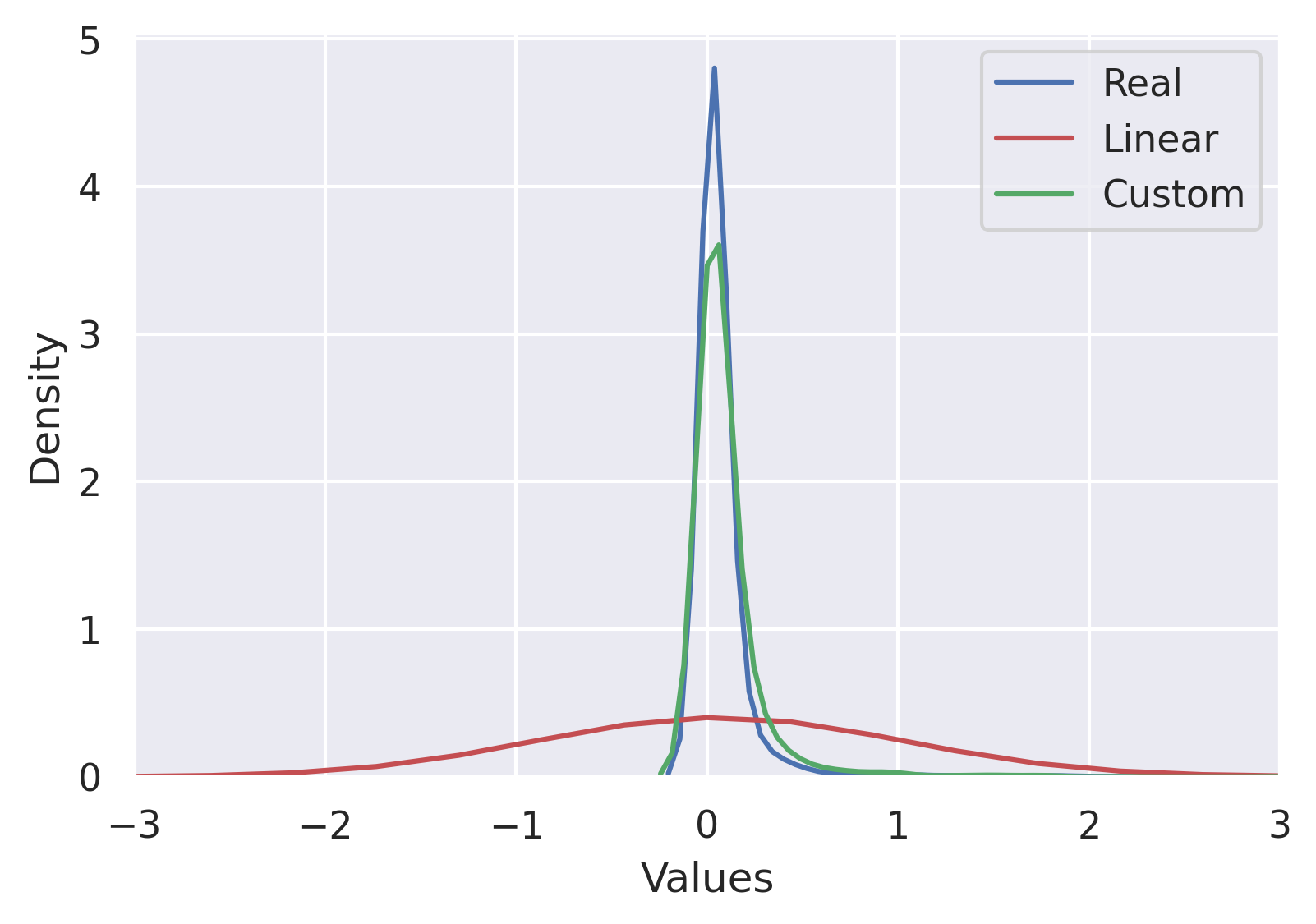} 
\caption{Feature 2.}\label{fig:SYN-distr-kde-l1-f2}
\end{subfigure}
\begin{subfigure}[t]{.24\textwidth}
\includegraphics[width=1\linewidth]{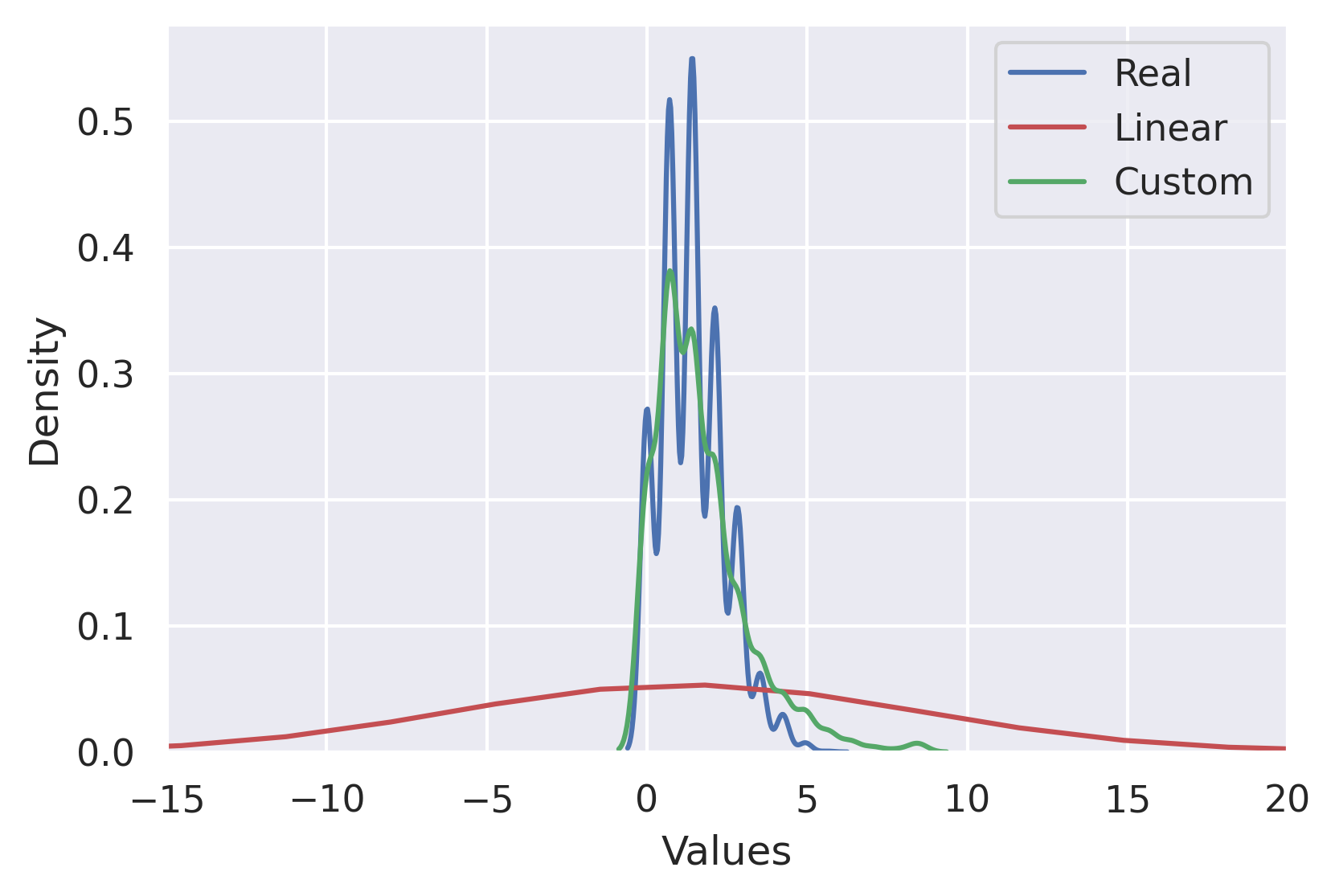} 
\caption{Feat 1. Real data.}\label{fig:SYN-distr-kde-l1-f3}
\end{subfigure}

\caption{Rendered data set (first use case). Frequency distribution from Label $1$ of real data and standard and ST-based WGAN's synthetic data. Histogram (top row) and Kernel Density Estimator (KDE) function (bottom row) are shown for the four variables.}
\label{fig:SYN-distr-l1}
\end{figure*}


\begin{figure*}[!t]
\centering
\begin{subfigure}[t]{.24\textwidth}
\includegraphics[width=1\linewidth]{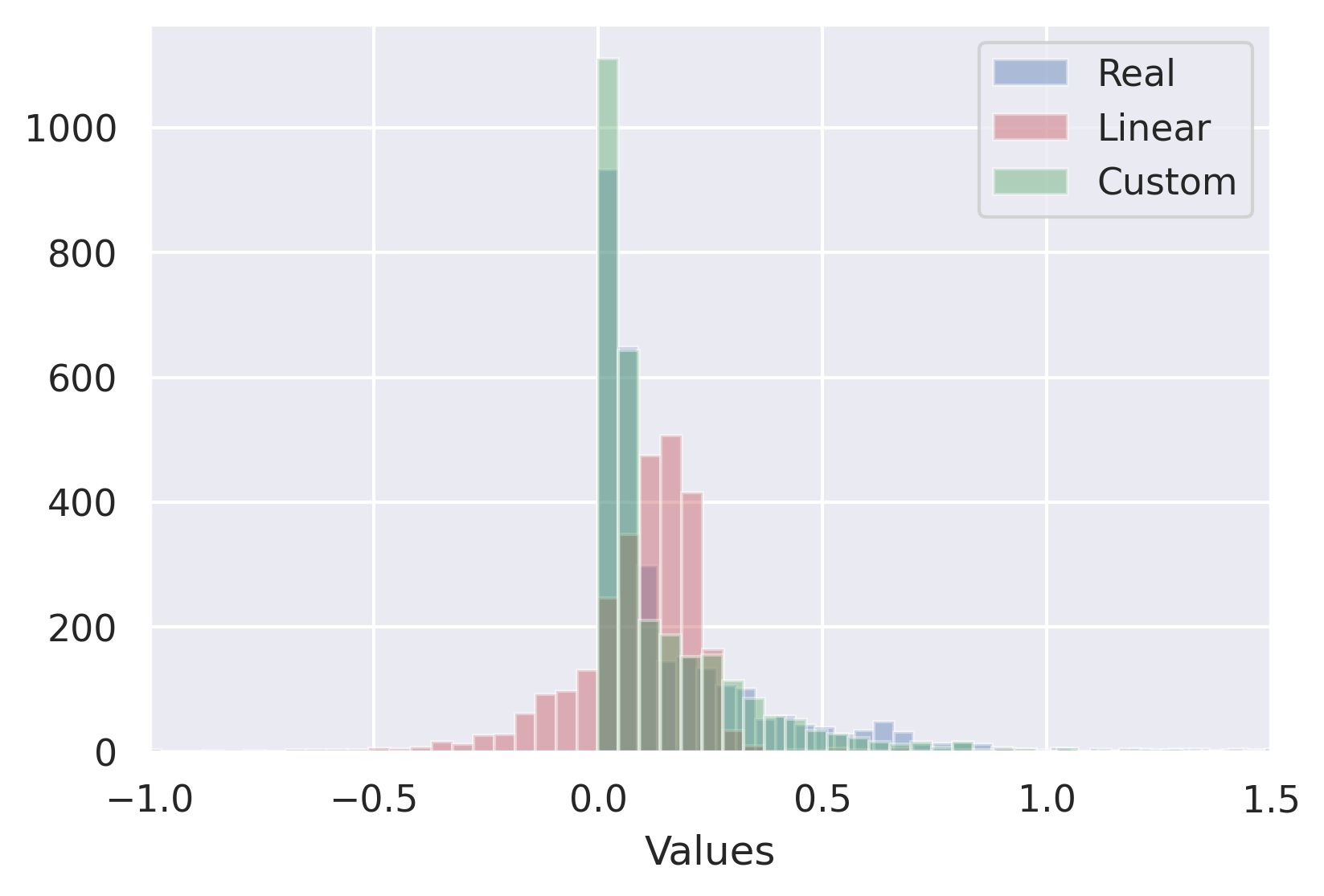} 
\caption{Feature 0.}\label{fig:CR-distr-l0-f0}
\end{subfigure}
\begin{subfigure}[t]{.24\textwidth}
\includegraphics[width=1\linewidth]{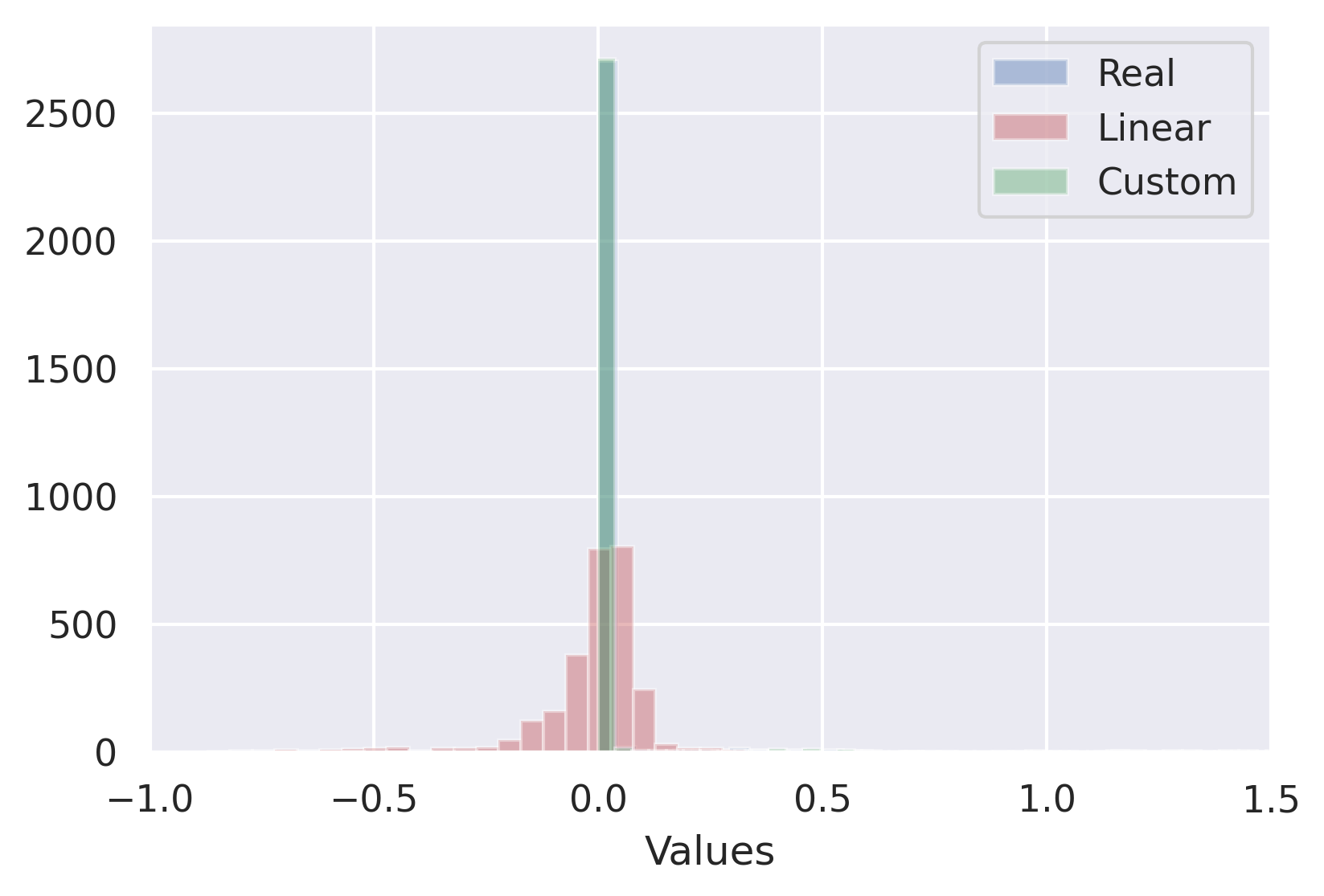} 
\caption{Feature 1.}\label{fig:CR-distr-l0-f1}
\end{subfigure}
\begin{subfigure}[t]{.24\textwidth}
\includegraphics[width=1\linewidth]{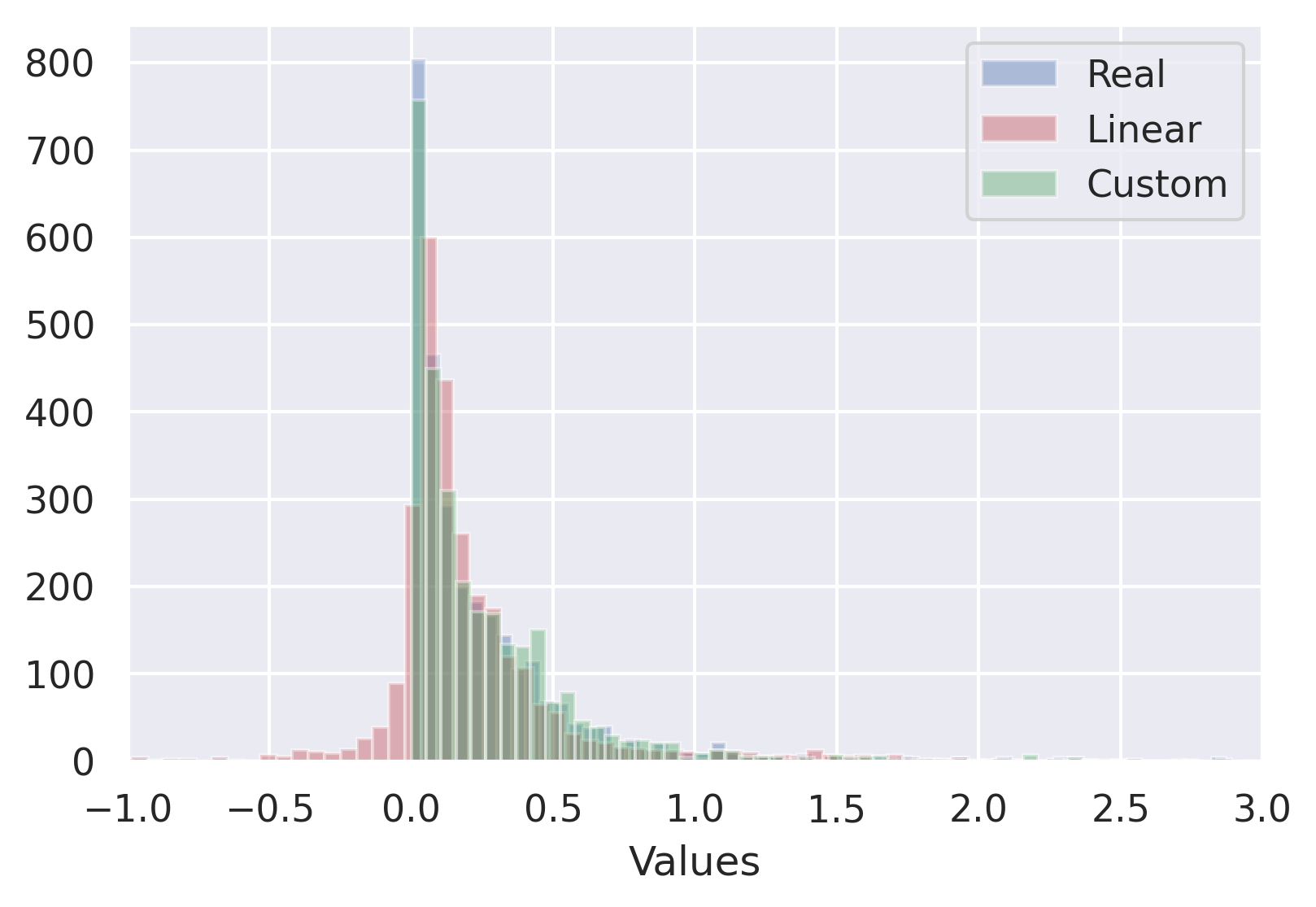} 
\caption{Feature 2.}\label{fig:CR-distr-l0-f2}
\end{subfigure}
\begin{subfigure}[t]{.24\textwidth}
\includegraphics[width=1\linewidth]{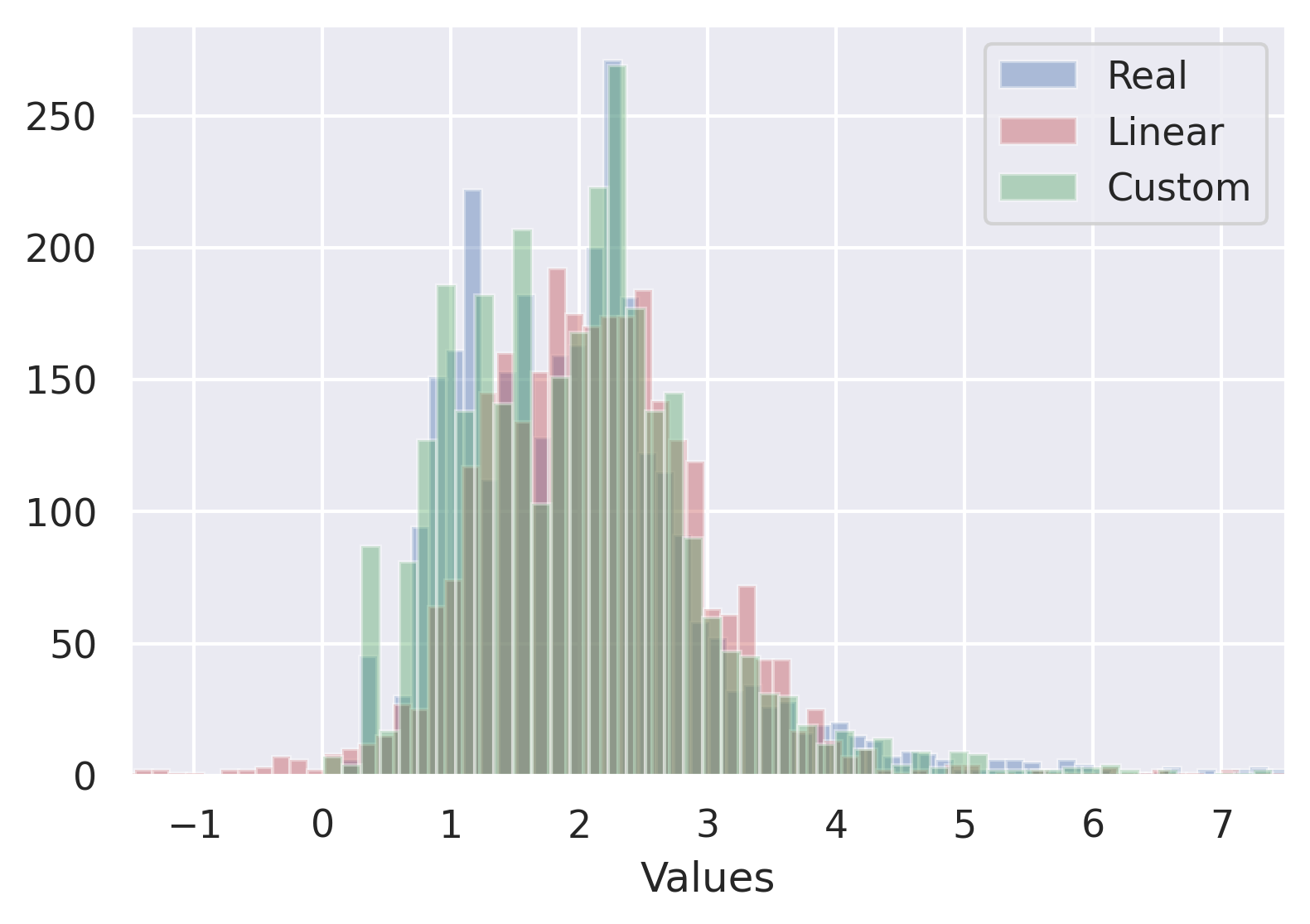} 
\caption{Feat 1. Real data.}\label{fig:CR-distr-l0-f31}
\end{subfigure}

\begin{subfigure}[t]{.24\textwidth}
\includegraphics[width=1\linewidth]{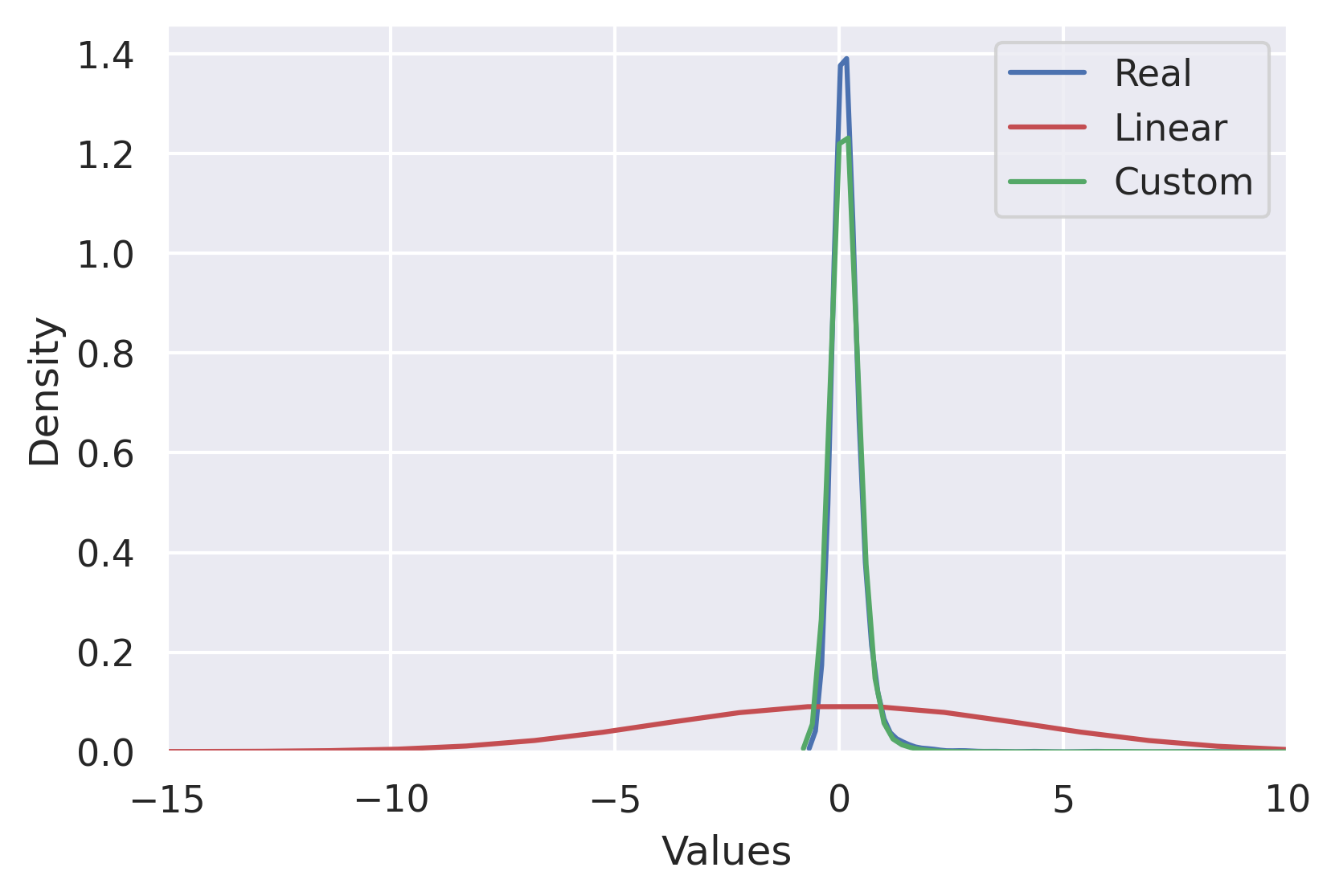} 
\caption{Feature 0.}\label{fig:CR-distr-kde-l0-f0}
\end{subfigure}
\begin{subfigure}[t]{.24\textwidth}
\includegraphics[width=1\linewidth]{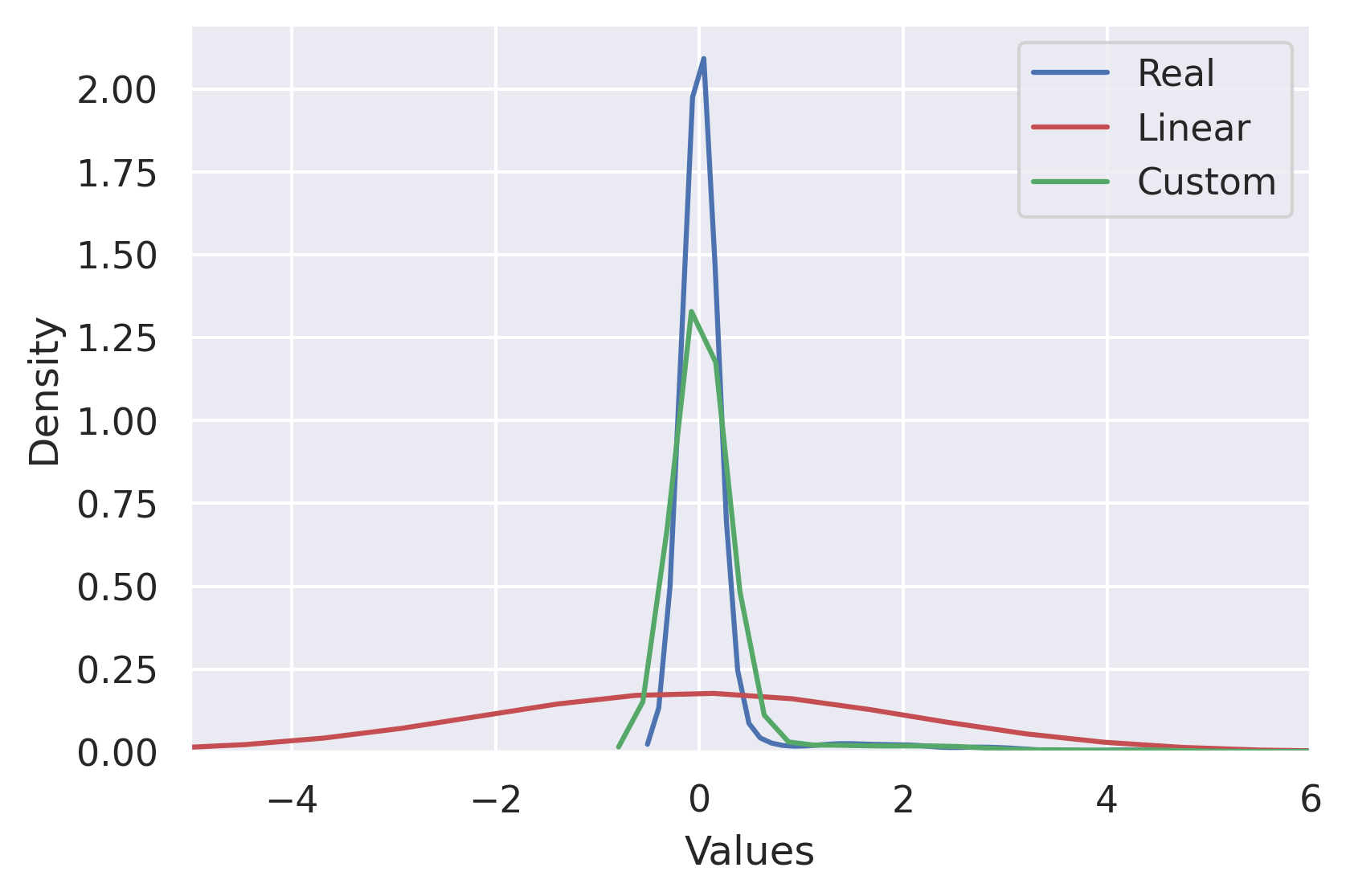} 
\caption{Feature 1.}\label{fig:CR-distr-kde-l0-f1}
\end{subfigure}
\begin{subfigure}[t]{.24\textwidth}
\includegraphics[width=1\linewidth]{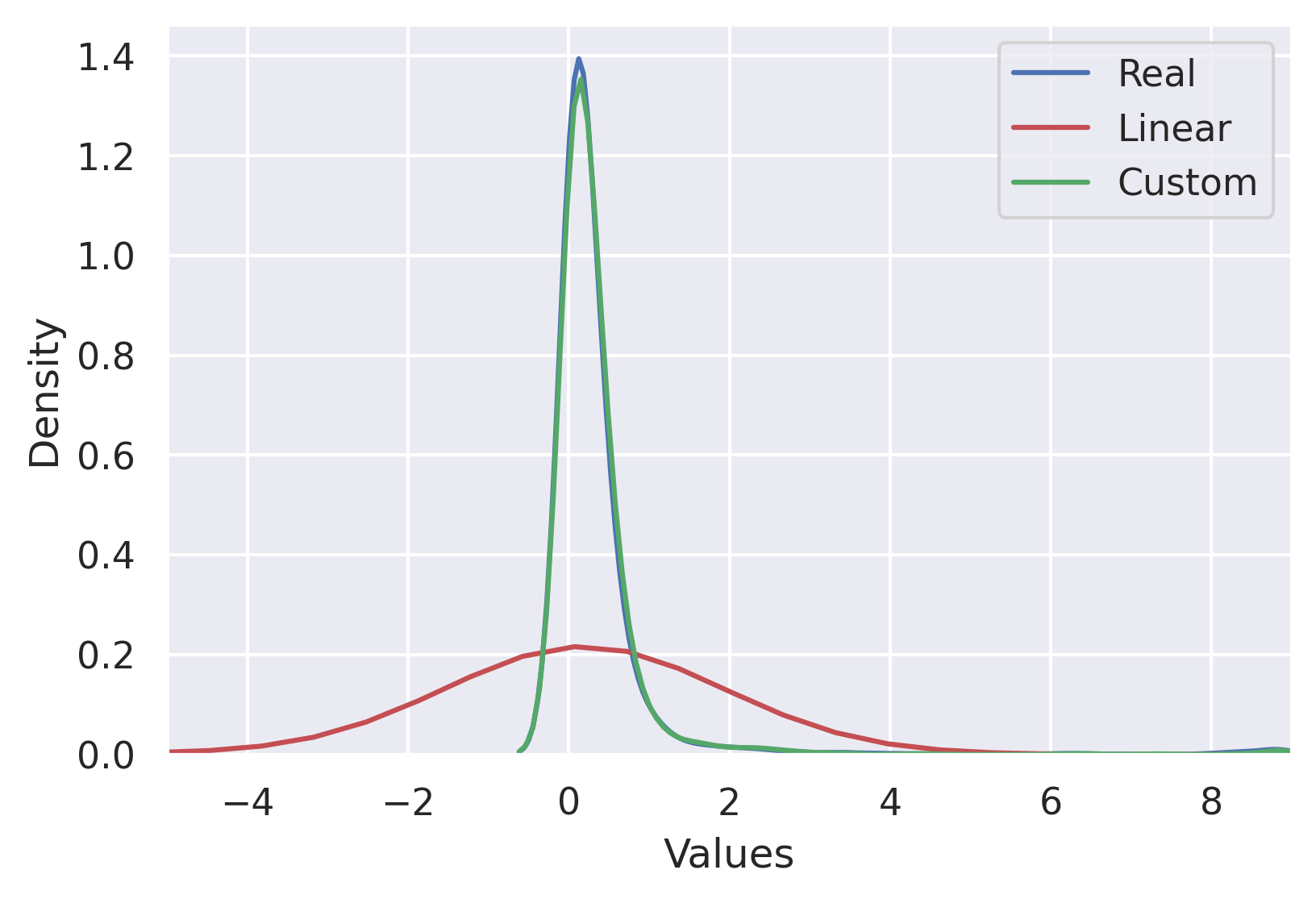} 
\caption{Feature 2.}\label{fig:CR-distr-kde-l0-f2}
\end{subfigure}
\begin{subfigure}[t]{.24\textwidth}
\includegraphics[width=1\linewidth]{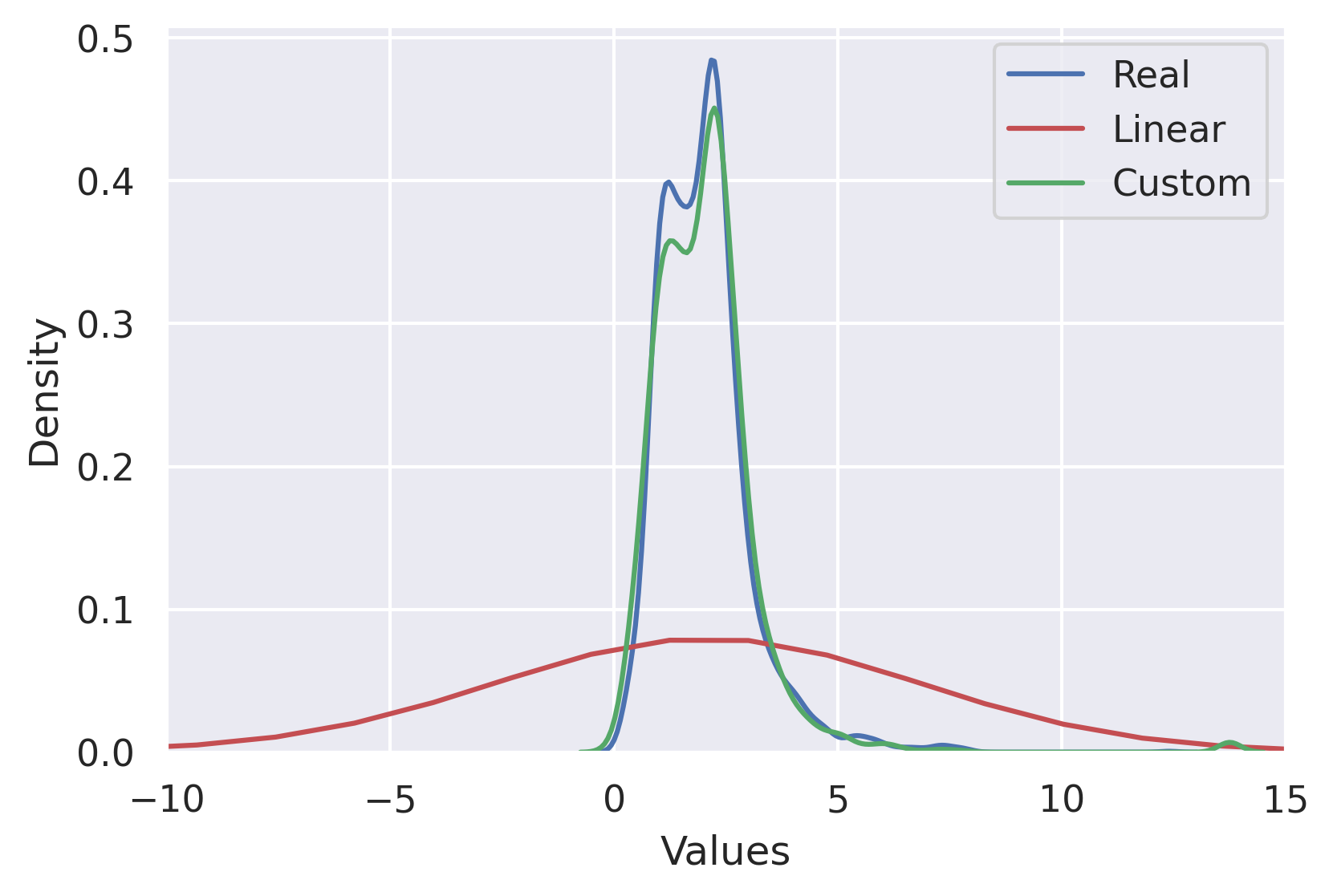} 
\caption{Feat 1. Real data.}\label{fig:CR-distr-kde-l0-f3}
\end{subfigure}

\caption{Cryptomining data set (second use case). Frequency distribution from Label $0$ of real data and standard and ST-based WGAN's synthetic data. Histogram (top row) and Kernel Density Estimator (KDE) function (bottom row) are shown for the four variables.}
\label{fig:CR-distr-l0}
\end{figure*}


\begin{figure*}[!t]
\centering
\begin{subfigure}[t]{.24\textwidth}
\includegraphics[width=1\linewidth]{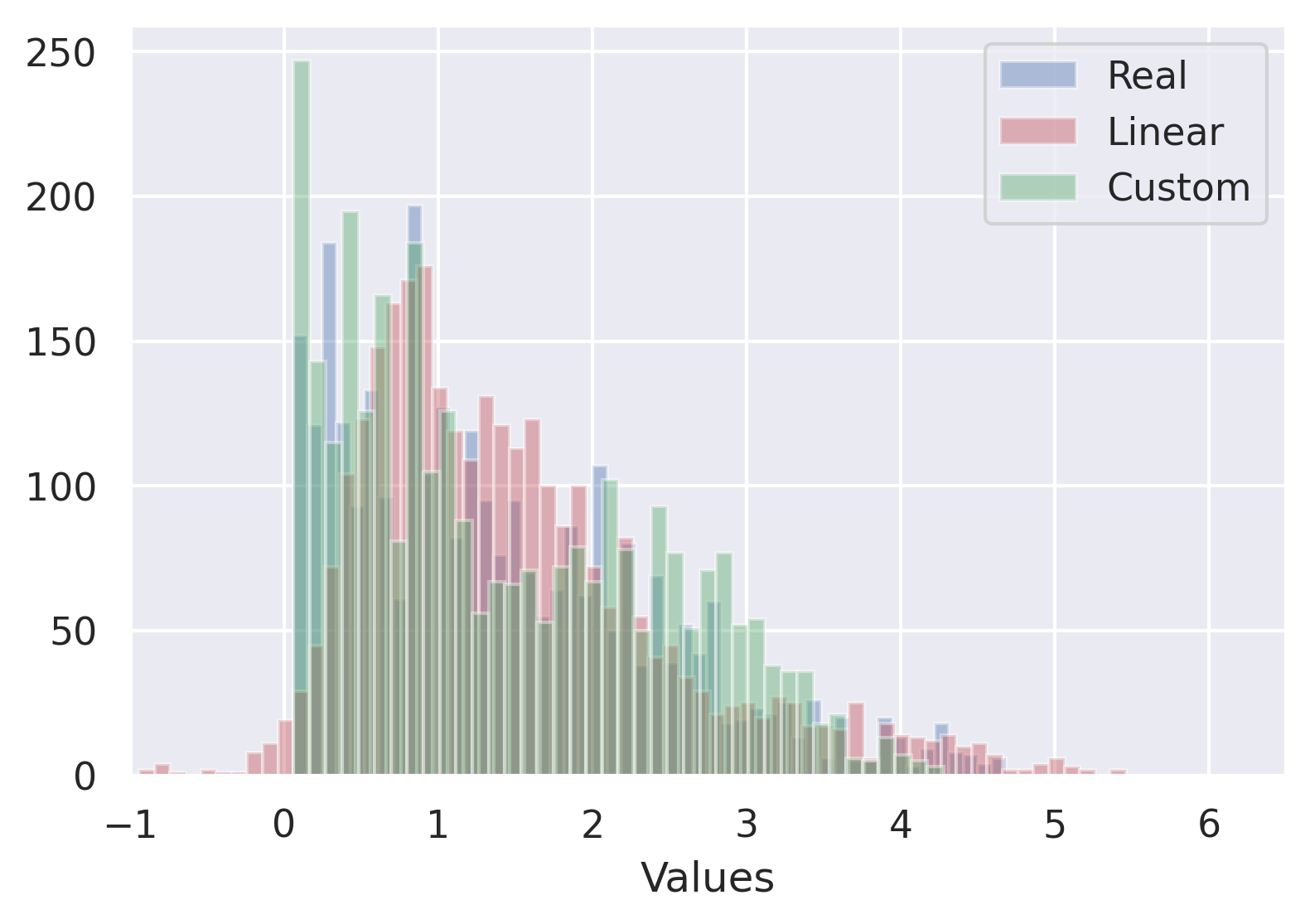} 
\caption{Feature 0.}\label{fig:CR-distr-l1-f0}
\end{subfigure}
\begin{subfigure}[t]{.24\textwidth}
\includegraphics[width=1\linewidth]{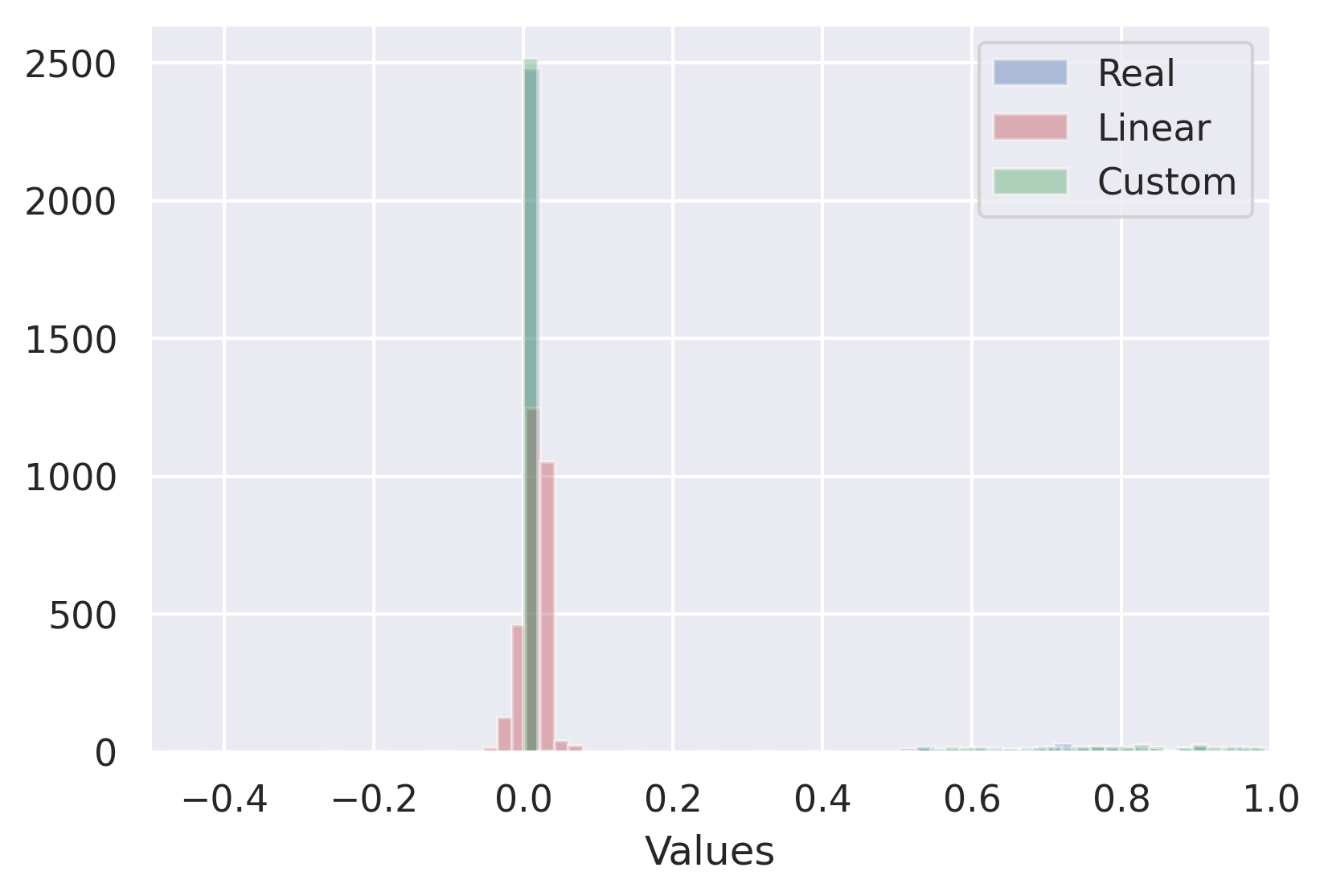} 
\caption{Feature 1.}\label{fig:CR-distr-l1-f1}
\end{subfigure}
\begin{subfigure}[t]{.24\textwidth}
\includegraphics[width=1\linewidth]{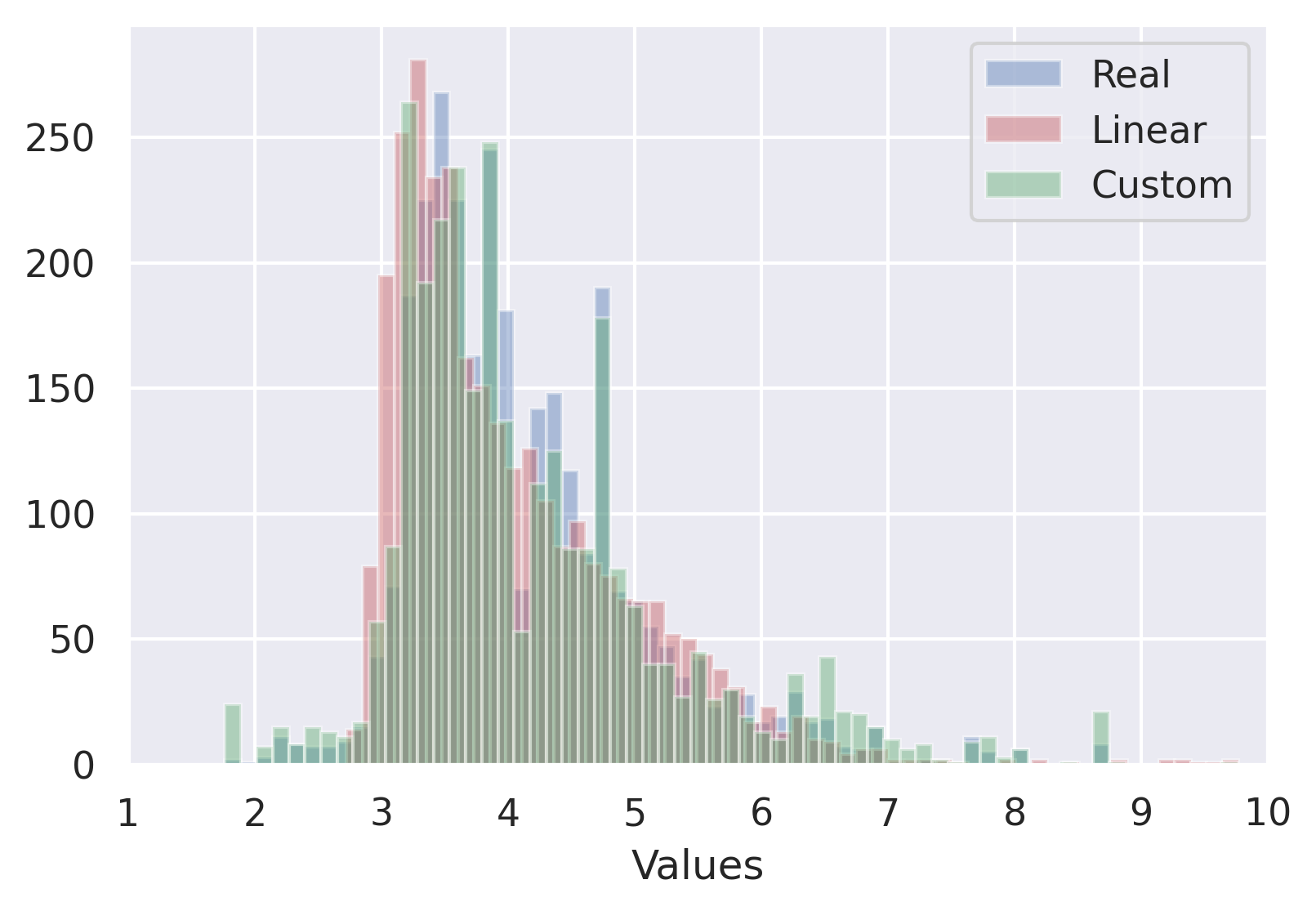} 
\caption{Feature 2.}\label{fig:CR-distr-l1-f2}
\end{subfigure}
\begin{subfigure}[t]{.24\textwidth}
\includegraphics[width=1\linewidth]{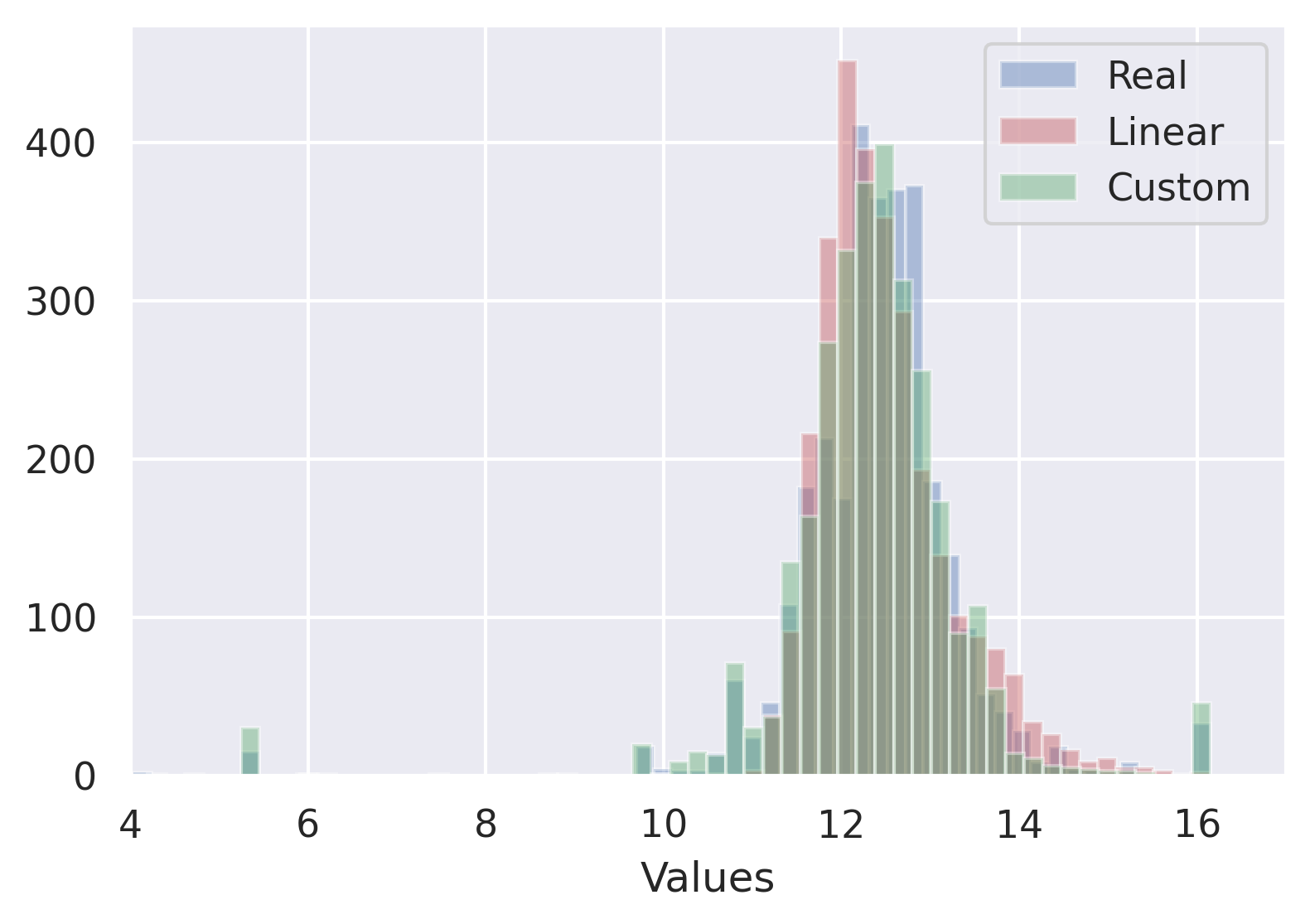} 
\caption{Feat 1. Real data.}\label{fig:CR-distr-l1-f3}
\end{subfigure}

\begin{subfigure}[t]{.24\textwidth}
\includegraphics[width=1\linewidth]{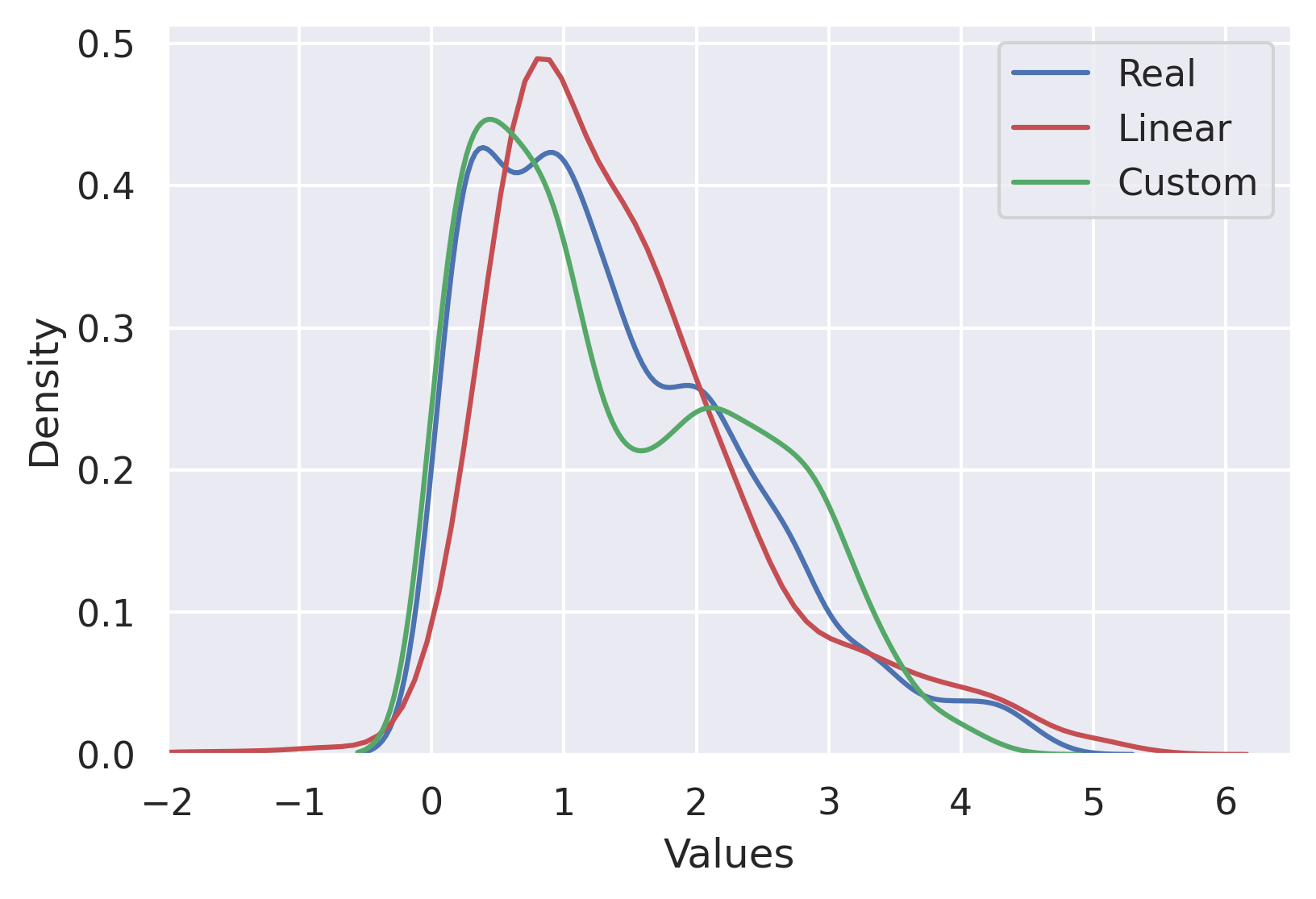} 
\caption{Feature 0.}\label{fig:CR-distr-kde-l1-f0}
\end{subfigure}
\begin{subfigure}[t]{.24\textwidth}
\includegraphics[width=1\linewidth]{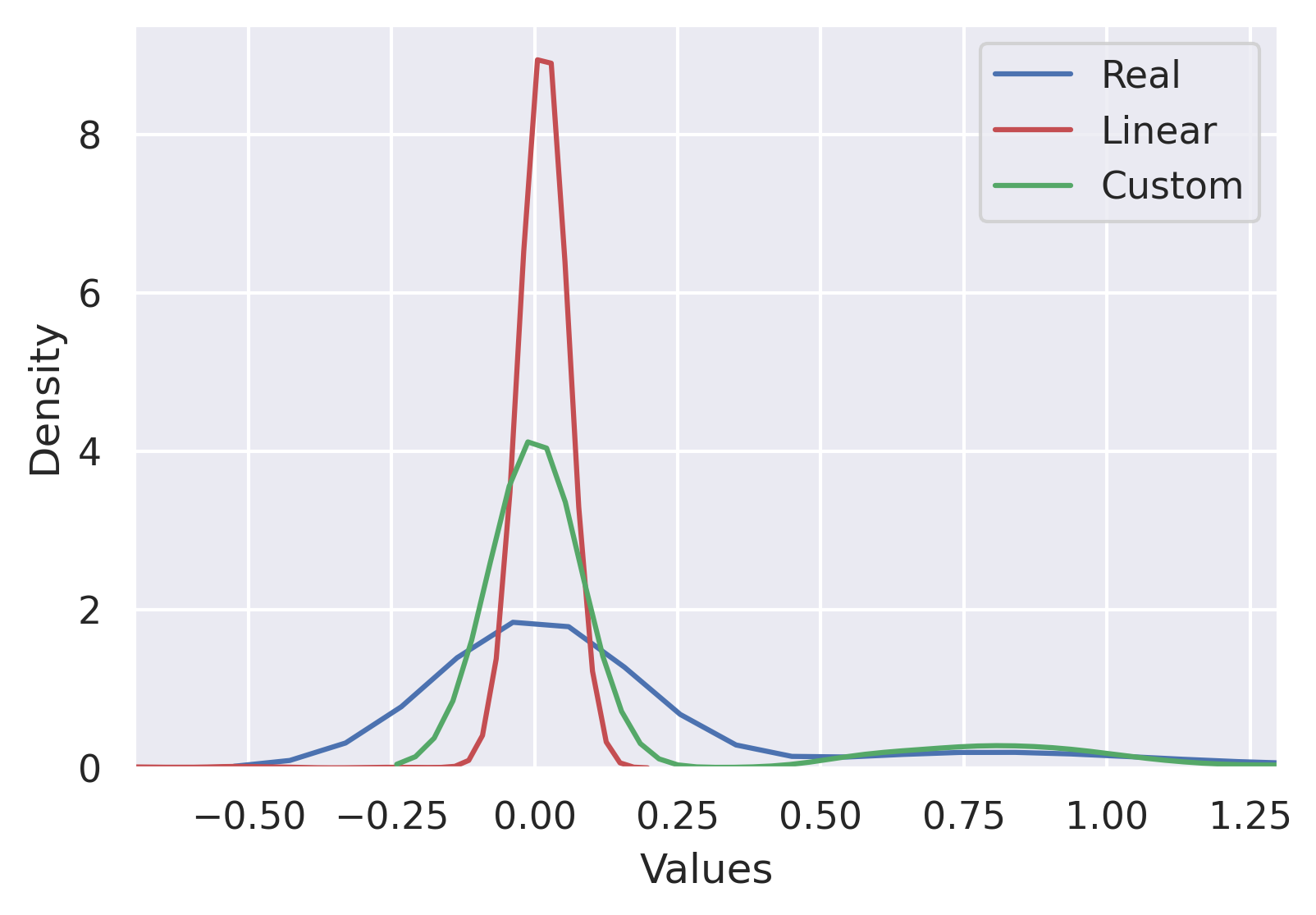} 
\caption{Feature 1.}\label{fig:CR-distr-kde-l1-f1}
\end{subfigure}
\begin{subfigure}[t]{.24\textwidth}
\includegraphics[width=1\linewidth]{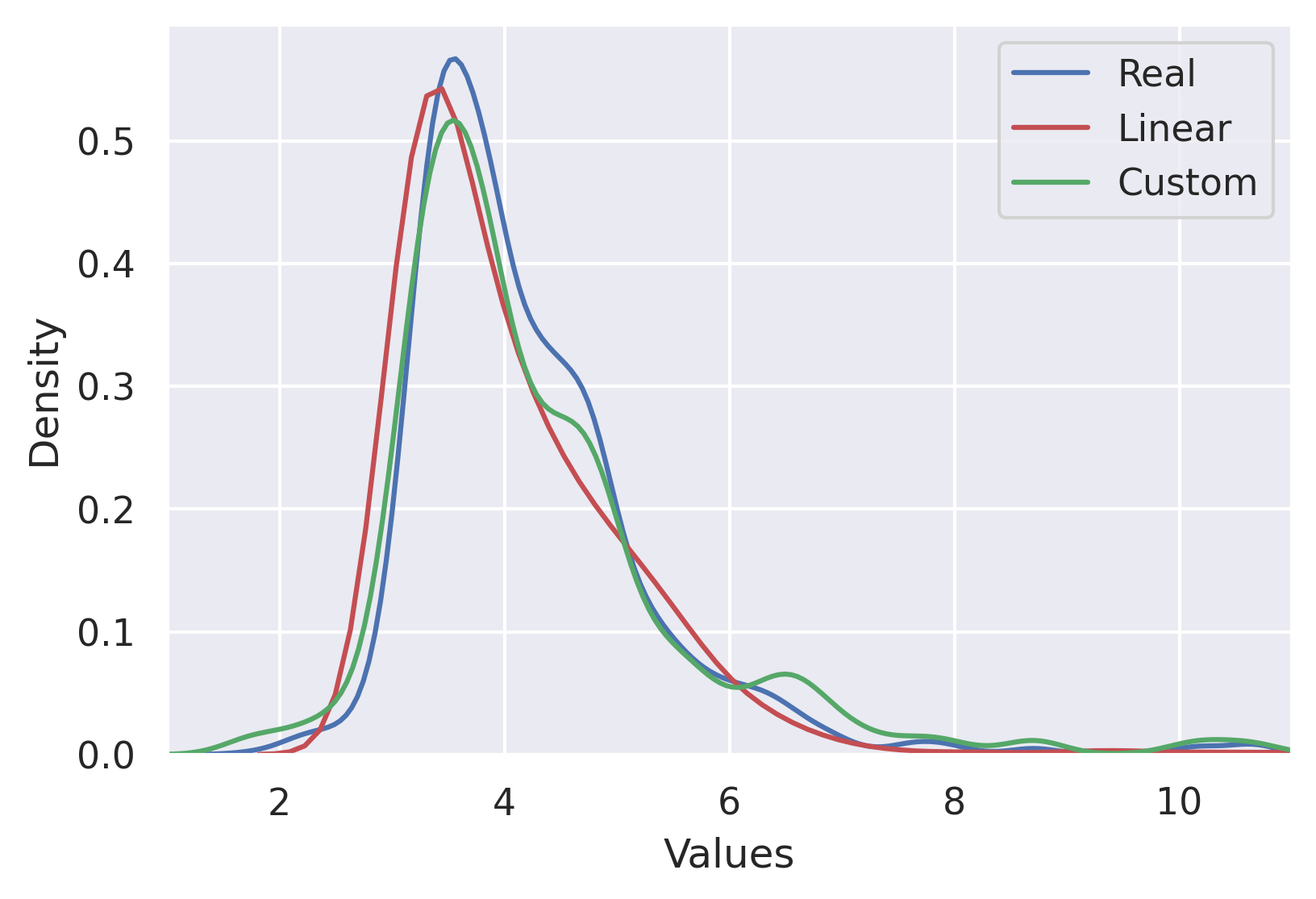} 
\caption{Feature 2.}\label{fig:CR-distr-kde-l1-f2}
\end{subfigure}
\begin{subfigure}[t]{.24\textwidth}
\includegraphics[width=1\linewidth]{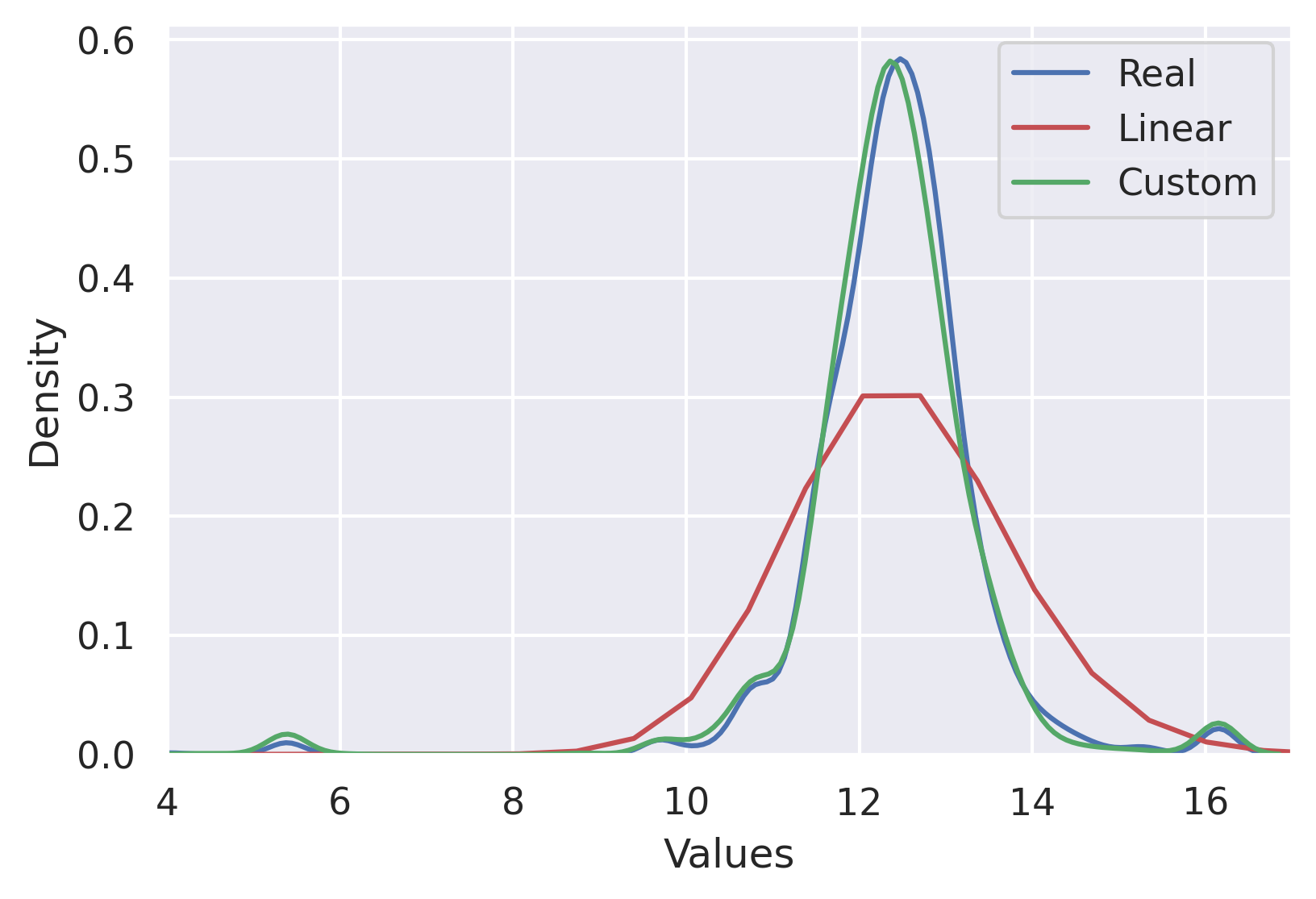} 
\caption{Feat 1. Real data.}\label{fig:CR-distr-kde-l1-f3}
\end{subfigure}

\caption{Cryptomining data set (second use case). Frequency distribution from Label $1$ of real data and standard and ST-based WGAN's synthetic data. Histogram (top row) and Kernel Density Estimator (KDE) function (bottom row) are shown for the four variables.}
\label{fig:CR-distr-l1}
\end{figure*}

\paragraph {Feature replication}
After having analysed in the previous subsections the quality of the synthetic data generated by ST-based WGANs,  we study the fidelity with which WGAN generators replicate the statistical distribution of each variable.  
We compare the statistical distributions of each of the 4 variables that compose the data elements in the first use case
in Figure \ref{fig:SYN-distr-l0} (label $0$) and Figure \ref{fig:SYN-distr-l1} (label $1$), and for the second use case in Figure \ref{fig:CR-distr-l0} (label $0$) and Figure \ref{fig:CR-distr-l1} (label $1$). 
At the top row of each figure, we plot the histogram of the real variables and the synthetic variables generated by the standard and ST-based WGANs. 
The bottom row shows the Kernel Density Estimator (KDE \cite{parzen1962estimation}) function of all of them.
Synthetic data samples were obtained at epoch $25$.

In general, it can be observed in all figures that the ST-based WGAN replicates the histogram and the KDE function of each variable with high accuracy. On the contrary, for the two uses cases and the two labels, the standard WGAN tends not to mimic the statistical distribution of the real data. 

It is worth noting that one of the key innovations of our proposal is the ability of ST-based WGANs to replicate discrete data variables. To the best of our knowledge, none of the existing solutions provides a clean approach to solve this crucial issue as the ST-based approach does.
In the first use case, an ad-hoc data distribution was designed to contain two discrete variables (features $1$ and $3$) in each label to highlight the advantage of our solution in comparison with current solutions.
It can be observed in the Label $0$ histograms for features $1$ and $3$, (Figure \ref{fig:SYN-distr-l0-f1} and Figure \ref{fig:SYN-distr-l0-f3}) that the ST-based WGAN perfectly replicates the discrete nature of these features. In sharp contrast, the standard WGAN fails on this task and generates two synthetic variables that follow a continuous distribution. The same situation appears for label $1$, as variables $1$ and $3$ (Figure \ref{fig:SYN-distr-l1-f1} and Figure \ref{fig:SYN-distr-l1-f3}) are perfectly replicated by the ST-based WGAN and on the contrary, the standard WGAN generates synthetic variables following a continuous distribution.

As a final remark, note that the discrete variables can be categorical or not (e.g., an ordered sequence of integer numbers) as the ST solution does not impose any assumption on this and therefore, provides a general solution for any discrete variable.

\paragraph{Marginal $F_1$-score}

\begin{figure*}[!t]

%
\begin{subfigure}[t]{.245\textwidth}
\centering
\includegraphics[width=1\linewidth]{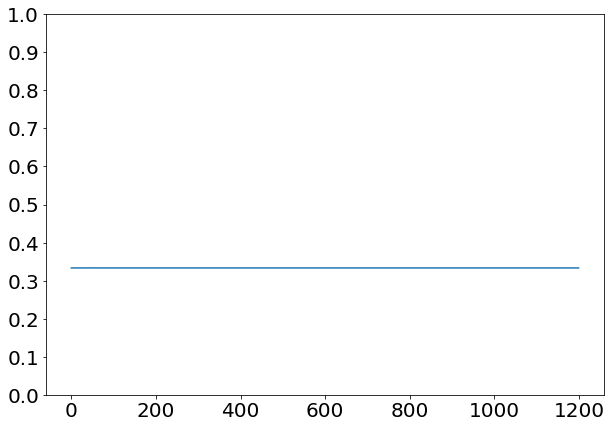} 
\caption{Linear: DS1 - L0}
\label{fig:syn-vainilla-f1-tr-l0}
\end{subfigure}
\begin{subfigure}[t]{.245\textwidth}
\centering
\includegraphics[width=1\linewidth]{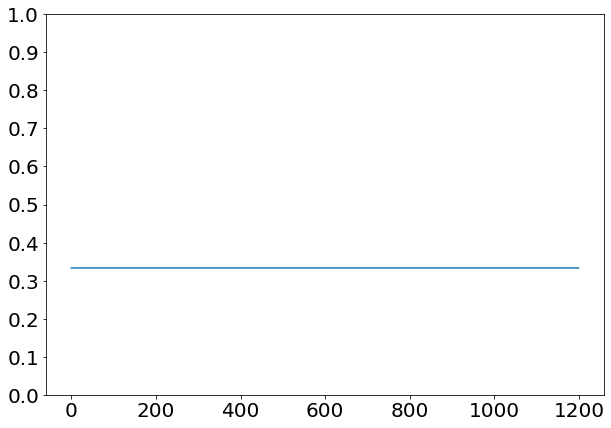} 
\caption{Linear: DS2 - L0}
\label{fig:syn-vainilla-f1-test-l0}
\end{subfigure}
%
\begin{subfigure}[t]{.245\textwidth}
\centering
\includegraphics[width=1\linewidth]{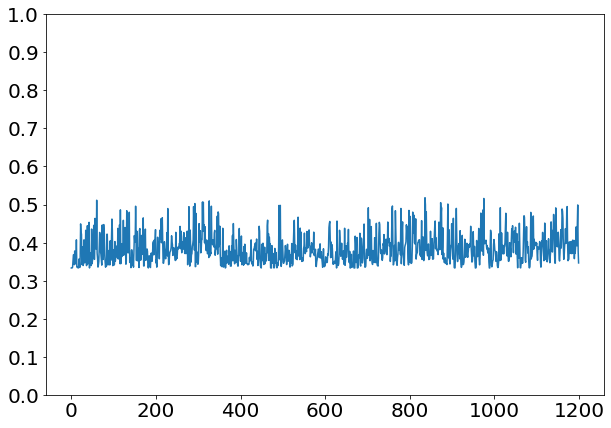} 
\caption{Linear: DS1 - L1}
\label{fig:syn-vainilla-f1-tr-l1}
\end{subfigure}
\begin{subfigure}[t]{.245\textwidth}
\centering
\includegraphics[width=1\linewidth]{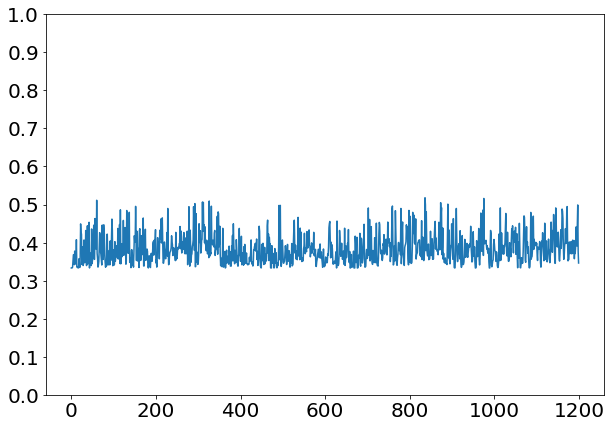} 
\caption{Linear: DS2 - L1}
\label{fig:syn-vainilla-f1-test-l1}
\end{subfigure}

\medskip


\begin{subfigure}[t]{.245\textwidth}
\centering
\includegraphics[width=1\linewidth]{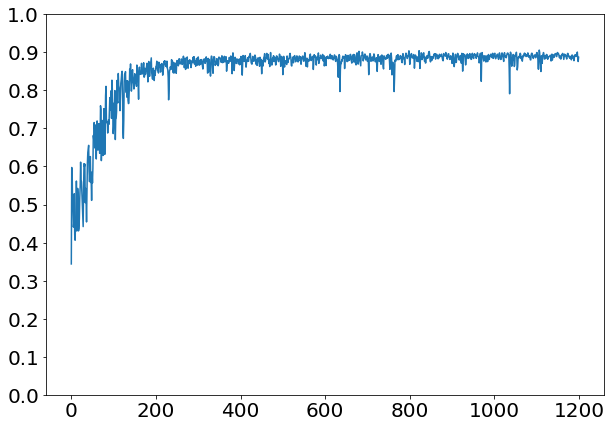} 
\caption{Custom: DS1 - L0}
\label{fig:syn-custom-f1-tr-l0}
\end{subfigure}
\begin{subfigure}[t]{.245\textwidth}
\centering
\includegraphics[width=1\linewidth]{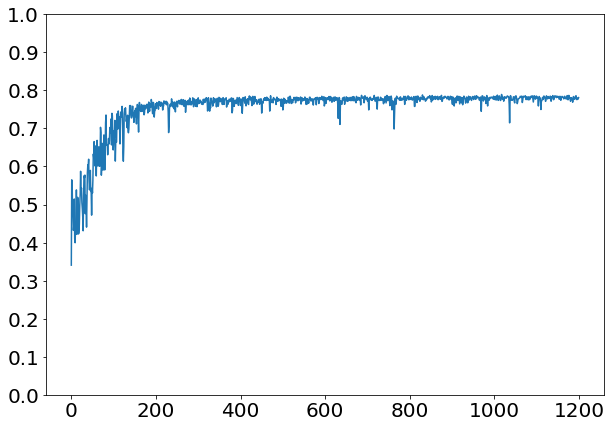} 
\caption{Custom: DS2 - L0}
\label{fig:syn-custom-f1-test-l0}
\end{subfigure}
%
\begin{subfigure}[t]{.245\textwidth}
\centering
\includegraphics[width=1\linewidth]{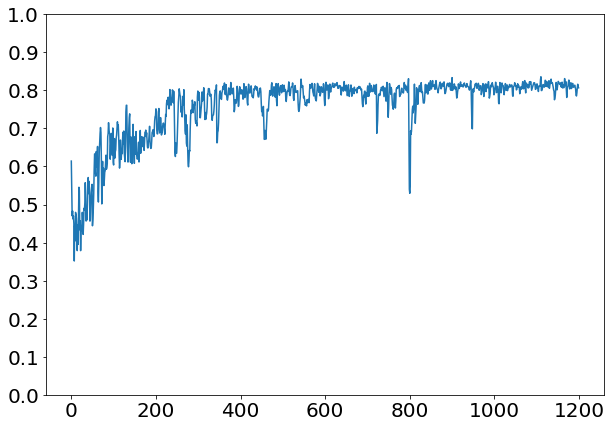} 
\caption{Custom: DS1 - L1}
\label{fig:syn-custom-f1-tr-l1}
\end{subfigure}
\begin{subfigure}[t]{.245\textwidth}
\centering
\includegraphics[width=1\linewidth]{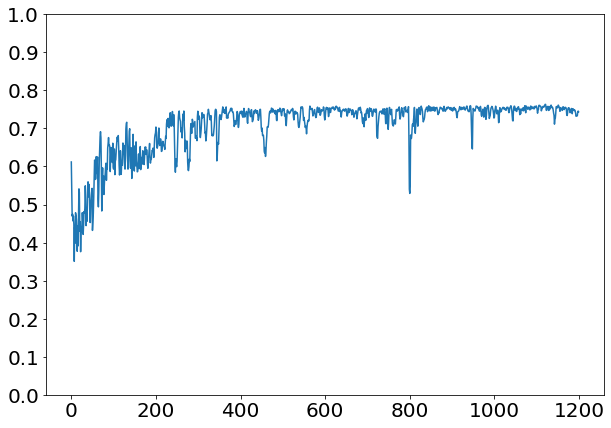} 
\caption{Custom: DS2 - L1}
\label{fig:syn-custom-f1-test-l1}
\end{subfigure}

\caption{Rendered data (first use case). Evolution of the marginal $F_1$-score on training and testing, using GAN generators with linear activation (top row) and IST-based activation (bottom row) for labels 0 and 1. The first and second columns correspond to the evolution for the GAN trained to generate label 0, whereas the third and forth columns correspond to the generation of label 1. The $x$-axis represents the GAN training epochs.}
\label{fig:SYN_F1}
\end{figure*}

Looking at the marginal $F_1$-score values obtained for each label in the first use case (Figure \ref{fig:SYN_F1}), we observe that the standard WGAN generators for both labels fail completely when they are used to marginally replace real data in training an ML classifier, as they obtain very bad  $F_1$-score metrics (around $0.35$ for label $0$ and $0.4$ for label $1$) in both testing (DS2-r) and training (DS1-r) data sets.
On the contrary, the ST-based WGAN obtains $F_1$-score values close to the ones obtained using real data in the training of the ML classifier. The benchmark classifier trained with real data obtained an $F_1$-score of $0.812$ with DS2-r and using the synthetic data generated by the ST-based WGAN we obtained  around $0.75$ for label $0$ and $0.7$ for label $1$ with DS2-r as testing data set. Furthermore, the ST-based WGAN did not generate any significant oscillations in the $F_1$-score curve after the maximum $F_1$-score values were reached at $4$ and $8$ epochs respectively for labels $0$ and $1$.

In general, the results of the marginal $F_1$-score for the second use case aligned with those obtained in the first use case. When the ST-based WGAN was used with DS2-c as the testing data set, higher $F_1$-score values were achieved (around $0.6$ for label $0$ and $0.85$ for label $1$) without incurring significant oscillations when GAN training was estabilised (around $8$ epochs for both labels). In contrast, the standard WGAN performance was not good, with small $F_1$-score values for label $0$ (from $0.1$ to $0.5$) and noticeable oscillations throughout the GAN training for both labels.  

We can conclude that in both use cases, the ST-based WGAN  obtained better marginal $F_1$-score values, stabilised to the maximum values faster and without producing significant oscillations once stabilised.


\begin{figure*}[!t]

%
\begin{subfigure}[t]{.245\textwidth}
\centering
\includegraphics[width=1\linewidth]{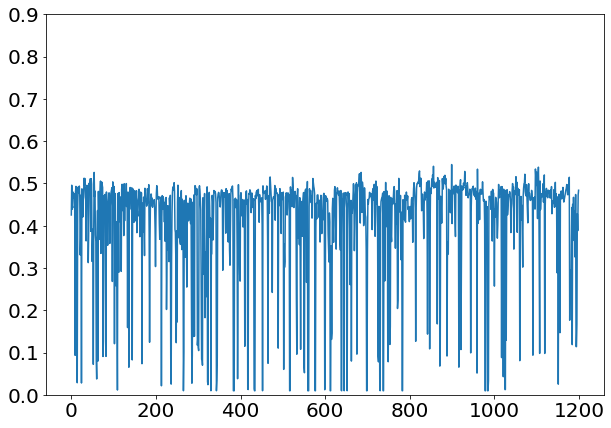} 
\caption{Linear: DS1 - L0}
\label{fig:vainilla-f1-tr-l0}
\end{subfigure}
\begin{subfigure}[t]{.245\textwidth}
\centering
\includegraphics[width=1\linewidth]{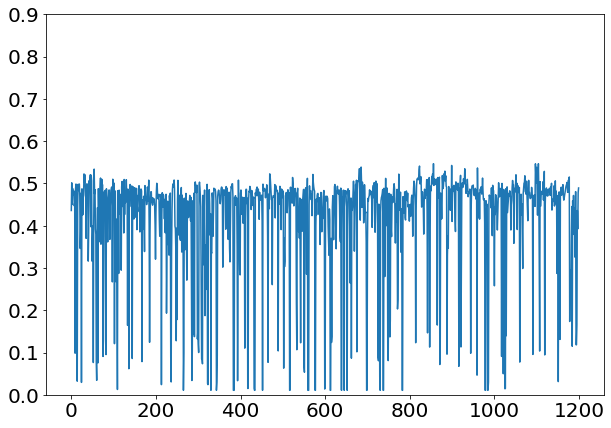} 
\caption{Linear: DS2 - L0}
\label{fig:vainilla-f1-test-l0}
\end{subfigure}
%
\begin{subfigure}[t]{.245\textwidth}
\centering
\includegraphics[width=1\linewidth]{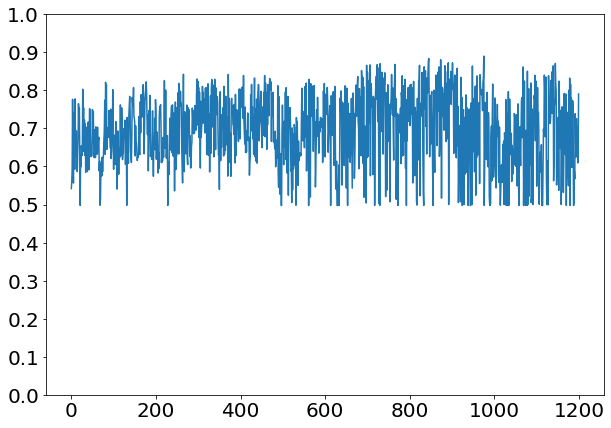} 
\caption{Linear: DS1 - L1}
\label{fig:vainilla-f1-tr-l1}
\end{subfigure}
\begin{subfigure}[t]{.245\textwidth}
\centering
\includegraphics[width=1\linewidth]{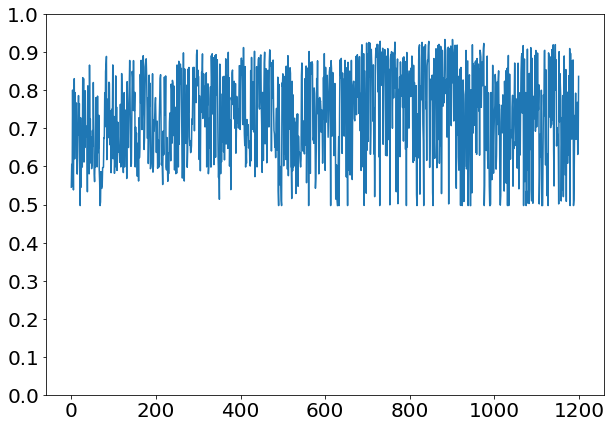} 
\caption{Linear: DS2 - L1}
\label{fig:vainilla-f1-test-l1}
\end{subfigure}

\medskip


\begin{subfigure}[t]{.245\textwidth}
\centering
\includegraphics[width=1\linewidth]{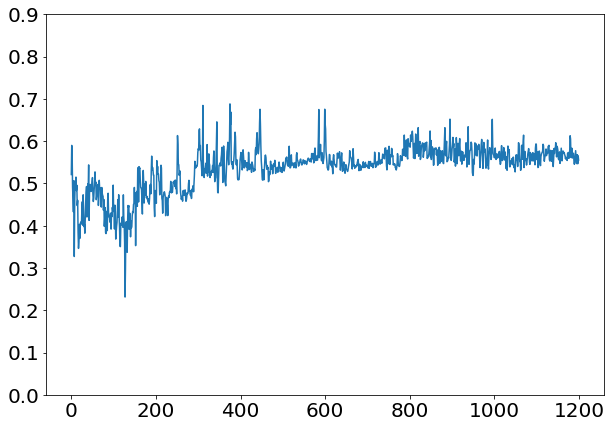} 
\caption{Custom: DS1 - L0}
\label{fig:custom-f1-tr-l0}
\end{subfigure}
\begin{subfigure}[t]{.245\textwidth}
\centering
\includegraphics[width=1\linewidth]{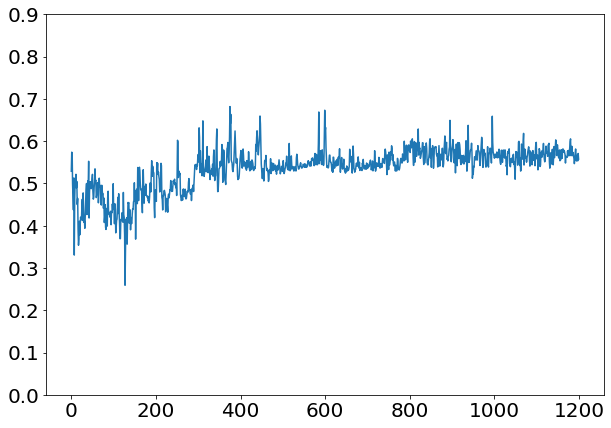} 
\caption{Custom: DS2 - L0}
\label{fig:custom-f1-test-l0}
\end{subfigure}
%
\begin{subfigure}[t]{.245\textwidth}
\centering
\includegraphics[width=1\linewidth]{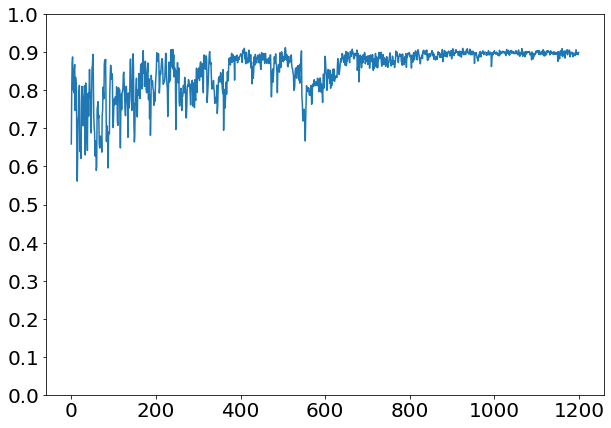} 
\caption{Custom: DS1 - L1}
\label{fig:custom-f1-tr-l1}
\end{subfigure}
\begin{subfigure}[t]{.245\textwidth}
\centering
\includegraphics[width=1\linewidth]{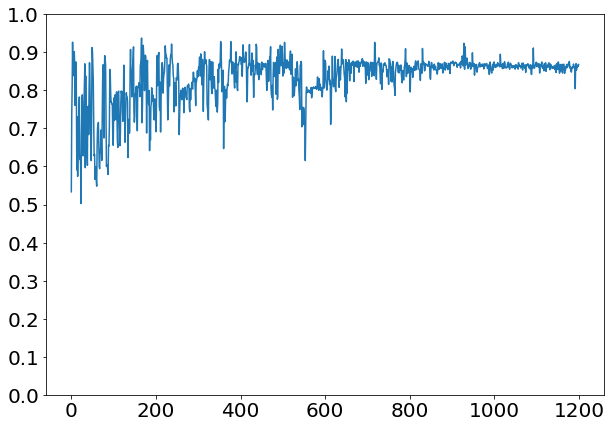} 
\caption{Custom: DS2 - L1}
\label{fig:custom-f1-test-l1}
\end{subfigure}

\caption{Cryptomining attack scenario (second use case). Evolution of the marginal $F_1$-score on training and testing, using GAN generators with linear activation (top row) and IST-based activation (bottom row) for labels 0 and 1. The first and second columns correspond to the evolution for the GAN trained to generate label 0, whereas the third and forth columns correspond to the generation of label 1. The $x$-axis represents the GAN training epochs.}
\label{fig:CR_F1}
\end{figure*}

\paragraph{Nested ML evaluation}

%
In this section, we analyze the performance of the nested ML classifier when a fully synthetic data set is generated using GANs. Recall that this data set is created by mixing samples created by the GAN generators of labels $0$ and $1$, and it is used to feed a ML classifier. As criteria for picking the generators to be used for each label, we applied two different strategies: (i) drawing a random sample among the whole set of stored generators, and (ii) drawing a random sample from the top $10$ models sorted by the marginal F1-score of the label (i.e., using $F_1$-score elitism).

\begin{figure*}[!ht]
\begin{mdframed}
\centering

\begin{subfigure}[t]{0.99\textwidth}
\centerline{
  \includegraphics[width=0.49\linewidth]{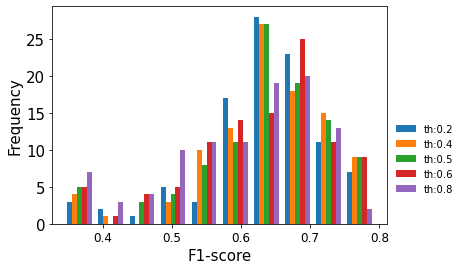} 
  \includegraphics[width=0.49\linewidth]{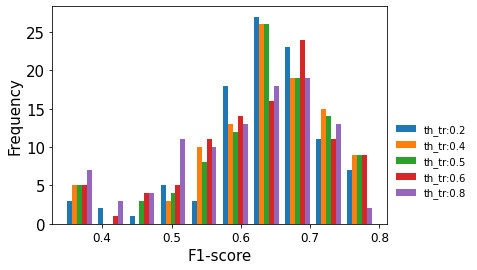}}
\caption{Generator with linear activation.}\label{fig:Mix_SYN_FA_L}
\end{subfigure}

\medskip

\begin{subfigure}[t]{.99\textwidth}
\centering
\vspace{0pt}
\includegraphics[width=0.49\linewidth]{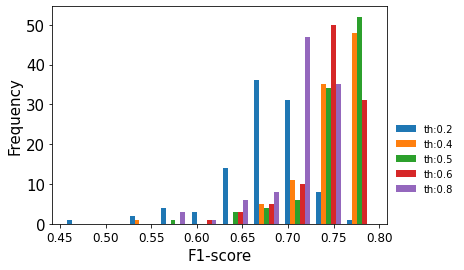}
\hfill
\includegraphics[width=0.49\linewidth]{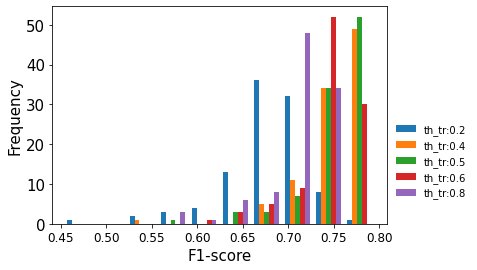}
\caption{Generator with custom activation.}\label{fig:Mix_SYN_FA_S}
\end{subfigure}

\medskip

\begin{subfigure}[t]{.99\textwidth}
\centering
\vspace{0pt}
\includegraphics[width=0.49\linewidth]{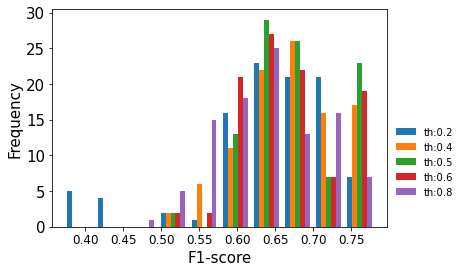}
\hfill
\includegraphics[width=0.49\linewidth]{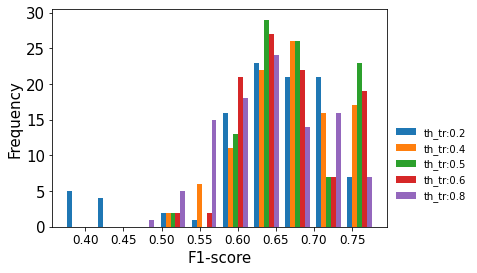}
\caption{Generator with linear activation. Elitism criterion: Top 10 on F1-score.}\label{fig:Mix_top10_SYN_FA_L}
\end{subfigure}

\medskip

\begin{subfigure}[t]{.99\textwidth}
\centering
\vspace{0pt}
\includegraphics[width=0.49\linewidth]{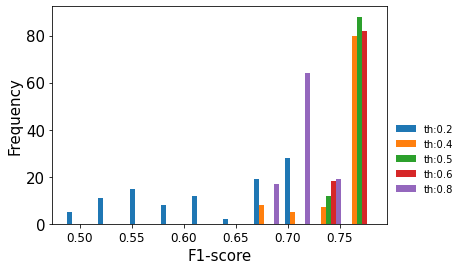}
\hfill
\includegraphics[width=0.49\linewidth]{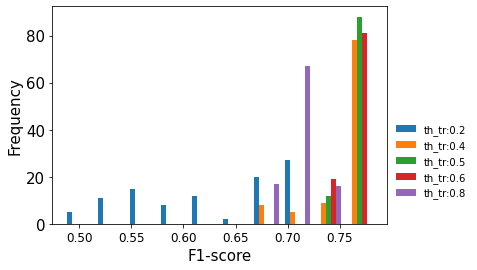}
\caption{Generator with custom activation. Elitism criterion: Top 10 on F1-score.}\label{fig:Mix_top10_SYN_FA_S}
\end{subfigure}

\end{mdframed}
\medskip
\begin{minipage}[t]{.99\textwidth}
\caption{Rendered data set (first use case). $F_1$-score on DS2-r (left) and DS1-r (right) using as training dataset GAN generators with linear and  IST-based activation. Random selection of generators was done from among all generators (\ref{fig:Mix_SYN_FA_L} and \ref{fig:Mix_SYN_FA_S}) and using an elitism criterion (top 10 of F1-scores) (\ref{fig:Mix_top10_SYN_FA_L} and \ref{fig:Mix_top10_SYN_FA_S}). Results for decision thresholds of 0.2, 0.4, 0.5, 0.6 and 0.8 are represented. Each experiment was run 100 times.} \label{fig:Mix_SYN}
\end{minipage}

\end{figure*}

Table \ref{tab:SYN_mix} summarises the results obtained for the first use case and Figure \ref{fig:Mix_SYN} shows detailed histograms of the $F_1$-score results obtained after running 100 times each experiment. In the light of the results presented in these plots, the main conclusions are: 
\begin{enumerate}[(i)]
    \item Both the ST-based and standard WGANs can be used for training a ML classifier as their performance is better than that of the noise generator and only slightly worse than that of the real data.
    \item The synthetic data generated by the ST-based WGAN performs slightly better than that of the standard WGAN, both when random selection is done on the whole set of models ($0.793$ against $0.789$ for the best $F_1$-score obtained) and when the $F_1$-score elitism is used ($0.783$ against $0.775$ for the best $F_1$-score obtained).
    \item The interval of the $F_1$-score values obtained when using the ST-based WGAN (from $0.75$ to $079$) was concentrated near to the maximum value and was significantly smaller than that obtained with the standard WGAN that are noticeably dispersed in a larger interval from $0.5$ to $0.79$. Hence, selecting at random ST-based generator models is highly likely to obtain a synthetic data set that performs close to the best model combination and the real data. In contrast, if we select generators at random from the standard WGAN, it is less likely to obtain a synthetic data set that comes close to the performance of the real data.
    \item When the ST-based activation was used, although the maximum value obtained with $F_1$-score elitism was slightly lower than that obtained with the uniform drawing method, the distribution of results was slightly more concentrated near the maximum value.
\end{enumerate}

\begin{table}
\parbox{1.\linewidth}{
\caption{
Rendered data set (first use case). Performance of synthetic traffic combining labels 0 and 1. Training with (i) a real data set (R-DS1), (ii) a mean-based noise generator and (iii and iv) GAN synthetic data (with linear and ST activation). Results on testing with real data (R-DS2). Experiment is drawn 100 times uniformly at random (Figure \ref{fig:Mix_SYN_B} and Figure \ref{fig:Mix_SYN}). For the GANs, in each sample we choose one generator from among all label 0 generators and one generator from among all label 1 generators.}
\label{tab:SYN_mix}
\centering

\resizebox{.99\linewidth}{!}{%
\begin{tabular}{|c|c|c|c|}
\hline \textbf{Dataset} & \textbf{Quality Measure} & \textbf{Best} & \textbf{Default} \\\hline\hline

\multirow{3}{*}{\vspace{-0.3cm}$\begin{matrix}\textbf{Training 200K/200K }\\\textbf{Real data (R-DS1)}\end{matrix}$}  & \textit{{Threshold}} & 0.5 & 0.5 \\ 
\cline{2-4} & \textit{{$F_1$-score}} & 0.812  &  0.812\\ \cline{2-4}
 & $\begin{matrix}\textit{{Confusion }} \\ \textit{{matrix}}\end{matrix}$ & \begin{tabular}{c|c} 349534 & 50466 \\\hline 99271 & 300729\end{tabular} & \begin{tabular}{c|c} 349534 & 50466\\\hline 99271 & 300729 \end{tabular}\\ \hline\hline
 
 \multirow{3}{*}{\vspace{-0.3cm}$\begin{matrix}\textbf{Training 200K/200K }\\\textbf{Noise generator with means}\end{matrix}$}  & \textit{{Threshold}} & 0.2 & 0.5 \\ 
\cline{2-4} & \textit{{$F_1$-score}} & 0.764 & 0.731 \\ \cline{2-4}
 & $\begin{matrix}\textit{{Confusion }} \\ \textit{{matrix}}\end{matrix}$ & \begin{tabular}{c|c} 349292 &  50708 \\\hline 135673 & 264327 \end{tabular} & \begin{tabular}{c|c} 379238 & 20762 \\\hline 185058 & 214942 \end{tabular}\\ \hline\hline
 
\multirow{3}{*}{\vspace{-0.3cm}$\begin{matrix}\textbf{Training 200K/200K }\\\textbf{WGAN with linear activation}\end{matrix}$}  & \textit{{Threshold}} & 0.5 & 0.5  \\ 
\cline{2-4} & \textit{{$F_1$-score}} & 0.789 & 0.789 \\ \cline{2-4}
 & $\begin{matrix}\textit{{Confusion }} \\ \textit{{matrix}}\end{matrix}$ & \begin{tabular}{c|c} 357013 & 42987 \\\hline 123978 & 276022 \end{tabular} & \begin{tabular}{c|c} 357013 & 42987 \\\hline 123978 & 276022 \end{tabular}\\ \hline\hline
 
 \multirow{3}{*}{\vspace{-0.3cm}$\begin{matrix}\textbf{Training 200K/200K }\\\textbf{WGAN with ST activation}\end{matrix}$}  & \textit{{Threshold}} & 0.4 & 0.5 \\ 
\cline{2-4} & \textit{{$F_1$-score}} & 0.795  & 0.793 \\ \cline{2-4}
 & $\begin{matrix}\textit{{Confusion }} \\ \textit{{matrix}}\end{matrix}$ & \begin{tabular}{c|c}338960 & 61040 \\\hline 102240 & 297760  \end{tabular} & \begin{tabular}{c|c} 333288 & 66712 \\\hline 98781 & 301219 \end{tabular}\\ \hline \hline
 
 \multirow{3}{*}{\vspace{-0.3cm}$\begin{matrix}\textbf{Training 200K/200K }\\\textbf{WGAN with linear activation}\\\textbf{$F_1$-score elitism (top 10)}\end{matrix}$}  & \textit{{Threshold}} & 0.4 & 0.5 \\ 
\cline{2-4} & \textit{{$F_1$-score}} & 0.779  & 0.775 \\ \cline{2-4}
 & $\begin{matrix}\textit{{Confusion }} \\ \textit{{matrix}}\end{matrix}$ & \begin{tabular}{c|c} 346911 &  53089 \\\hline 122176 & 277824 \end{tabular} & \begin{tabular}{c|c} 339251 & 60749 \\\hline 118428 & 281572 \end{tabular}\\ \hline \hline

\multirow{3}{*}{\vspace{-0.3cm}$\begin{matrix}\textbf{Training 200K/200K }\\\textbf{WGAN with ST activation}\\\textbf{$F_1$-score elitism (top 10)}\end{matrix}$}  & \textit{{Threshold}} & 0.5 & 0.5 \\ 
\cline{2-4} & \textit{{$F_1$-score}} & 0.783 & 0.783\\ \cline{2-4}
 & $\begin{matrix}\textit{{Confusion }} \\ \textit{{matrix}}\end{matrix}$ & \begin{tabular}{c|c} 339440 & 60560 \\\hline 111704 & 288296 \end{tabular} & \begin{tabular}{c|c} 339440 & 60560 \\\hline 111704 & 288296 \end{tabular}\\ \hline

\end{tabular}
}
}
\end{table}

\begin{figure*}[!ht]
\begin{mdframed}
\centering

\begin{subfigure}[t]{0.99\textwidth}
\centerline{
  \includegraphics[width=0.49\linewidth]{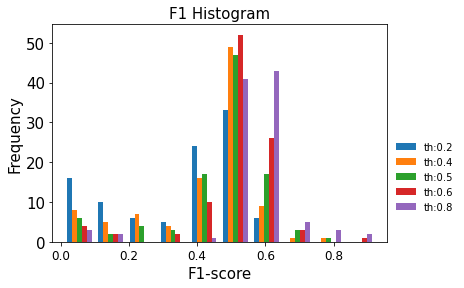} 
  \includegraphics[width=0.49\linewidth]{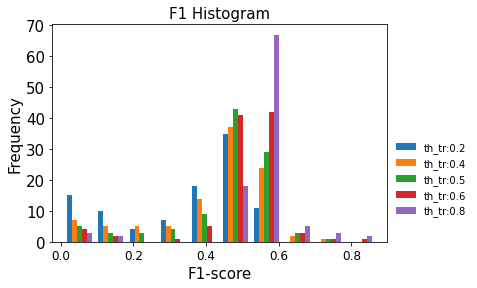}}
\vspace{-0.3cm}
\caption{Generator with linear activation.}\label{fig:Mix_CR_FA_L}
\end{subfigure}

\medskip

\begin{subfigure}[t]{.99\textwidth}
\centering
\vspace{0pt}
\includegraphics[width=0.49\linewidth]{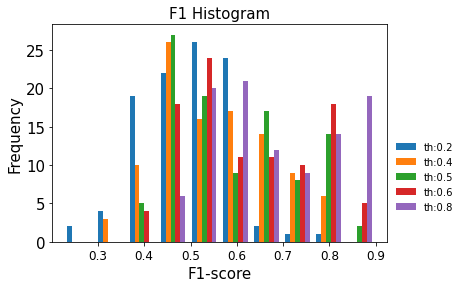}
\hfill
\includegraphics[width=0.49\linewidth]{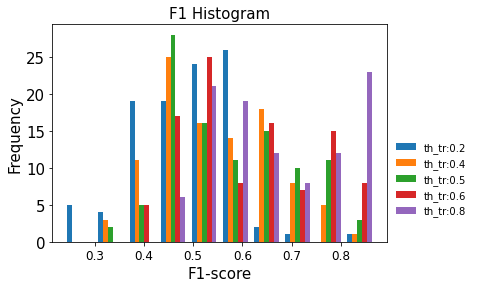}
\vspace{-0.3cm}
\caption{Generator with custom activation.}\label{fig:Mix_CR_FA_S}
\end{subfigure}

\medskip

\begin{subfigure}[t]{.99\textwidth}
\centering
\vspace{0pt}
\includegraphics[width=0.49\linewidth]{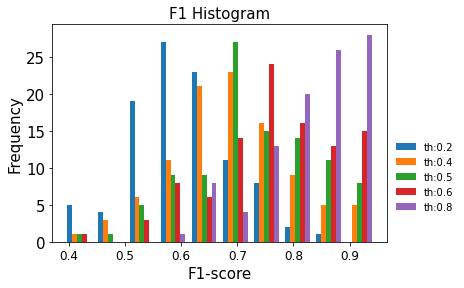}
\hfill
\includegraphics[width=0.49\linewidth]{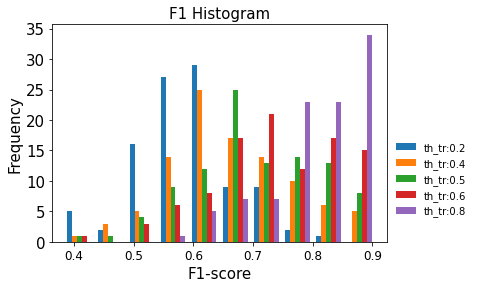}
\vspace{-0.3cm}
\caption{Generator with linear activation. Elitism criterion: Top 10 on F1-score.}\label{fig:Mix_top10_CR_FA_L}
\end{subfigure}

\medskip

\begin{subfigure}[t]{.99\textwidth}
\centering
\vspace{0pt}
\includegraphics[width=0.49\linewidth]{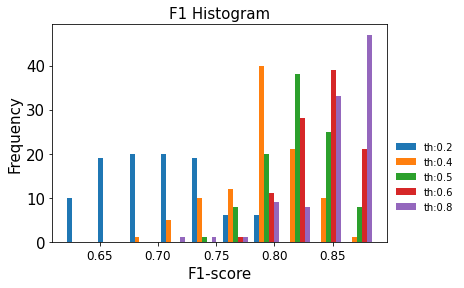}
\hfill
\includegraphics[width=0.49\linewidth]{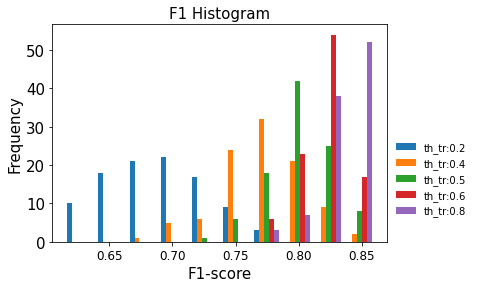}
\vspace{-0.3cm}
\caption{Generator with custom activation. Elitism criterion: Top 10 on F1-score.}\label{fig:Mix_top10_CR_FA_S}
\end{subfigure}

\end{mdframed}

\medskip
\begin{minipage}[t]{.99\textwidth}
\caption{Cryptomining attack data set (second use case). $F_1$-score on DS2-c (left) and DS1-c (right) using as training dataset GAN generators with linear and  custom activation. Random selection of generators was done from among all generators (\ref{fig:Mix_CR_FA_L} and \ref{fig:Mix_CR_FA_S}) and using an elitism criterion (top 10 of F1-scores) (\ref{fig:Mix_top10_CR_FA_L} and \ref{fig:Mix_top10_CR_FA_S}). Results for decision thresholds of 0.2, 0.4, 0.5, 0.6 and 0.8 are represented. Each experiment was run 100 times.} \label{fig:Mix_CR}
\end{minipage}

\end{figure*}

Table \ref{tab:CR_mix} summarizes the results obtained for the second use case and Figure \ref{fig:Mix_CR} shows detailed histograms of the $F_1$-score results obtained after running each experiment  $100$ times.
In a general sense, the results are aligned with the ones of the first use case:
\begin{enumerate}[(i)]
    \item ML classifiers trained with synthetic data generated by WGANs obtain a similar performance than when trained with real data. In fact, the best value of standard WGAN with $F_1$-elitism was slightly greater than the best value obtained with real data.
    \item The interval of the $F_1$-score values obtained when using the ST-based WGAN was concentrated near to the maximum value and this effect was accentuated when $F_1$-elitism was applied. 
Hence, selecting at random ST-based generator models is highly likely to obtain a synthetic data set that performs close to the best model combination and the real data.
    \item In contrast, when using the standard WGAN, the interval of $F_1$-score values is significantly wider and only a very small percentage of them are close to the best value obtained with real data, which precludes to use this method as it is not likely to produce a realistic synthetic data set when generators are selected at random. However, this effect slightly decreases when $F_1$-score elitism is used for selecting the generators.
\end{enumerate}

\begin{table}
\parbox{1.\linewidth}{
\caption{
Performance of synthetic traffic combining labels 0 and 1. Training with (i) a real data set (DS1), (ii) a mean-based noise generator and (iii and iv) GAN synthetic data (with linear and ST activation). Results on testing with real data (DS2). Experiment is drawn 100 times uniformly at random (Figure \ref{fig:Mix_CR_B} and Figure \ref{fig:Mix_CR}). For the GANs, in each sample we choose one generator from among all label 0 generators and one generator from among all label 1 generators.}
\label{tab:CR_mix}
\centering

\resizebox{.99\linewidth}{!}{%
\begin{tabular}{|c|c|c|c|}
\hline \textbf{Dataset} & \textbf{Quality Measure} & \textbf{Best} & \textbf{Default} \\\hline\hline

\multirow{3}{*}{\vspace{-0.3cm}$\begin{matrix}\textbf{Training 400K/4K }\\\textbf{Real data (DS1)}\end{matrix}$}  & \textit{{Threshold}} & 0.8 & 0.5 \\ 
\cline{2-4} & \textit{{$F_1$-score}} & 0.957  & 0.898 \\ \cline{2-4}
 & $\begin{matrix}\textit{{Confusion }} \\ \textit{{matrix}}\end{matrix}$ & \begin{tabular}{c|c} 399426  &  574 \\\hline 205  & 4183 \end{tabular} & \begin{tabular}{c|c} 399873 &   127 \\\hline 1384 &  3004 \end{tabular}\\ \hline\hline
 
 \multirow{3}{*}{\vspace{-0.3cm}$\begin{matrix}\textbf{Training 400K/4K }\\\textbf{Noise generator with means}\end{matrix}$}  & \textit{{Threshold}} & 0.8 & 0.5 \\ 
\cline{2-4} & \textit{{$F_1$-score}} & 0.710  & 0.624 \\ \cline{2-4}
 & $\begin{matrix}\textit{{Confusion }} \\ \textit{{matrix}}\end{matrix}$ & \begin{tabular}{c|c} 393386 &  6614 \\\hline 1364  & 3024 \end{tabular} & \begin{tabular}{c|c} 381851 & 18149 \\\hline 816  & 3572 \end{tabular}\\ \hline\hline
 
\multirow{3}{*}{\vspace{-0.3cm}$\begin{matrix}\textbf{Training 400K/4K }\\\textbf{WGAN with linear activation}\end{matrix}$}  & \textit{{Threshold}} & 0.8 & 0.5  \\ 
\cline{2-4} & \textit{{$F_1$-score}} & 0.920 & 0.820 \\ \cline{2-4}
 & $\begin{matrix}\textit{{Confusion }} \\ \textit{{matrix}}\end{matrix}$ & \begin{tabular}{c|c} 399692 &   308 \\\hline 969  & 3419 \end{tabular} & \begin{tabular}{c|c} 398651 &  1349 \\\hline 1657 &  2731 \end{tabular}\\ \hline\hline
 
 \multirow{3}{*}{\vspace{-0.3cm}$\begin{matrix}\textbf{Training 400K/4K }\\\textbf{WGAN with ST activation}\end{matrix}$}  & \textit{{Threshold}} & 0.8 & 0.5 \\ 
\cline{2-4} & \textit{{$F_1$-score}} & 0.897  & 0.869 \\ \cline{2-4}
 & $\begin{matrix}\textit{{Confusion }} \\ \textit{{matrix}}\end{matrix}$ & \begin{tabular}{c|c} 399780 &   220 \\\hline 1334 &  3054 \end{tabular} & \begin{tabular}{c|c} 399123  &  877 \\\hline 1288 &  3100 \end{tabular}\\ \hline \hline
 
 \multirow{3}{*}{\vspace{-0.3cm}$\begin{matrix}\textbf{Training 400K/4K }\\\textbf{WGAN with linear activation}\\\textbf{$F_1$-score elitism (top 10)}\end{matrix}$}  & \textit{{Threshold}} & 0.8 & 0.5 \\ 
\cline{2-4} & \textit{{$F_1$-score}} & 0.941  & 0.933 \\ \cline{2-4}
 & $\begin{matrix}\textit{{Confusion }} \\ \textit{{matrix}}\end{matrix}$ & \begin{tabular}{c|c} 399646  &  354 \\\hline 626 &  3762 \end{tabular} & \begin{tabular}{c|c} 399696  &  304 \\\hline 782 &  3606 \end{tabular}\\ \hline \hline

\multirow{3}{*}{\vspace{-0.3cm}$\begin{matrix}\textbf{Training 400K/4K }\\\textbf{WGAN with ST activation}\\\textbf{$F_1$-score elitism (top 10)}\end{matrix}$}  & \textit{{Threshold}} & 0.6 & 0.5 \\ 
\cline{2-4} & \textit{{$F_1$-score}} & 0.879  & 0.873 \\ \cline{2-4}
 & $\begin{matrix}\textit{{Confusion }} \\ \textit{{matrix}}\end{matrix}$ & \begin{tabular}{c|c} 399572 &   428 \\\hline 1425 &  2963 \end{tabular} & \begin{tabular}{c|c} 399261  &  739 \\\hline 1316 &  3072 \end{tabular}\\ \hline

\end{tabular}
}
}
\end{table}

\section{Conclusions}

We proposed a novel activation function to be used as output of the generator agent of a GAN. This activation function is based on the Smirnov probabilistic transformation (ST) and it is specifically designed to improve the quality of the generated data. The ST-based activation function provides a general approach that deals not only with the replication of categorical variables but with any type of data distribution (continuous or discrete).
Moreover, this activation function is derivable and therefore, it can be seamlessly integrated in the backpropagation computations during the GAN training processes.

The experimental results evidence a clear outperformance of the GAN network tuned with ST-based activation function with respect to a standard GAN. The quality of the data is so high that the generated data can fully substitute real data for training a nested classifier without a fall in the obtained accuracy. 
This result encourages the use of GANs to produce high-quality synthetic data that are applicable in scenarios in which data privacy must be guaranteed. 

\section*{Acknowledgements}

This work was partially supported by the European Union’s Horizon 2020 Research and Innovation Programme under Grant 833685 (SPIDER) and Grant 101015857 (Teraflow).
The first-named author wants to thank the hospitality of the Department of Mathematics at Universidad Aut\'onoma de Madrid, where this work was partially completed. 

\bibliography{sample}

\bibliographystyle{abbrv}

\end{document}